\documentclass{article}

\PassOptionsToPackage{numbers, compress, sort}{natbib}

\usepackage[final]{neurips_data_2024}

\usepackage[utf8]{inputenc} 
\usepackage[T1]{fontenc}    
\usepackage{url}            
\usepackage{booktabs}       
\usepackage{amsfonts}       
\usepackage{nicefrac}       
\usepackage{microtype}      
\usepackage{xcolor}         

\usepackage{amsmath}
\usepackage{amssymb}
\usepackage{mathtools}
\usepackage{amsthm}

\usepackage[labelfont=bf]{caption} 
\usepackage{graphicx} 
\usepackage{nicefrac} 
\usepackage{pifont}
\newcommand{\cmark}{\ding{51}}

\usepackage{color,colortbl} 
\definecolor{tablegray}{gray}{0.94} 
\usepackage{multirow} 
\usepackage{xcolor} 
\usepackage{adjustbox}
\usepackage{algorithmic}

\usepackage{wrapfig} 
\usepackage{enumitem} 
\usepackage{titletoc} 
\DeclareMathOperator*{\argmin}{arg\,min} 

\usepackage[hidelinks,colorlinks=true,citecolor=blue!50!black,linkcolor=black,urlcolor=green!50!black]{hyperref}
\usepackage[capitalize,noabbrev]{cleveref}
\usepackage[acronym]{glossaries}
\makeglossaries
\allowdisplaybreaks

\newglossaryentry{formula}{name=formula,
                           description={A mathematical expression}}

\newacronym{SOTA}{SOTA}{state-of-the-art}

\newacronym{iid}{iid}{independent and identically distributed}
\newacronym{iff}{iff}{if and only if}
\newacronym{wlog}{w.l.o.g.}{without loss of generality}

\newacronym{AI}{AI}{artificial intelligence}
\newacronym{XAI}{XAI}{explainable artificial intelligence}
\newacronym{ML}{ML}{machine learning}
\newacronym{DL}{DL}{deep learning}

\newacronym{SV}{SV}{Shapley Value}
\newacronym{BV}{BV}{Banzhaf Value}
\newacronym{CHII}{CHII}{Chaining II}
\newacronym{BII}{BII}{Banzhaf II}
\newacronym{SII}{SII}{Shapley II}
\newacronym{CHGV}{CHGV}{Chaining GV}
\newacronym{BGV}{BGV}{Banzhaf GV}
\newacronym{SGV}{SGV}{Shapley GV}
\newacronym{k-SII}{$k$-SII}{$k$-SV}
\newacronym{SI}{SI}{Shapley Interaction}
\newacronym{MI}{MI}{Möbius Interaction}
\newacronym{STII}{STII}{Shapley Taylor II}
\newacronym{FSII}{FSII}{Faithful SII}
\newacronym{FBII}{FBII}{Faithful BII}
\newacronym{BSHAP}{BShap}{Baseline Shapley}
\newacronym{II}{II}{Interaction Index}
\newacronym{GV}{GV}{Generalized Value}

\newacronym{LXAI}{Loc.\ Exp.}{Local Explanation}
\newacronym{GXAI}{Glo.\ Exp.}{Global Explanation}
\newacronym{TreeXAI}{Tree Exp.}{TreeSHAP-IQ Explanation}
\newacronym{UXAI}{Unc.\ Exp.}{Uncertainty Explanation}
\newacronym{FS}{Fea.\ Sel.}{Feature Selection}
\newacronym{ES}{Ens.\ Sel.}{Ensemble Selection}
\newacronym{RFES}{RF Ens.\ Sel.}{RF Ensemble Selection}
\newacronym{DV}{Dat.\ Val.}{Data Valuation}
\newacronym{DSV}{Dst.\ Val.}{Dataset Valuation}
\newacronym{CE}{Clu.\ Exp.}{Cluster Explanation}
\newacronym{UFI}{Uns.\ FI}{Unsupervised Feature Imp.}
\newacronym{SOUM}{SOUM}{Sum of Unanimity Model}
\newacronym{UM}{Una.}{Unanimity}

\newacronym{GT}{GT}{ground truth}
\newacronym{MSE}{MSE}{mean squared error}
\newacronym{prec}{Precision@5}{precision at five}
\newacronym{MAE}{MAE}{mean absolute error}

\usepackage{xspace}

\newcommand{\shapiq}{\texttt{shapiq}\xspace}

% ...

% ...

% ...

% ...

% ...
\def\bx{\mathbf{x}}

\def\bxsb{\mathbf{x}_{\bar{S}}}
\def\xs{x_S}
\def\xsb{x_{\bar{S}}}
% ...

% ...

\usepackage[frozencache,cachedir=.]{minted} 
\definecolor{codebackground}{RGB}{240, 240, 235}
\definecolor{highlightcolor}{RGB}{218, 218, 213}
\newsavebox{\mintedbox}
\newenvironment{boxminted}[1]
 {%
  \VerbatimEnvironment
  \RecustomVerbatimEnvironment{Verbatim}{BVerbatim}{}%
  \begin{lrbox}{\mintedbox}
  \begin{minted}[highlightlines={#1},highlightcolor=highlightcolor]%
 }
 {%
  \end{minted}%
  \end{lrbox}%
  \noindent\colorbox{codebackground}{\makebox[\textwidth][l]{\usebox{\mintedbox}}}%
 }

\title{shapiq: Shapley Interactions for Machine Learning}

\author{%
  Maximilian Muschalik\\
  LMU Munich,\\
  Munich Center for Machine Learning
  \And 
  Hubert Baniecki\\
  University of Warsaw,\\
  Warsaw University of Technology\\
  \And
  Fabian Fumagalli\\
  Bielefeld University, CITEC
  \And
  Patrick Kolpaczki\\
  Paderborn University
  \And
  Barbara Hammer\\
  Bielefeld University, CITEC
  \And
  Eyke Hüllermeier\\
  LMU Munich,\\
  Munich Center for Machine Learning
}

\begin{document}

\maketitle

\begin{abstract}
Originally rooted in game theory, the \gls{SV} has recently become an important tool in machine learning research.
Perhaps most notably, it is used for feature attribution and data valuation in explainable artificial intelligence.
\glspl{SI} naturally extend the \gls{SV} and address its limitations by assigning joint contributions to groups of entities, which enhance understanding of black box machine learning models.
Due to the exponential complexity of computing \glspl{SV} and \glspl{SI}, various methods have been proposed that exploit structural assumptions or yield probabilistic estimates given limited resources.
In this work, we introduce \shapiq, an open-source Python package that unifies state-of-the-art algorithms to efficiently compute \glspl{SV} and any-order \glspl{SI} in an application-agnostic framework.
Moreover, it includes a benchmarking suite containing 11 machine learning applications of \glspl{SI} with pre-computed games and ground-truth values to systematically assess computational performance across domains.
For practitioners, \shapiq is able to explain and visualize any-order feature interactions in predictions of models, including vision transformers, language models, as well as XGBoost and LightGBM with TreeSHAP-IQ.
With \shapiq, we extend \texttt{shap} beyond feature attributions and consolidate the application of \glspl{SV} and \glspl{SI} in machine learning that facilitates future research. 
The source code and documentation are available at \url{https://github.com/mmschlk/shapiq}.
\end{abstract}

\glsresetall

\section{Introduction}

\begin{figure}
    \centering
    \includegraphics[width=\textwidth]{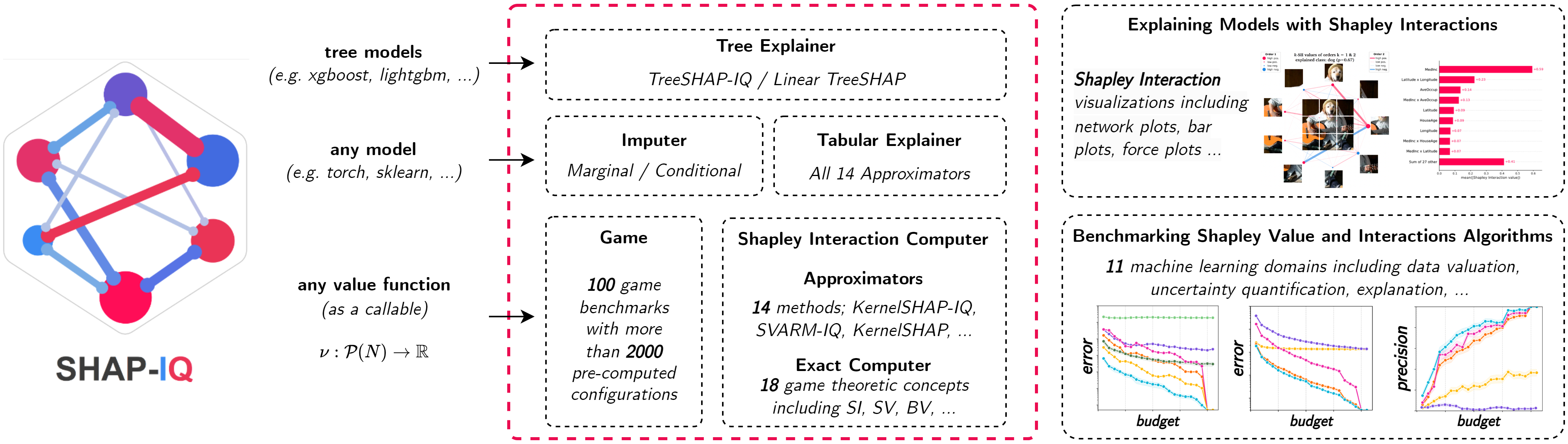}
    \caption{The \shapiq Python package facilitates research on game theory for machine learning, including state-of-the-art approximation algorithms and pre-computed benchmarks. Moreover, it provides a simple interface for explaining predictions of machine learning models beyond feature attributions.}
    \label{fig:abstract}
\end{figure}

Assigning \emph{value} to entities collectively performing a task is essential in various real-world applications of \gls{ML} \cite{DBLP:conf/ijcai/RozemberczkiWBY22,olsen2024comparative}.
For instance, when reimbursing data providers based on the \emph{value of data}~\cite{Ghorbani.2019,Tay.2022}, or justifying a model's prediction based on \emph{value of feature information}~\cite{Strumbelj.2014,Lundberg.2017,Covert.2020, Covert.2021b, Chen.2023}.
The \emph{fair} distribution of value among a group of entities is a central aspect of cooperative game theory, where the \gls{SV} \cite{Shapley.1953} defines a \emph{unique} allocation scheme based on intuitive axioms. 
The \gls{SV} is applicable to any \emph{game}, i.e.\ a function that specifies the worth of all possible groups of entities, called coalitions.
In \gls{ML}, application-specific games were introduced~\cite{DBLP:conf/ijcai/RozemberczkiWBY22,Ghorbani.2019,Tay.2022,Balestra.2022,Watson.2023}, which typically require a definition of the overall worth and a notion of entities' absence~\cite{Covert.2021b}.
The \gls{SV} fairly distributes the overall worth among individuals by evaluating the game for all coalitions.
However, it does not give insights on \emph{synergies or redundancies} between entities.
For instance, while two features such as \emph{latitude} and \emph{longitude} convey separate information, only their joint consideration reveals 
the synergy of encoding an \emph{exact location}.
The value of such a group of entities is known as an \emph{interaction} \cite{Grabisch.1999}, or in this context \emph{feature interaction} \cite{Fumagalli.2023}, and is crucial to understand predictions of complex \gls{ML} models~\cite{Kumar.2020, Tsang.2020, Kumar.2021, deng2022discovering, Patel.2021, Sundararajan.2020, Muschalik.2024, Wright.2016}, as illustrated in \cref{fig:abstract}.

\glspl{SI} \cite{Grabisch.1999,Sundararajan.2020,Tsai.2022,Bord.2023} distribute the overall worth to all groups of entities up to a maximum \emph{explanation order}. They satisfy axioms similar to the \gls{SV}, to which they reduce for individuals, i.e.\ the lowest explanation order.
In contrast, for the highest explanation order, which comprises an interaction for every coalition, the \glspl{SI} yield the \gls{MI}, or Möbius transform~\cite{rota1964foundations,harsanyi1963simplified,Fujimoto.2006}.
The \glspl{MI} are a fundamental concept in cooperative game theory that captures the \emph{isolated joint contribution}, which allows to additively describe every coalition's worth by a sum of the contained \glspl{MI}.
With an increasing explanation order, the \glspl{SI} comprise more components that finally yield the \glspl{MI} as the most comprehensive explanation of the game at the cost of highest complexity~\cite{Bord.2023,Tsai.2022}.
While the \gls{SV} and \glspl{SI} provide an appealing theoretical concept, computing them without structural assumptions on the game requires exponential complexity \cite{Deng.1994}.
For tree-based models, it was shown that \glspl{SV} \cite{Lundberg.2020,Yu.2022} and \glspl{SI} \cite{Zern.2023,Muschalik.2024} can be efficiently computed by exploiting the architecture.
Moreover, game-agnostic stochastic approximators estimate the SV~\cite{Castro.2009,Maleki.2013,Owen.2014,Lundberg.2017,Covert.2021,Pelegrina.2023,Kolpaczki.2024a} and \glspl{SI} \cite{Sundararajan.2020,Tsai.2022,Fumagalli.2023,Kolpaczki.2024b,Fumagalli.2024} with a limited budget of game evaluations.

Diverse applications of the \gls{SV} have led to various techniques for its efficient computation~\cite{Chen.2023}.
Recently, extensions to any-order \glspl{SI} addressed limitations of the \gls{SV} and complemented interpretation of model predictions with higher-order feature interactions~\cite{Tsang.2020,Sundararajan.2020, Tsai.2022, Bord.2023, Fumagalli.2023}.
While stochastic approximators are applicable to any game, their evaluation is typically performed in an isolated application \cite{Fumagalli.2023,li2024faster}, such as feature interactions.
Moreover, implementing such algorithms requires a strong mathematical background and specific design choices.
Existing Python packages, such as \texttt{shap}~\citep{Lundberg.2017}, provide a relatively small number of approximators, which are limited to the \gls{SV} and feature attributions.

\paragraph{Contribution.}
In this work, keeping within the scope of the NeurIPS 2024 Datasets \& Benchmark track, we present \shapiq, an open-source \emph{Python} library for any-order \glspl{SI} that consolidates research for computing \glspl{SV} and \glspl{SI} across \gls{ML} domains.
Therein, we contribute
\begin{enumerate}
    \item[\textbf{1.}] a general \texttt{approximation} interface for state-of-the art \gls{SI} algorithms and methods without focus on a specific application like explanations or data valuation,
    \item[\textbf{2.}] an explanation API for using \glspl{SI} to explain \gls{ML} models and visualizing interactions,
    \item[\textbf{3.}] a benchmarking suite to evaluate \gls{SI} approximators across several real-world scenarios,
    \item[\textbf{4.}] and a cross-domain empirical evaluation of approximators guiding practitioners.
\end{enumerate}

\paragraph{Related software tools and benchmarks.} \shapiq extends the popular \texttt{shap}~\citep{Lundberg.2017} Python package beyond feature attributions aiming to fully embrace the application of \glspl{SV} and \glspl{SI} in \gls{ML}. While \texttt{shap} implements a single index for 2-order feature interactions to explain the predictions of tree-based models, \shapiq implements a dozen approximators for any-order \glspl{SI} and offers a benchmarking suite for these algorithms across 10 different domains~(\cref{tab_method_overview}). Related software such as \texttt{aix360}~\citep{arya2020aix360}, \texttt{alibi}~\citep{klaise2021alibi} and \texttt{dalex}~\citep{baniecki2021dalex} are general toolboxes offering the implementation and visualization of the most popular \gls{ML} explanations for end-users. We specify in \glspl{SI} to provide a comprehensive tool facilitating research in game theory for \gls{ML}, including the exact computation of 18 interaction indices and game-theoretic concepts (\cref{tab_background_indices}). Notably, the \texttt{innvestigate}~\citep{alber2019innvestigate} and \texttt{captum}~\citep{kokhlikyan2020captum} Python packages offer feature attribution explanation methods specific to (deep) neural networks. Most recently, \texttt{quantus}~\citep{hedstrom2023quantus} implements evaluation metrics for these explanation methods.

We build upon recent advances in benchmarking \gls{XAI} methods such as feature attributions~\citep{liu2021synthetic,agarwal2022openxai,li2023m4,olsen2024comparative} and algorithms for data valuation~\citep{jiang2023opendataval}. \texttt{XAI-Bench}~\citep{liu2021synthetic} focuses on synthetic tabular data. \texttt{OpenXAI}~\citep{agarwal2022openxai} provides 7 real-world tabular datasets with pre-trained neural network models, 7 feature attribution methods and 8 metrics to compare them. $\mathcal{M}^{4}$~\citep{li2023m4} extends \texttt{OpenXAI} to benchmark feature attributions of deep neural networks for image and text modalities. In~\citep{olsen2024comparative}, the authors benchmark several algorithms for approximating \glspl{SV} based on the conditional feature distribution. \texttt{OpenDataVal}~\citep{jiang2023opendataval} provides 9 real-world datasets, 11 data valuation methods and 4 metrics to compare them. \shapiq puts more focus on benchmarking higher-order \gls{SI} algorithms and provides an interface to state-of-the-art explanation methods that base on \glspl{SI}, e.g.\ TreeSHAP-IQ~\citep{Muschalik.2024}.  While open data repositories such as \texttt{OpenML}~\citep{bischl2021openml} offer easy access to datasets for \gls{ML}, we pre-compute and share ground-truth \glspl{SI} for various games (i.e.\ dataset--model pairs) that saves considerate time and resources when benchmarking approximation algorithms.

\section{Theoretical Background}\label{sec_theoretical_background}

\begin{table}[t]
\centering
\caption{Available concepts in the \texttt{ExactComputer} class in \protect{\shapiq} with \textbf{SIs} in \textbf{bold}.}
\vspace{0.5em}
\label{tab_background_indices}
{\footnotesize
\begin{tabular}{@{}ll|l|l@{}}
\toprule
\textbf{Setting} &
\begin{tabular}[l]{@{}l@{}}\textbf{Interaction Index (II)} \cite{Fujimoto.2006}\end{tabular} &
\begin{tabular}[l]{@{}l@{}}\textbf{Base Semivalue} \cite{Dubey.1981} \end{tabular} &
\begin{tabular}[l]{@{}l@{}}\textbf{Generalized Value (GV)} \cite{DBLP:journals/dam/MarichalKF07} \end{tabular}
\\
\midrule
\multirow{4}{*}{\textbf{\begin{tabular}[l]{@{}l@{}}Machine \\ Learning \end{tabular}}} &
\textbf{$k$-Shapley Values ($k$-SII)} \cite{Bord.2023} &
\multirow{4}{*}{\begin{tabular}[l]{@{}l@{}}\textbf{Shapley (SV)} \cite{Shapley.1953} \end{tabular}} &
\multirow{4}{*}{\begin{tabular}[l]{@{}l@{}} Joint SV \cite{Harris.2022} \end{tabular}}  
\\
& \textbf{Shapley Taylor II (STII)} \cite{Sundararajan.2020} & &
\\
& \textbf{Faithful Shapley II (FSII)} \cite{Tsai.2022} & & 
\\
& $k_\text{ADD}$-SHAP \cite{Pelegrina.2023} & & 
\\
& Faithful Banzhaf II (FBII) \cite{Tsai.2022}  & Banzhaf (BV) \cite{Banzhaf.1964} & --
\\ \midrule
\multirow{5}{*}{\textbf{\begin{tabular}[l]{@{}l@{}}Game \\ Theory \end{tabular}}} &
\textbf{M\"obius (MI)} \cite{harsanyi1963simplified,rota1964foundations,Fujimoto.2006} & \multirow{2}{*}{ -- } &
Internal GV (IGV) \cite{DBLP:journals/dam/MarichalKF07}
\\
& Co-M\"obius (Co-MI) \cite{Grabisch.2000} & & External GV (EGV) \cite{DBLP:journals/dam/MarichalKF07} 
\\
& Shapley II (SII) \cite{Grabisch.1999} & 
\multirow{2}{*}{\begin{tabular}[c]{@{}c@{}}\textbf{Shapley (SV)} \cite{Shapley.1953} \end{tabular}} &
Shapley GV (SGV) \cite{Marichal_2000} 
\\
& Chaining II (CHII) \cite{Marichal.1999} & & 
Chaining GV (CHGV) \cite{DBLP:journals/dam/MarichalKF07} 
\\
& Banzhaf II (BII) \cite{Grabisch.1999} &
Banzhaf (BV) \cite{Banzhaf.1964} & 
Banzhaf GV (BGV) \cite{Marichal.1999} 
\\
\bottomrule
\end{tabular}
}
\end{table}

In \gls{ML}, various concepts are based on synergies of entities to optimize performance in a given task.
For example, weak learners construct powerful model ensembles \citep{Rozemberczki.2021}, collected data instances and features are used to train supervised \gls{ML} models \cite{Ghorbani.2019,Cohen.2007}, where feature values collectively predict outputs.
To better understand such processes, \gls{XAI} quantifies the contributions of these entities to the task, most prominently for feature values in predictions (local feature attribution \cite{Strumbelj.2014,Lundberg.2017}), features in models (global feature importance \cite{Cohen.2007,Pfannschmidt.2016,Covert.2021}), and data instances in model training (data valuation \cite{Ghorbani.2019}).
Assigning such contributions is closely related to the field of cooperative game theory \cite{DBLP:conf/ijcai/RozemberczkiWBY22}, which studies the notion of value for players that collectively obtain a payout.
To adequately assess the impact of individual players, it is necessary to analyze the payout for different coalitions.
More formally, a cooperative game $\nu: \mathcal{P}(N) \rightarrow \mathbb{R}$ with $\nu(\emptyset)=0$ is defined by a \emph{value function} on the power set of $N := \{1,\dots,n\}$ entities, which describes such payouts for all possible coalitions of players.
We later summarize such prominent examples in the context of \gls{ML} in~\cref{tab_benchmark_overview}.
Here, we summarize existing contribution concepts for individuals and groups of entities, outlined in~\cref{tab_background_indices}.

The \textbf{\gls{SV}} \cite{Shapley.1953} and \textbf{\gls{BV}} \cite{Banzhaf.1964} are instances of \emph{semivalues} \cite{Dubey.1981}.
Semivalues assign contributions to individual players and adhere to intuitive axioms: \emph{Linearity} enforces linearly composed contributions for linearly composed games; 
\emph{Dummy} requires that players without impact receive zero contribution; 
\emph{Symmetry} enforces that entities contributing equally to the payout receive equal value. 
The \gls{SV} \cite{Shapley.1953} is the unique semivalue that additionally satisfies \emph{efficiency}, i.e.\ the sum of all contributions yields the total payout $\nu(N)$.
In contrast, the \gls{BV} \cite{Banzhaf.1964} is the unique semivalue that additionally satisfies \emph{2-Efficiency}, i.e.\ the contributions of two players sum to the contribution of a joint player in a reduced game, where both players are merged.
The \gls{SV} and \gls{BV} are represented as a weighted average over \emph{marginal contributions} $\Delta_i(T) := \nu(T \cup \{i\}) - \nu(T)$ for $i \in N$ as
\begin{align*}
    &\phi^{\text{SV}}(i) := \sum_{T \subseteq N \setminus \{i\}} \frac{1}{n \binom{n-1}{\vert T \vert}} \Delta_i(T)
    &&\text{ and }    
    &&\phi^{\text{BV}}(i) := \sum_{T \subseteq N \setminus \{i\}} \frac{1}{2^{n-1}} \Delta_i(T) \, .
\end{align*}

In \gls{ML} applications, the \gls{SV} is typically preferred over the \gls{BV} due to the efficiency axiom \cite{DBLP:conf/ijcai/RozemberczkiWBY22}.
For instance, in local feature attribution, the \gls{SV} is utilized to fairly distribute the model's prediction to individual features \cite{Strumbelj.2014,Lundberg.2017}.
However, it was shown that the \gls{SV} is limited when explaining complex decision systems, and \emph{feature interactions}, i.e.\ the joint contributions of features' groups, are required to understand such processes~\cite{Kumar.2020, Tsang.2020, Kumar.2021, deng2022discovering, Tsai.2022, Patel.2021, Fumagalli.2023,Muschalik.2024, Sundararajan.2020, Wright.2016}.

The \textbf{\gls{GV}} \cite{DBLP:journals/dam/MarichalKF07} and \textbf{\gls{II}} \cite{Fujimoto.2006}  are two paradigms to extend the notion of value to groups of entities.
The \glspl{GV} are based on weighted averages over (joint) marginal contributions $\nu(T \cup S) - \nu(T)$ for $S \subseteq N$ given $T \subseteq N \setminus S$.
In contrast, \glspl{II} are based on discrete derivatives that account for lower-order effects of subsets of $S$.
For instance, for two players $i,j \in N$, the discrete derivative $\Delta_{\{i,j\}}(T)$ is defined as the joint marginal contribution $\nu(T \cup \{i,j\}) - \nu(T)$ minus the individual marginal contributions $\Delta_i(T)$ and $\Delta_j(T)$.
More generally, the \emph{discrete derivative} $\Delta_S(T)$ for $S\subseteq N$ in the presence of $T \subseteq N \setminus S$ is defined as 
\begin{align*}
    \Delta_S(T) := \sum_{L \subseteq S} (-1)^{\vert S \vert-\vert L \vert} \nu(T \cup L) \hspace{0.2cm} \text{ with } \hspace{0.2cm} \Delta_S(T) = \underbrace{\nu(T \cup S) - \nu(T)}_{\text{joint marginal contribution}} - \sum_{\emptyset \neq L\subset S}\underbrace{\Delta_L(T)}_{\text{lower-order effects}} \, .
\end{align*}
A positive value indicates synergy, whereas a negative value indicates redundancy of $S$ given $T$.
Lastly, a zero value indicates (additive) independence, i.e. the joint marginal contribution is equal to the sum of all lower-order effects.
\glspl{GV} and \glspl{II} are uniquely represented \cite{DBLP:journals/dam/MarichalKF07,Fujimoto.2006} by
\begin{align*}    
    &\phi^{\text{GV}}(S) := \sum_{T \subseteq N \setminus S} p_{\vert T \vert}^{\vert S \vert}(n)  \left(\nu(T \cup S) - \nu(T)\right)  &&\text{ and }   &&\phi^{\text{II}}(S) := \sum_{T \subseteq N \setminus S} p_{\vert T \vert}^{\vert S \vert}(n) \Delta_S(T) \, .
\end{align*}
The most prominent examples are the \emph{\gls{SGV}}~\cite{Marichal_2000} and the \emph{\gls{SII}} \cite{Grabisch.1999} with $p_t^s(n) = \big((n-s+1)\binom{n-s}{t}\big)^{-1}$, which naturally extend the \gls{SV} (cf.\ \cref{appx_sec_gv_ii}).
While the \gls{SGV} and \gls{SII} are natural extensions to the \gls{SV}, they are not suitable for interpretability, since they are defined on the powerset and comprise an exponential number of components.
Moreover, neither \glspl{GV} nor \glspl{II} satisfy the efficiency axiom for higher-orders, which is desirable for \gls{ML} applications.

\textbf{Shapley Interactions (SIs) for Machine Learning} assign joint contribution $\Phi_k$ up to an \emph{explanation order} $k$, i.e.\ for all coalitions $S \subseteq N$ with $\vert S \vert \leq k$, which satisfy generalized \emph{efficiency}
$
    \nu(N) = \sum_{S\subseteq N, \vert S \vert \leq k}\Phi_k(S) \, .
$
The \emph{\glspl{k-SII}} \cite{Bord.2023} are the unique \glspl{SI} that coincide with SII for the highest order.
The \emph{\gls{STII}} \cite{Sundararajan.2020} puts a stronger emphasis on the top-order interactions, and \emph{\gls{FSII}}~\cite{Tsai.2022} optimizes \emph{Shapley-weighted faithfulness}
\begin{align*}
    \mathcal{L}(\nu,\Phi_k) := \sum_{T \subseteq N} \mu(t) \left(\nu(T) - \sum_{S \subseteq T, \vert S \vert \leq k} \Phi_k(S)\right)^2 \text{ with } \mu(t) := \begin{cases}
        \mu_\infty &\text{ if } t \in \{0,n\} 
        \\
        \frac{1}{\binom{n-2}{t-1}} &\text{ else}
    \end{cases} \, .
\end{align*}
\gls{FSII} is thus $\Phi_k^{\text{FSII}} := \argmin_{\Phi_k} \mathcal{L}(\nu,\Phi_k)$, where $\mu_\infty \gg 1$ ensures efficiency.
It was recently~shown that pairwise SII and consequently \gls{k-SII} with $k=2$ optimize a faithfulness metric with slightly different weights \cite{Fumagalli.2024}.
For $k=1$, all \glspl{SI} reduce to the \gls{SV} $\Phi_1 \equiv \phi^{\text{SV}}$, which minimizes faithfulness~$\mathcal{L}(\nu,\Phi_1)$~\cite{Charnes.1988} with the efficiency constraint, or equivalently $\mu_\infty \to \infty$~\cite{Lundberg.2017,Fumagalli.2024}.
Finally, for $k=n$, all \glspl{SI} $\Phi_n$ are the \emph{\glspl{MI}} (cf.\ \cref{appx_sec_mi}), which are faithful to all game values, i.e.\ $\mathcal{L}(\nu,\Phi_n) = 0$.
Notably, all \glspl{SI} can be uniquely represented by the \glspl{MI} \cite{Grabisch.2016}.
In this context, \glspl{SI} yield a complexity-accuracy trade-off, ranging from the least complex (\gls{SV}) to the most comprehensive (\gls{MI}) explanation.
Other extensions include \emph{$k_\text{ADD}$-SHAP} \cite{Pelegrina.2023} of the \gls{SV} and \emph{\gls{FBII}}~\cite{Tsai.2022} of the \gls{BV}, which do not satisfy efficiency, as well as \emph{Joint \glspl{SV}} \cite{Harris.2022}, a \gls{GV} with efficiency.
However, in the context of feature interactions and \gls{ML}, \glspl{SI} are preferred over \gls{GV}-based (Joint \glspl{SV}) or \gls{BV}-based \glspl{II}~(\gls{FBII}), as they account for lower-order \emph{interactions} and adhere to the \gls{SV} and \gls{MI} as edge cases.

\section{Overview of the \shapiq Python package}\label{sec_shapiq_overview}

\begin{table}[t]
\centering
\caption{Overview of methods in \protect{\shapiq} and applicable \glspl{SI}. Explainers rely on approximators or model assumptions. (\cmark) indicates only top-order approximation.}
\vspace{0.5em}
\label{tab_method_overview}
{\footnotesize
\begin{tabular}{@{}lllcccc@{}}
\toprule
\textbf{Class} & \textbf{Implementation} & \textbf{Source} & \textbf{SV} & \textbf{($k$-)SII} & \textbf{STII} & \textbf{FSII} \\ \midrule
\multirow{14}{*}{\textbf{\vspace{-0.165em}Approximator}} 
 & SHAP-IQ & \citep{Fumagalli.2023} & \cmark & \cmark & \cmark & (\cmark)\\ 
 & SVARM-IQ & \citep{Kolpaczki.2024b} & \cmark & \cmark & \cmark & (\cmark) \\
 & Permutation Sampling (SII) & \citep{Tsai.2022} & \cmark & \cmark & -- & -- \\
 & Permutation Sampling (STII) & \citep{Sundararajan.2020} & \cmark & -- & \cmark & -- \\
 & KernelSHAP-IQ & \citep{Fumagalli.2024} & \cmark & \cmark & -- & -- \\
 & Inconsistent KernelSHAP-IQ & \citep{Fumagalli.2024} & \cmark & \cmark & -- & -- \\
 & FSII Regression & \citep{Tsai.2022} & \cmark & -- & -- & \cmark \\
 \cmidrule{2-7}
 & KernelSHAP & \citep{Lundberg.2017} & \cmark & -- & -- & --\\
 & $k_\text{ADD}$-SHAP & \citep{Pelegrina.2023} & \cmark & -- & -- & -- \\
 & Unbiased KernelSHAP & \citep{Covert.2021}  & \cmark & -- & -- & -- \\
 & SVARM & \citep{Kolpaczki.2024a} &  \cmark & -- & -- & -- \\
 & Permutation Sampling & \citep{Castro.2009}  & \cmark & -- & -- & -- \\
 & Owen Sampling & \citep{Owen.2014} & \cmark & -- & -- & -- \\
 & Stratified Sampling & \citep{Maleki.2013} & \cmark & -- & -- & --\\
 \midrule
\multirow{4}{*}{\textbf{Explainer}} 
 & Agnostic (Marginal) & -- & \cmark & {\cmark} & \cmark & \cmark \\ 
 & Agnostic (Conditional) & -- & \cmark & {\cmark} & \cmark & \cmark \\ 
 & TreeSHAP-IQ & \citep{Muschalik.2024} & \cmark & {\cmark} & \cmark & (\cmark) \\ 
 & Linear TreeSHAP & \citep{Yu.2022,Lundberg.2020} & \cmark & -- & -- & -- \\
  \midrule
 \multirow{2}{*}{\textbf{Computer}} & Möbius Converter & -- & \cmark & \cmark & \cmark & \cmark \\
 & Exact Computer & -- & \cmark & \cmark & \cmark & \cmark \\ 
 \bottomrule
\end{tabular}%
}
\end{table}

The \shapiq package accelerates research on \glspl{SI} for ML, and provides a simple interface for explaining any-order feature interactions in predictions of \gls{ML} models. Its code is open-source on GitHub at \url{https://github.com/mmschlk} while the documentation with notebook examples and API reference is available at \url{https://shapiq.readthedocs.io}.

\subsection{\shapiq Facilitates Research on Shapley Interactions for Machine Learning}

\textbf{Approximators.} We implement 7 algorithms for approximating \glspl{SI} across 4 different interaction indices, and another 7 algorithms for approximating \glspl{SV}.
\cref{tab_method_overview} provides a comprehensive overview of this effort, where the \texttt{shapiq.Approximator} class is extended with each implementation.
We unify common approximation methods by including a general \texttt{shapiq.CoalitionSampler} interface offering approximation performance increases through sampling procedures like the \textit{border-} and \textit{pairing-tricks} introduced in~\cite{Covert.2021,Fumagalli.2023}. 
Algorithms are primarily benchmarked based on how well they approximate the ground truth \glspl{SI} values that often cannot be computed in practice due to exponential complexity and constrained resources.

\textbf{Exact computer.} A key functionality of \shapiq lies in computing the \glspl{SI} \emph{exactly}, which is feasible for smaller games, but reaches its limit for growing player numbers. The \texttt{shapiq.ExactComputer} class provides an interface for computing 18 interaction indices and game-theoretic concepts, including the \glspl{MI} (see \cref{tab_background_indices}).

\textbf{Games.} Approximators and computers work given a specified cooperative game. Table~\ref{tab_benchmark_overview} describes in detail 11 benchmark games implemented in \shapiq. Beyond synthetic games, our benchmark spans the 5 most prominent domains where \glspl{SI} can be applied for \gls{ML}. The \texttt{shapiq.Game} class can be easily extended to include future benchmarks in the package. We pre-compute and share exact \glspl{SI} for $2\,042$ benchmark game configurations in total (\cref{appx_sec_overview_benchmark}), facilitating future work on improving the approximators, which we elaborate on further in Section~\ref{sec_benchmark}.

\subsection{Explaining Machine Learning Predictions with \shapiq}

\begin{figure}
    \begin{minipage}[l]{0.60\textwidth}
{\scriptsize
\begin{boxminted}{}{Python}
X, model = ...
import shapiq
# create an explainer object
explainer = shapiq.Explainer(model=model, data=X, max_order=3)
# choose a sample point to be explained
x = X[0]
# approximate feature interactions given the specificed budget 
interaction_values = explainer.explain(x=x, budget=1024)
# retrieve 3-order feature interactions 
interaction_values.get_n_order_values(3)
# visualize 1-order and 2-order feature interactions on a graph 
interaction_values.plot_network(feature_names=...)
\end{boxminted}
}
\hfill
\end{minipage}
\begin{minipage}[r]{0.40\textwidth}
    \begin{flushright}
    \includegraphics[width=0.9\textwidth]{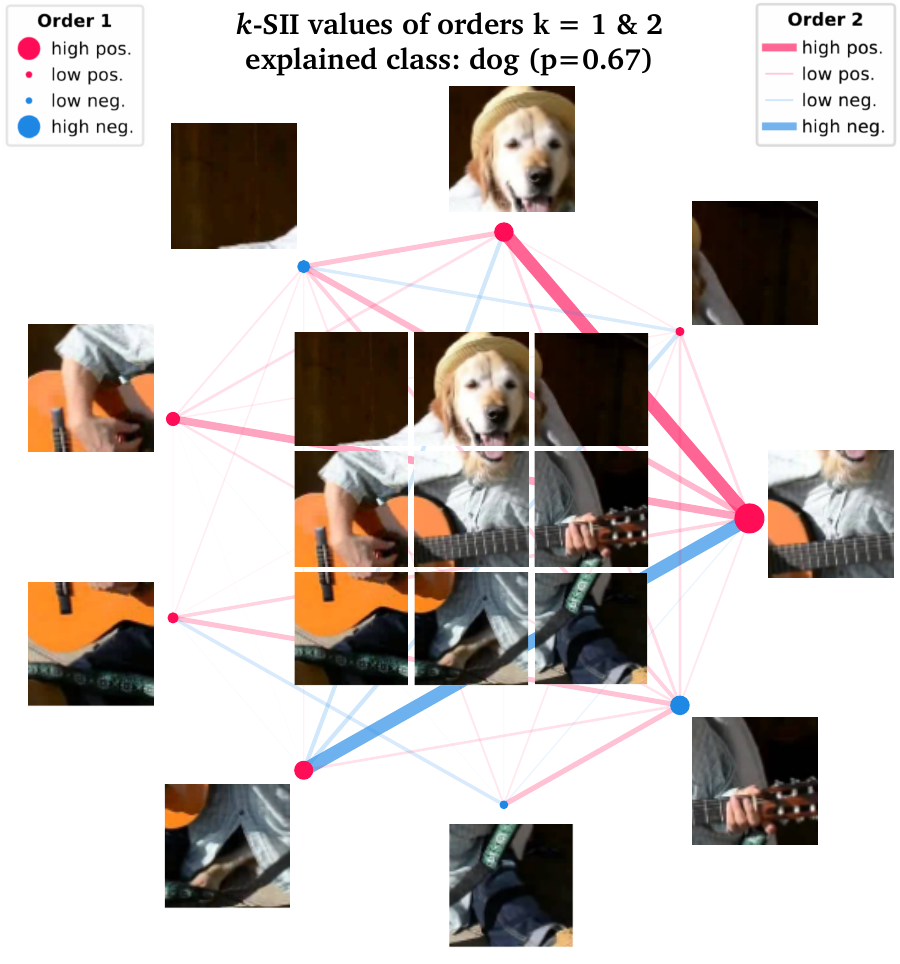}
    \end{flushright}
\end{minipage}
\caption{\textbf{Left:} Exemplary code for locally explaining a single model's prediction with \shapiq. \textbf{Right:} Local feature interactions visualized on a network plot.}
\label{fig_local_explanation}
\end{figure}

\begin{figure}
    \begin{minipage}[l]{0.60\textwidth}
{\scriptsize
\begin{boxminted}{}{Python}
X, model = ...
import shapiq
# create an explainer object
explainer = shapiq.Explainer(model=model, data=X, max_order=3)
# approximate feature interactions for multiple sample points
interaction_values_list = explainer.explain_X(X, budget=1024)
# retrieve 3-order feature interactions for the first point
interaction_values_list[0].get_n_order_values(3)
# visualize the global feature interaction importance  
shapiq.plot.bar_plot(interaction_values_list, feature_names=...)
\end{boxminted}
}
\hfill
\end{minipage}
\begin{minipage}[r]{0.40\textwidth}
    \begin{flushright}
    \includegraphics[width=0.95\textwidth]{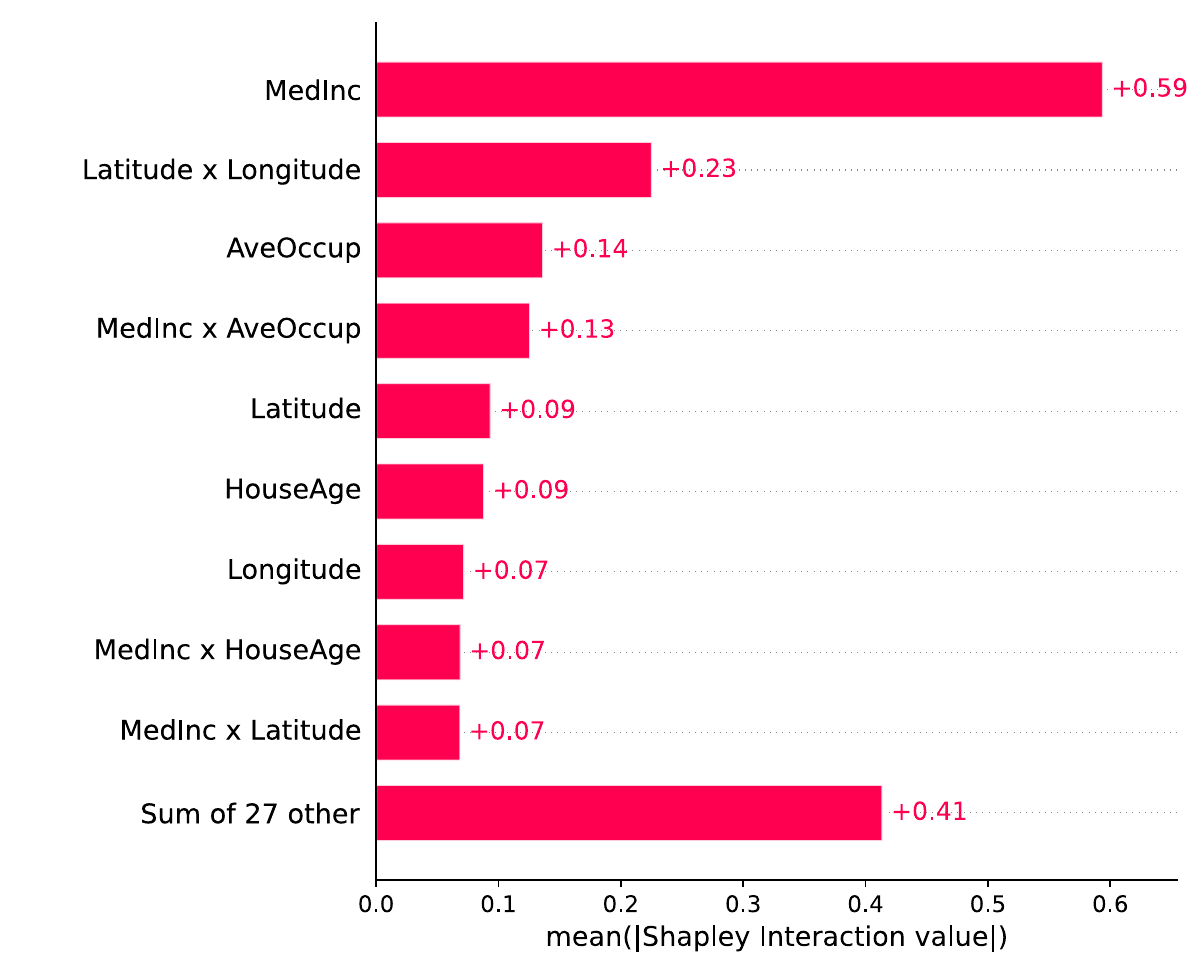}
    \end{flushright}
\end{minipage}
\caption{\textbf{Left:} Exemplary code for globally explaining multiple model's predictions with \shapiq. \textbf{Right:} Global feature interaction importance visualized as a bar plot.}
\label{fig_global_explanation}
\end{figure}

\textbf{Explainer.} The \texttt{shapiq.Explainer} class is a simplified interface to explain any-order feature interactions in \gls{ML} models. \cref{fig_local_explanation} goes through exemplary code used to approximate \glspl{SI} for a single prediction and visualize them on a graph plot. Currently two classes are further distinguished within the API, but we envision extending \texttt{shapiq.Explainer} to include more data modalities and model algorithms. \texttt{shapiq.TabularExplainer} allows for model-agnostic explanation based on feature marginalization with either marginal or conditional imputation (refer to \cref{appx_imputers} for details). \texttt{shapiq.TreeExplainer} implements TreeSHAP-IQ~\citep{Muschalik.2024} for efficient explanations specific to decision tree-based models, e.g.\ random forest or gradient boosting decision trees, with native support for \texttt{scikit-learn}~\citep{pedregosa2011scikitlearn}, \texttt{xgboost}~\citep{chen2016xgboost}, and \texttt{lightgbm}~\citep{ke2017lightgbm}. \cref{fig_global_explanation} goes through exemplary code for explaining a set of predictions and visualizing their aggregation in a bar plot, which represents the global feature interaction importance.

\textbf{Utility functions.} \shapiq offers additional useful tools that are described in detail in the documentation. Interaction values are stored and processed using the \texttt{shapiq.InteractionValues} data class, which is rich in utility functions. The \texttt{shapiq.plot} module supports the visualization of interaction values, including our custom network plot, but also wrapping the well-known \texttt{force} and \texttt{bar} plots from \texttt{shap}~\citep{Lundberg.2017}. Finally, \texttt{shapiq.datasets} loads datasets used for testing and examples.

\section{Benchmarking Analysis}\label{sec_benchmark}

The \shapiq library enables computation of various \glspl{SI} for a broad class of application domains.
To illustrate its versatility, we conduct benchmarks across a wide variety of traditional \gls{ML}-based \gls{SV} application scenarios.
The \gls{ML} benchmark demonstrates how higher-order \glspl{SI} enable an accuracy--complexity trade-off for model interpretability (\cref{sec_experiments_acc_complexity}) and highlights that different approximation techniques in \shapiq achieve the state-of-the-art performance depending on the application domains~(\cref{sec_experiments_benchmark}).
\cref{tab_benchmark_overview,tab_app_benchamrk_configs_1,tab_app_benchamrk_configs_2} present an overview of different application domains and associated benchmarks.
Depending on the benchmark, it can be instantiated with different datasets, models, player numbers or benchmark-specific configuration parameters, e.g.\ uncertainty type: \emph{epistemic} for Uncertainty Explanation\ or imputer: \emph{conditional} for Local Explanation.
In total, \shapiq offers 100 unique benchmark games, i.e. applications times dataset--model pairs.

\begin{table}
\centering
\caption{Overview of the available benchmark games and domains in \protect{\shapiq}. Each benchmark can be instantiated with different datasets, models, player sizes, or benchmark-specific configuration parameters. This results in $2\,042$ pre-computed individual configurations (see  \cref{tab_app_benchamrk_configs_1,tab_app_benchamrk_configs_2})
.}
\vspace{0.5em}
\label{tab_benchmark_overview}
{\footnotesize
\begin{tabular}{@{}cllll@{}}
\toprule
\textbf{Domain} & \textbf{Benchmark (Game)} & \textbf{Source} & \textbf{Players} & \textbf{Coalition Worth} \\ \midrule
\multirow{3}{*}{\textbf{XAI}} & Local Explanation & \cite{Strumbelj.2014, Lundberg.2017} & Features & Model Output \\
& Global Explanation & \cite{Covert.2020} & Features & Model Loss \\
& Tree Explanation & \citep{Lundberg.2020,Muschalik.2024} & Features & Model Output \\[0.33em]
\textbf{Uncertainty} & Uncertainty Explanation & \citep{Watson.2023} & Features & Prediction Entropy \\[0.33em]
\multirow{3}{*}{\textbf{\begin{tabular}[c]{@{}c@{}}Model\\ Selection\end{tabular}}} & Feature Selection & \citep{Cohen.2007,Pfannschmidt.2016} & Features & Performance \\
& Ensemble Selection & \citep{Rozemberczki.2021} & Weak Learners & Performance\\
& RF Ensemble Selection & \citep{Rozemberczki.2021} & Tree Models & Performance \\[0.33em]
\multirow{2}{*}{\textbf{Valuation}} & Data Valuation & \citep{Ghorbani.2019} & Data Points & Performance \\
& Dataset Valuation & \citep{Tay.2022} & Data Subsets & Performance \\[0.33em]
\multirow{2}{*}{\textbf{\begin{tabular}[c]{@{}c@{}}Unsupervised \\ Learning\end{tabular}}} & Cluster Explanation & -- & Features & Cluster Score \\
& Unsupervised Feature Importance & \citep{Balestra.2022} & Features & Total Correlation \\ \midrule
\textbf{Synthetic} & Sum of Unanimity Model & \cite{Tsai.2022,Fumagalli.2024} & Players & Sum of Unan. Votes \\ \bottomrule
\end{tabular}
}
\end{table}

For all games that include $n \leq 16$ players, the value functions have been pre-computed by evaluating all coalitions and storing the games to file.
Reading a pre-computed game from file, instead of performing up to $2^{16} = 65\,536$ value function calls with each new experiment run, saves valuable computational time and contributes to reproducibility as well as sustainability.
This is particularly beneficial for tasks that involve remove-and-refit strategies~\cite{Covert.2021b}, such as Data Valuation or Feature Selection.
For $n > 16$, where pre-computing a game and ground truth values becomes computationally prohibitive, we rely on analytical solutions to compute the ground truth like TreeSHAP-IQ~\cite{Muschalik.2024} for tree-based ensembles or the \gls{MI} representation of Sum of Unanimity Models \cite{Tsai.2022,Fumagalli.2023,Fumagalli.2024}.
For details regarding the experimental setting and reproducibility, we refer to \cref{appx_sec_experiment_description}.

\subsection{Accuracy -- Complexity Trade-Off of Shapley Interactions for Interpretability}\label{sec_experiments_acc_complexity}
\begin{figure}[hb]
    \centering
    \begin{minipage}[c]{0.45\textwidth}
    \centering
        \textbf{(a) Single synthetic interaction}\\[0.5em]
        \includegraphics[width=\textwidth]{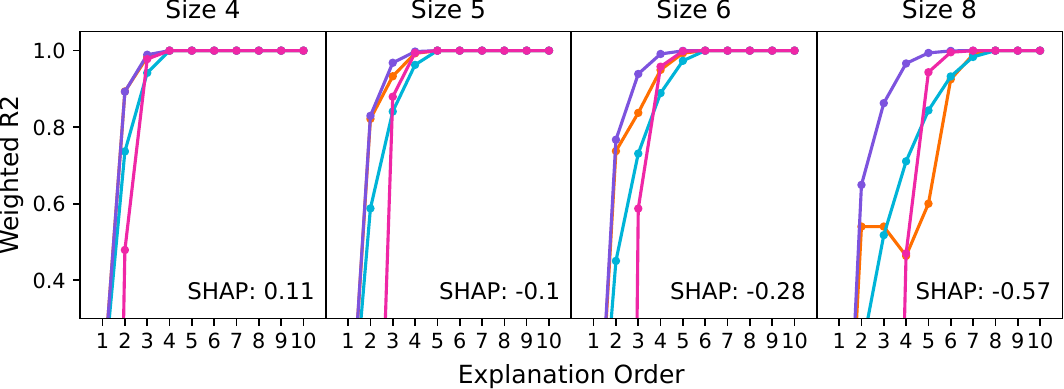}
    \end{minipage}
    \hfill
    \begin{minipage}[c]{0.45\textwidth}
        \centering
        \textbf{(b) \gls{ML} applications}\\[0.5em]
        \includegraphics[width=\textwidth]{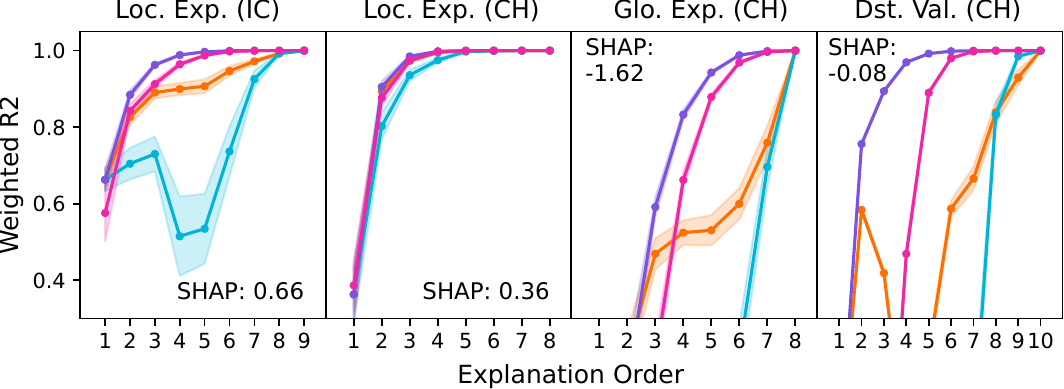}
    \end{minipage}
    \hfill
    \begin{minipage}[c]{0.07\textwidth}
        \vspace{1.1em}
        \includegraphics[width=\textwidth]{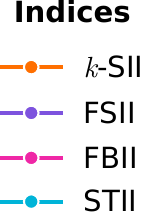}
    \end{minipage}
    \caption{Shapley-weighted $R^2$ of interaction indices by explanation order for \textbf{(a)} single synthetic interactions and  \textbf{(b)} \gls{ML} applications. \gls{FSII} is optimized for this metric, and increases faithfulness with each order. Interactions improve faithfulness over SHAP and yield an exact decomposition for the highest order. However, increasing interaction size negatively affects faithfulness.}
    \label{fig_complexity_accuracy}
\end{figure}

In this experiment, we empirically evaluate faithfulness of \glspl{SI} for varying explanation orders~$k$.
To this end, we rely on the Shapley-weighted faithfulness $\mathcal{L}(\nu,\Phi_k)$ introduced in \cref{sec_theoretical_background}.
\glspl{SI} range from the \gls{SV} (least complex) to the \gls{MI} (most complex) explanation, where the \gls{SV} minimizes $\mathcal{L}(\nu,\Phi_1)$ for $k=1$ \cite{Charnes.1988,Lundberg.2017}, and \glspl{MI} are faithful to all game values with $\mathcal{L}(\nu,\Phi_n)=0$ for $k=n$ \cite{Bord.2023,Tsai.2022}.
\cref{fig_complexity_accuracy} shows the Shapley-weighted $R^2$ value for $k=1,\dots,n$ for a synthetic game with a single interaction of varying size (a) and real-world \gls{ML} applications (b).
Here, we used $\mu_\infty = 1$ instead of $\mu_\infty \gg 1$, which affects \gls{FBII} that violates efficiency.
The results show that in general \glspl{SI} become more faithful with higher explanation order.
Notably, the difference between pairwise \glspl{SI} and \glspl{SV} (SHAP textbox) is remarkable, where pairwise interactions ($k=2$) already yield a strong improvement in faithfulness.
If higher-order interactions dominate, then \glspl{SI} require a larger explanation order to maintain faithfulness.
While \gls{FSII} and \gls{FBII} are optimized for faithfulness, \gls{STII} and \gls{k-SII} do not necessarily yield a strict improvement in this metric.
In fact, it was shown that \gls{SII} and \gls{k-SII} optimize a slightly different faithfulness metric, which changes for every order \cite{Fumagalli.2024}.
Yet, we observe a consistent strong improvement of pairwise \gls{k-SII} over the \gls{SV} (SHAP).
While \gls{FSII} and \gls{FBII} optimize faithfulness, \gls{k-SII} and \gls{STII} adhere to strict structural assumptions, where \gls{STII} projects all higher-order interactions to the top-order \glspl{SI}, and \gls{k-SII} is consistent with \gls{SII}.
Pracitioners may choose \glspl{SI} tailored to their specific application, where \gls{k-SII} is a good default choice for \shapiq.

\subsection{Comparison of Approximation Methods}\label{sec_experiments_benchmark}
\begin{figure}
    \centering
    \textbf{(a) Approximation quality in different application domains}
    \\
    \hspace{1.5em}
    \includegraphics[width=\textwidth]{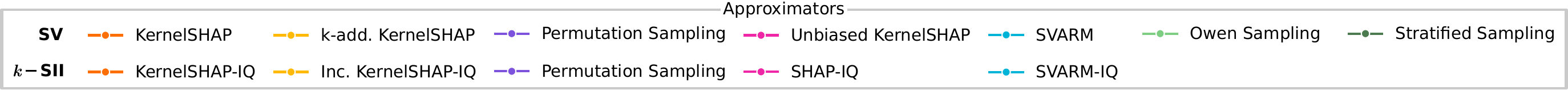}
    \\[0.5em]
    \begin{minipage}[c]{0.24\textwidth}
        \includegraphics[width=\textwidth]{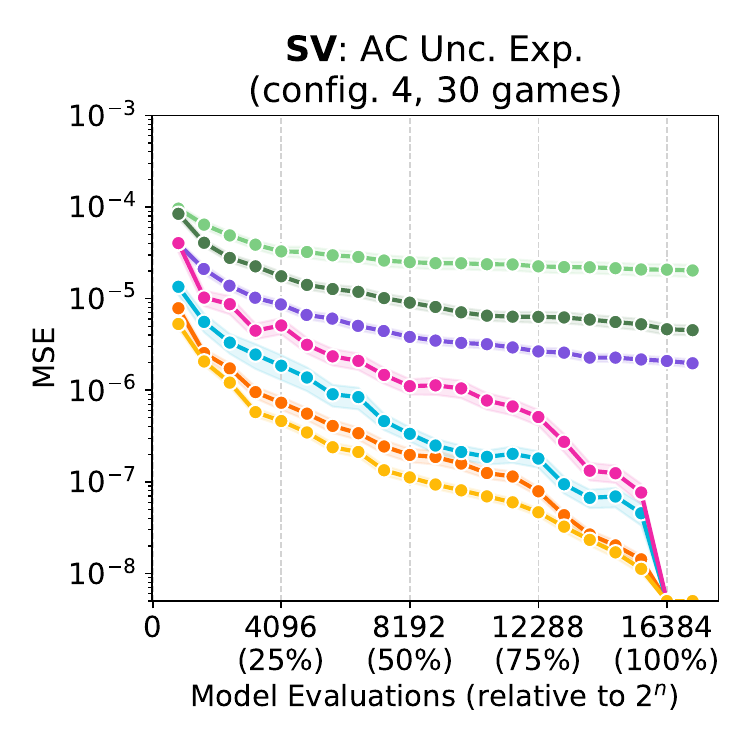}
    \end{minipage}
    \begin{minipage}[c]{0.24\textwidth}
        \includegraphics[width=\textwidth]{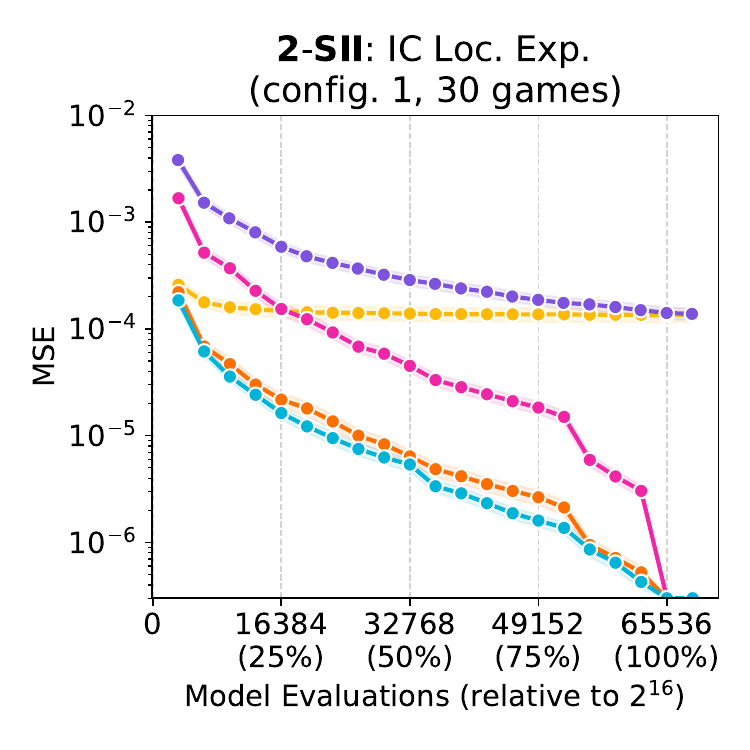}
    \end{minipage}
    \begin{minipage}[c]{0.24\textwidth}
        \includegraphics[width=\textwidth]{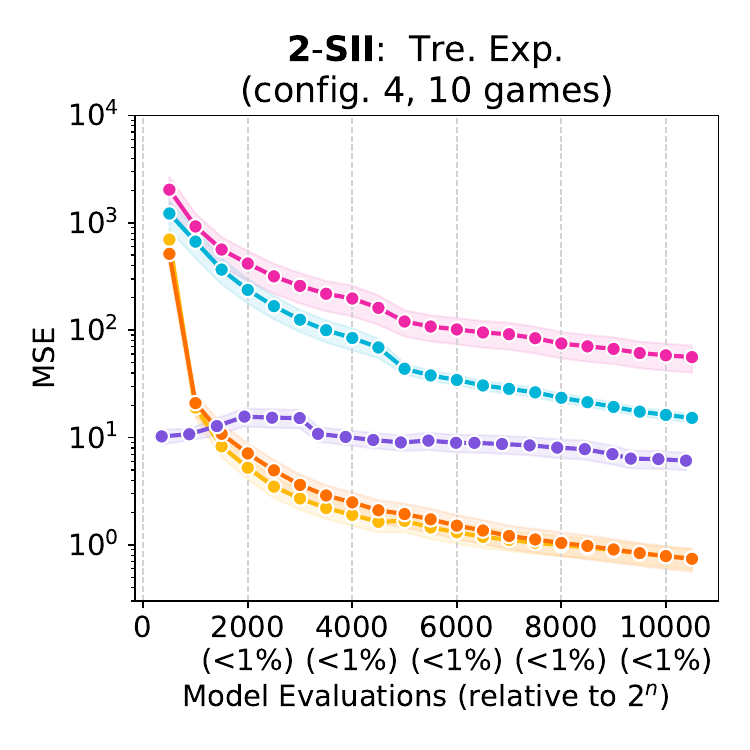}
    \end{minipage}
    \begin{minipage}[c]{0.24\textwidth}
        \includegraphics[width=\textwidth]{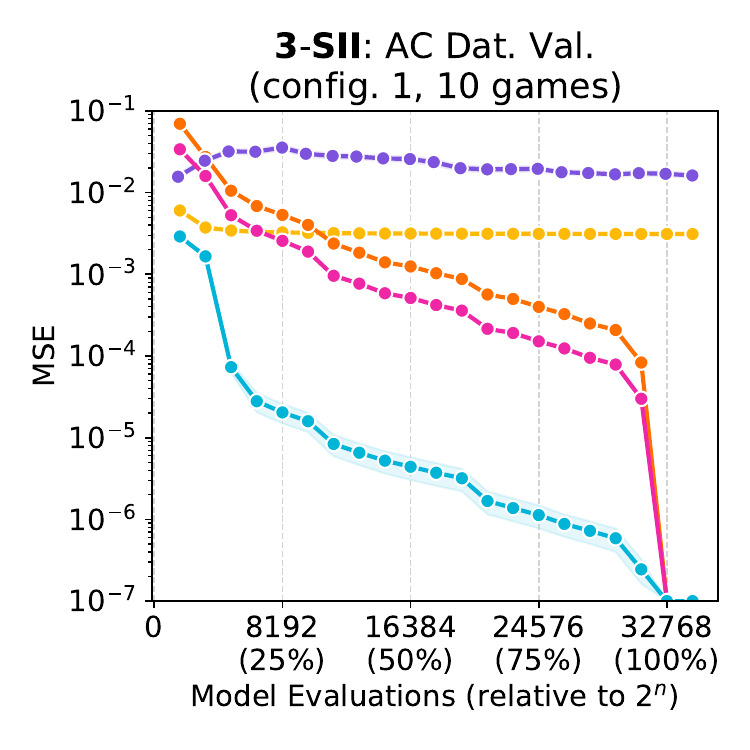}
    \end{minipage}
    \\
    \begin{minipage}[b]{0.75\textwidth}
        \centering
        \includegraphics[width=0.49\textwidth]{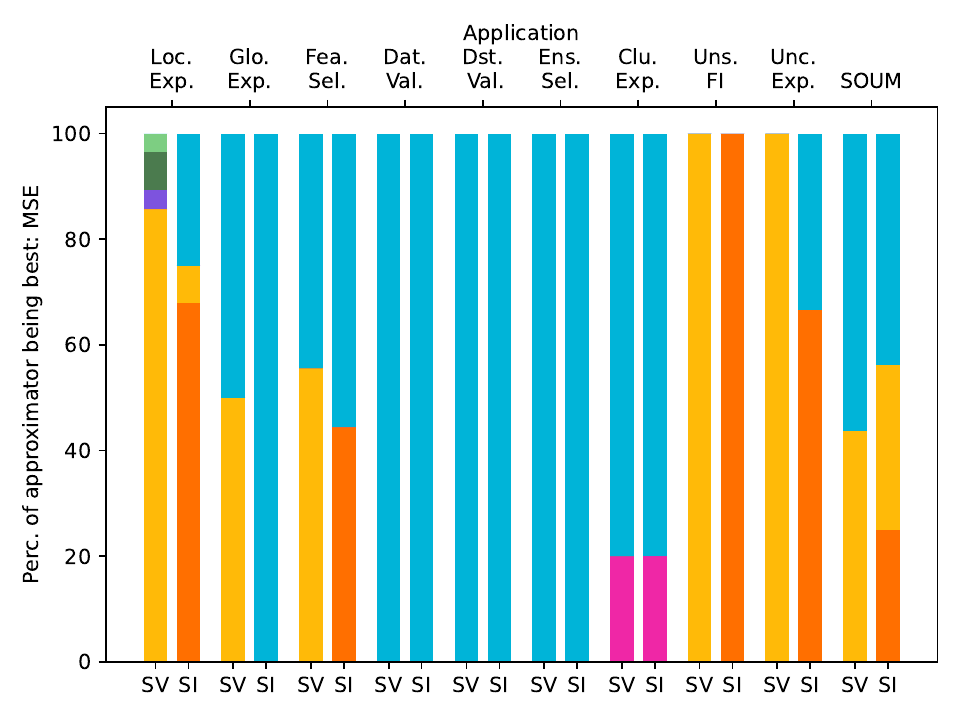}
        \includegraphics[width=0.49\textwidth]{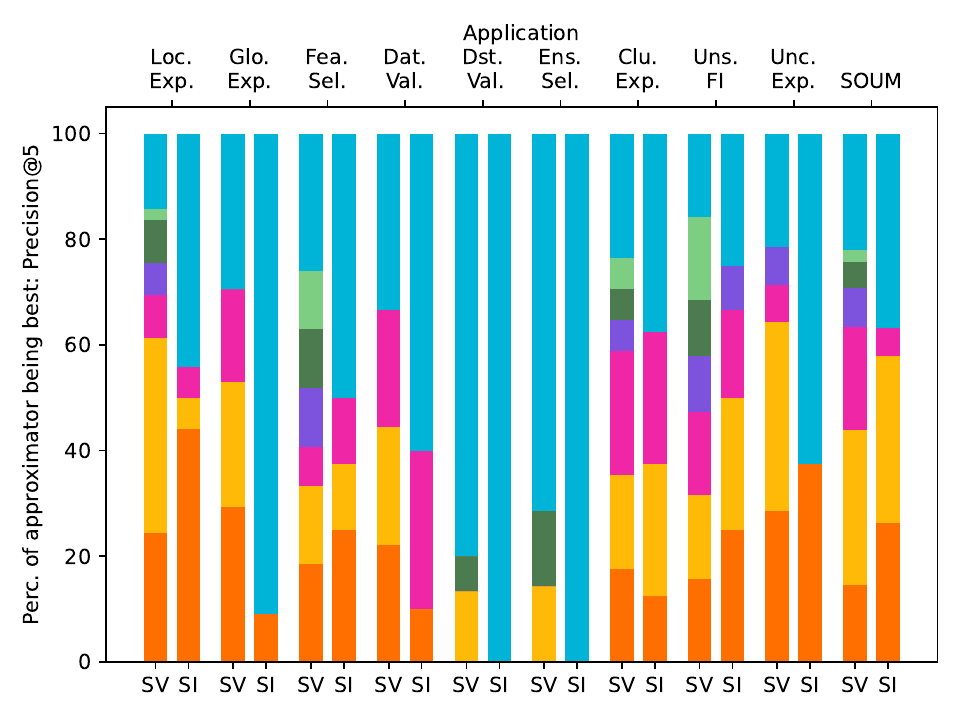}
        \\
        \textbf{(b) Overall approximator performance}
    \end{minipage}
    \begin{minipage}[b]{0.24\textwidth}
        \centering
        \includegraphics[width=\textwidth]{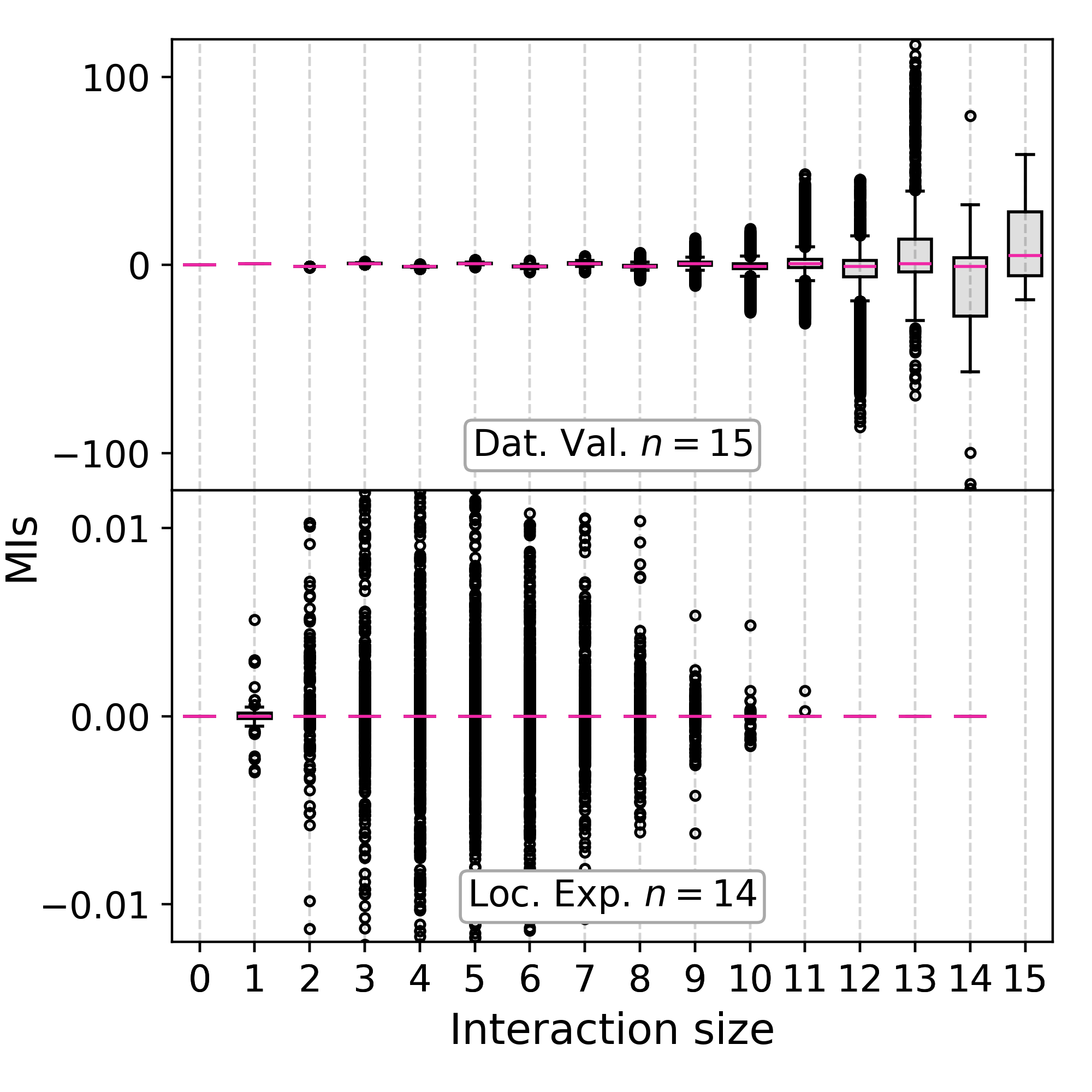}
        \textbf{(c) \glspl{MI} per size}
    \end{minipage}
    \caption{Overview of the benchmark results containing \textbf{(a)} budget-dependent \gls{MSE} approximation curves on different benchmark settings, \textbf{(b)} a summary of the best performing approximators per setting over all 100 benchmark games measured by \gls{MSE} \textbf{(left)} and \gls{prec} \textbf{(right)}, and \textbf{(c)}~exemplary \glspl{MI} for ten games of Data Valuation \textbf{(top)} and Local Explanation \textbf{(bottom)}.}
    \label{fig_approximation}
\end{figure}

Various approximation methods for computing \glspl{SI} are included in \shapiq for a variety of \glspl{SI} (cf. \cref{tab_method_overview}).
The possibility of attributing (domain-specific) state-of-the-art performance to a single algorithm has been investigated empirically by multiple works \cite{Fumagalli.2023,Fumagalli.2024,Kolpaczki.2024b,Kolpaczki.2024a,Tsai.2022,Maleki.2013,Owen.2014,Pelegrina.2023,vanCampen.2018}.
We use the collection of 100 unique benchmark games in \shapiq to evaluate the performance of different \gls{SV} and \gls{SI} approximation methods on a broad spectrum of \gls{ML} applications. 
For each domain and configuration (see \cref{tab_app_benchamrk_configs_1} and \ref{tab_app_benchamrk_configs_2} in \cref{appx_sec_overview_benchmark}), we compute ground truth \glspl{SV}, $2$-SIIs, and $3$-SIIs and compare them with estimates provided by all approximators from \cref{tab_method_overview}.
The approximators are run with a wide range of budget values and assessed by their achieved \gls{MSE} or \gls{prec}.
\cref{fig_approximation} summarizes the approximation results.

Most notably, the ranking of approximators varies strongly between the different applications domains, which is depicted in \cref{fig_approximation} (a) and (b). 
This observation holds for both \glspl{SV} and \glspl{SI}.
In general, two types of approximation methods dominate the application landscape in terms of \gls{MSE} and \gls{prec}.
First, \textit{kernel-based} approaches including KernelSHAP, $k_\text{ADD}$-SHAP, KernelSHAP-IQ and Inconsistent KernelSHAP-IQ perform best for Local Explanation, Uncertainty Explanation, and Unsupervised Feature Importance.
Second, the two \textit{stratification-based} estimators SVARM and SVARM-IQ achieve state-of-the-art performance for Data Valuation, Dataset Valuation, or Ensemble Selection.
Traditional \textit{mean-estimation} methods including Permutation Sampling (SV and SII), Unbiased KernelSHAP, SHAPIQ, and Owen Sampling achieve moderate estimation qualities in comparison.
Our findings give rise to the conclusion that
\textit{stratification-based} estimators perform superior in settings where the size of a coalition naturally impacts its worth (e.g.\ training size for Dataset Valuation), which is plausible as these methods group coalitions by size and thus leverage this dependency.
Meanwhile, \textit{kernel-based} estimators achieve state-of-the-art in settings where the dependency between size and worth of a coalition is less pronounced (e.g.\ sudden jumps of model predictions in Local Explanation).

Interestingly, the settings where \textit{stratified-sampling} outperforms \textit{kernel-based} variants exhibit different internal structures in the games' \glspl{MI}.
Generally, \glspl{MI} disentangle a game into all of its additive components (cf.\ \cref{sec_theoretical_background}) and can be computed exactly with \shapiq's pre-computed games.
The accuracy of \textit{kernel-based} estimators drops when higher-order interactions dominate the games' structure instead of lower-order interactions.
This is depicted by \cref{fig_approximation} (c) where the \glspl{MI} for Local Explanation are of lower order than the Data Valuation games.

\section{Conclusion}\label{sec_conclusion}

As \glspl{SI} are increasingly employed to analyze \gls{ML} models, it becomes pivotal to ensure that these are accurately and efficiently approximated. To this end, we contributed \shapiq, an open-source toolbox that implements state-of-the-art algorithms, defines a dozen of benchmark games, and provides ready-to-use explanations of any-order feature interactions. \shapiq comes with a comprehensive documentation and is designed to be extendable by contributors. 

\textbf{Limitations and future work.} We identify three main limitations of \shapiq that provide natural opportunities for future work. First, the TreeSHAP-IQ algorithm is currently implemented in Python, but by-design requires no access to model inference, which allows for a more efficient implementation in C++ alike TreeSHAP~\citep{Lundberg.2020,Yu.2022}. Second, \glspl{SI} can be misinterpreted based on choosing the wrong index for the application scenario, which we comment on across \cref{sec_theoretical_background,sec_experiments_acc_complexity}.
The selection of a particular \gls{SI} index, enabled by \shapiq, offers great opportunities for application-specific research.
We also acknowledge that visualization of higher-order feature interactions is itself challenging and a potential research direction in human-computer interaction. 
Certainly, a human-centric evaluation of explanations may be required for their broader adoption in practical applications~\citep{rong2024towards}.

\textbf{Broader impact.} A potential negative societal implication of visualizing higher-order feature interactions may be an \emph{information overload}~\citep{poursabzisangdeh2021manipulating,baniecki2024grammar} that leads to users misinterpreting model explanations. Nevertheless, we hope our contribution sparks the advancement of game-theoretical indices motivated by various applications in \gls{ML}. Specifically in the context of explainability, \shapiq may impact the way users interact with \gls{ML} models when having access to previously inaccessible information, e.g. higher-order feature interactions.

\begin{ack}
We gratefully thank the anonymous reviewer for their valuable feedback for improving this work.
Fabian Fumagalli and Maximilian Muschalik gratefully acknowledge funding by the Deutsche Forschungsgemeinschaft (DFG, German Research Foundation): TRR 318/1 2021 – 438445824. Hubert Baniecki was supported from the state budget within the Polish Ministry of Education and Science program ``Pearls of Science'' project number PN/01/0087/2022. Patrick Kolpaczki was supported by the research training group Dataninja (Trustworthy AI for Seamless Problem Solving: Next Generation Intelligence Joins Robust Data Analysis) funded by the German federal state of North Rhine-Westphalia.
\end{ack}

\bibliographystyle{apalike}
\bibliography{references}

\clearpage
\appendix
\onecolumn
\section*{Organization of the Supplementary Material}
The supplementary material is organized as follows:

\startcontents[sections]
\printcontents[sections]{l}{1}{\setcounter{tocdepth}{3}}

\clearpage
\section{Extended Theoretical Background}
In this section, we introduce further theoretical background.
Specifically, we discuss in more detail the class of \glspl{GV} and \glspl{II} in \cref{appx_sec_gv_ii}, and the \glspl{MI} in \cref{appx_sec_mi}.

\subsection{Probabilistic and Cardinal-Probabilistic GVs and IIs} \label{appx_sec_gv_ii}
Probabilistic \glspl{GV} \cite{DBLP:journals/dam/MarichalKF07} extend semivalues with a focus on \emph{monotonicity}, i.e.\ games that satisfy $\nu(S) \leq \nu(T)$, if $S \subseteq T \subseteq N$.
\glspl{GV} satisfy the \emph{positivity axiom}, which requires non-negative joint contributions, i.e.\ $\phi^{\text{GV}}(S) \geq 0$, for all $S \subseteq N$ in monotone games \cite{DBLP:journals/dam/MarichalKF07}.
It was shown that \glspl{GV} are uniquely represented as weighted averages over (joint) marginal contributions $\nu(T \cup S) - \nu(T)$.
On the other hand, cardinal-probabilistic \glspl{II} \cite{Fujimoto.2006} are centered around synergy, independence and redundancy between entities.
\glspl{II} are based on discrete derivatives, which extend (joint) marginal contributions by accounting for lower-order effects.
\glspl{II} focus on \emph{$k$-monotonicity}, i.e.\ games that have non-negative discrete derivatives $\Delta_S(T) \geq 0$ for $S \subseteq U \subseteq N$ with $2\leq \vert S \vert \leq k$.
\glspl{II} satisfy the \emph{$k$-monotonicity axiom}, i.e.\ non-negative interactions $\phi^{\text{II}}(S) \geq 0$ for $k$-monotone games.
Both, \glspl{GV} and \glspl{II} are uniquely represented \cite{DBLP:journals/dam/MarichalKF07,Fujimoto.2006} as
\begin{align*}    
    &\phi^{\text{GV}}(S) := \sum_{T \subseteq N \setminus S} p_{\vert T \vert}^{\vert S \vert}(n) \cdot \left(\nu(T \cup S) - \nu(T)\right)  &&\text{ and }   &&\phi^{\text{II}}(S) := \sum_{T \subseteq N \setminus S} p_{\vert T \vert}^{\vert S \vert}(n) \cdot \Delta_S(T),
\end{align*}
where $p_t^s(n)$ are index-specific weights based on the sizes of $S,T$ and $N$.
The \emph{\gls{SGV}}~\cite{Marichal_2000} and the \emph{\gls{SII}} \cite{Grabisch.1999} with 
\begin{align*}
    \textbf{Shapley: } p_t^s(n) = \frac{1}{n-s+1} \binom{n-s}{t}^{-1}
\end{align*}
naturally extend the \gls{SV}.
An alternative extension for the \gls{SV} is the \emph{\gls{CHGV}} \cite{DBLP:journals/dam/MarichalKF07} and \emph{\gls{CHII}} \cite{Marichal.1999} with
\begin{align*}
    \textbf{Chaining: } p_t^s(n) = \frac{s}{s + t} \binom{n}{s + t}^{-1}.
\end{align*}
The main difference of the \gls{SGV}/\gls{SII} and the \gls{CHGV}/\gls{CHII} is the quantification of so-called
\emph{partnerships} \cite{DBLP:journals/dam/MarichalKF07,Fujimoto.2006}, i.e. coalitions that only influence the value of the game if all members of the partnership are present.
The \gls{CHGV} and \gls{CHII} adhere to the \emph{partnership-allocation axiom} \cite{Fujimoto.2006,DBLP:journals/dam/MarichalKF07}, which states that the contribution of an individual member of the partnership and the interaction of the whole partnership are proportional.
In contrast, the \gls{SGV} and \gls{SII} satisfy the \emph{reduced partnership consistency axiom} \cite{Fujimoto.2006,DBLP:journals/dam/MarichalKF07}, which states that the interaction of the whole partnership is equal to the contribution of the partnership in a game, where the partnership is a single player.

On the other hand, the \emph{\gls{BGV}} \cite{Marichal_2000} and \emph{\gls{BII}} \cite{Grabisch.1999} extend the \gls{BV} with 
\begin{align*}
    \textbf{Banzhaf: } p_t^s(n) := \frac{1}{2^{n-s}}.
\end{align*}

\subsection{Möbius Interactions (MIs)}\label{appx_sec_mi}

The \glspl{MI} $\Phi_n$ are a prominent concept in discrete mathematics, which appears in many different forms.
In discrete mathematics, it also known as the Möbius transform \cite{rota1964foundations}.
In cooperative game theory, the concept is known as Harsanyi dividend \cite{harsanyi1963simplified} or internal \gls{II} \cite{Fujimoto.2006}.
The \gls{MI} for $S \subseteq N$ is defined as
\begin{equation*}
    \Phi_n(S) := \sum_{T \subseteq S} (-1)^{\vert S\vert - \vert T \vert} \nu(T).
\end{equation*}
In this context, the \glspl{MI} are the unique measure that satisfy the \emph{recovery property}
\begin{align*}
    \nu(T) = \sum_{S \subseteq T}\Phi_n(S) \text{ for every } T\subseteq N.
\end{align*}
The \glspl{MI} are a basis of the vector space of cooperative games, and thus every game can be uniquely represented in terms of its \glspl{MI}.
The \emph{Co-M\"obius transform (Co-MI)} \cite{Grabisch.2000} is another fundamental concept linked to the \glspl{MI} of the conjugate game, i.e.\ $\bar\nu(T) := \nu(N\setminus T)$ \cite{Grabisch.2016}.

\clearpage
\section{Detailed Overview of all Benchmark Games and Configurations}\label{appx_sec_overview_benchmark}

\subsection{Benchmark Overview}

We list in \cref{tab_app_benchamrk_configs_1} and \ref{tab_app_benchamrk_configs_2} all configurations available within our benchmark.
Depending on the application task, a configuration represents a combination of multiple parameters which specify the generated cooperative games.
For \gls{ML} games such a combination includes at least the used dataset and number of features or datapaoints.
If a prediction model or imputer for feature values is employed, as for example for Local XAI games, these are also specified.

\begin{table}[!htb]
\caption{Overview of all Benchmark Configurations: Each configuration is assigned a distinctive identifier (ID), has a name (Benchmark) indicating dataset and application if available, is pre-computed (P.) if the player number $n$ does not exceed 16, compromises $|\mathcal{P}(N)|$ many coalitions to be evaluated, is iterated over multiple game instances ($g$), and has a set of parameters (Game Configuration).}
\vspace{0.5em}
\resizebox{\columnwidth}{!}{%
\begin{tabular}{cccccccc}
\toprule
\textbf{ID} & \textbf{Benchmark} & \textbf{Data} & \textbf{P.} & \textbf{$n$} & \textbf{$|\mathcal{P}(N)|$} & \textbf{$g$} & \textbf{Game Configuration} \\\midrule
\textbf{1} & Data Valuation & AD & \cmark & 15 & 32768 & 10 & model\_name=decision\_tree, n\_data\_points=15 \\
\textbf{2} & Data Valuation & AD & \cmark & 15 & 32768 & 10 & model\_name=random\_forest, n\_data\_points=15 \\
\midrule
\textbf{3} & Data Valuation & AD & \cmark & 10 & 1024 & 30 & model\_name=decision\_tree, player\_sizes=increasing, n\_players=10 \\
\textbf{4} & Data Valuation & AD & \cmark & 10 & 1024 & 30 & model\_name=random\_forest, player\_sizes=increasing, n\_players=10 \\
\textbf{5} & Data Valuation & AD & \cmark & 10 & 1024 & 30 & model\_name=gradient\_boosting, player\_sizes=increasing, n\_players=10 \\
\textbf{6} & Data Valuation & AD & \cmark & 14 & 16384 & 5 & model\_name=decision\_tree, player\_sizes=increasing, n\_players=14 \\
\midrule
\textbf{7} & Ensemble Selection & AD & \cmark & 10 & 1024 & 30 & loss\_function=accuracy\_score, n\_members=10 \\
\midrule
\textbf{8} & Feature Selection & AD & \cmark & 14 & 16384 & 30 & model\_name=decision\_tree \\
\textbf{9} & Feature Selection & AD & \cmark & 14 & 16384 & 30 & model\_name=random\_forest \\
\textbf{10} & Feature Selection & AD & \cmark & 14 & 16384 & 30 & model\_name=gradient\_boosting \\
\midrule
\textbf{11} & Global Explanation & AD & \cmark & 14 & 16384 & 30 & model\_name=decision\_tree, loss\_function=accuracy\_score \\
\textbf{12} & Global Explanation & AD & \cmark & 14 & 16384 & 30 & model\_name=random\_forest, loss\_function=accuracy\_score \\
\textbf{13} & Global Explanation & AD & \cmark & 14 & 16384 & 30 & model\_name=gradient\_boosting, loss\_function=accuracy\_score \\
\midrule
\textbf{14} & Local Explanation & AD & \cmark & 14 & 16384 & 30 & model\_name=decision\_tree, imputer=marginal \\
\textbf{15} & Local Explanation & AD & \cmark & 14 & 16384 & 30 & model\_name=random\_forest, imputer=marginal \\
\textbf{16} & Local Explanation & AD & \cmark & 14 & 16384 & 30 & model\_name=gradient\_boosting, imputer=marginal \\
\textbf{17} & Local Explanation & AD & \cmark & 14 & 16384 & 30 & model\_name=decision\_tree, imputer=conditional \\
\textbf{18} & Local Explanation & AD & \cmark & 14 & 16384 & 30 & model\_name=random\_forest, imputer=conditional \\
\textbf{19} & Local Explanation & AD & \cmark & 14 & 16384 & 30 & model\_name=gradient\_boosting, imputer=conditional \\
\midrule
\textbf{20} & RF Ensemble Selection & AD & \cmark & 10 & 1024 & 30 & loss\_function=accuracy\_score, n\_members=10 \\
\midrule
\textbf{21} & Uncertainty Explanation & AD & \cmark & 14 & 16384 & 30 & uncertainty\_to\_explain=total, imputer=marginal \\
\textbf{22} & Uncertainty Explanation & AD & \cmark & 14 & 16384 & 30 & uncertainty\_to\_explain=total, imputer=conditional \\
\textbf{23} & Uncertainty Explanation & AD & \cmark & 14 & 16384 & 30 & uncertainty\_to\_explain=aleatoric, imputer=marginal \\
\textbf{24} & Uncertainty Explanation & AD & \cmark & 14 & 16384 & 30 & uncertainty\_to\_explain=aleatoric, imputer=conditional \\
\textbf{25} & Uncertainty Explanation & AD & \cmark & 14 & 16384 & 30 & uncertainty\_to\_explain=epistemic, imputer=marginal \\
\textbf{26} & Uncertainty Explanation & AD & \cmark & 14 & 16384 & 30 & uncertainty\_to\_explain=epistemic, imputer=conditional \\
\midrule
\textbf{27} & Unsupervised FI. & AD & \cmark & 14 & 16384 & 1 & - \\
\midrule
\textbf{28} & Cluster Explanation & BS & \cmark & 12 & 4096 & 1 & cluster\_method=kmeans, score\_method=silhouette\_score \\
\textbf{29} & Cluster Explanation & BS & \cmark & 12 & 4096 & 1 & cluster\_method=agglomerative, score\_method=calinski\_harabasz\_score \\
\midrule
\textbf{30} & Data Valuation & BS & \cmark & 15 & 32768 & 10 & model\_name=decision\_tree, n\_data\_points=15 \\
\textbf{31} & Data Valuation & BS & \cmark & 15 & 32768 & 10 & model\_name=random\_forest, n\_data\_points=15 \\
\midrule
\textbf{32} & Data Valuation & BS & \cmark & 10 & 1024 & 30 & model\_name=decision\_tree, player\_sizes=increasing, n\_players=10 \\
\textbf{33} & Data Valuation & BS & \cmark & 10 & 1024 & 30 & model\_name=random\_forest, player\_sizes=increasing, n\_players=10 \\
\textbf{34} & Data Valuation & BS & \cmark & 10 & 1024 & 30 & model\_name=gradient\_boosting, player\_sizes=increasing, n\_players=10 \\
\textbf{35} & Data Valuation & BS & \cmark & 14 & 16384 & 5 & model\_name=decision\_tree, player\_sizes=increasing, n\_players=14 \\
\midrule
\textbf{36} & Ensemble Selection & BS & \cmark & 10 & 1024 & 30 & loss\_function=r2\_score, n\_members=10 \\
\midrule
\textbf{37} & Feature Selection & BS & \cmark & 12 & 4096 & 30 & model\_name=decision\_tree \\
\textbf{38} & Feature Selection & BS & \cmark & 12 & 4096 & 30 & model\_name=random\_forest \\
\textbf{39} & Feature Selection & BS & \cmark & 12 & 4096 & 30 & model\_name=gradient\_boosting \\
\midrule
\textbf{40} & Global Explanation & BS & \cmark & 12 & 4096 & 30 & model\_name=decision\_tree, loss\_function=r2\_score \\
\textbf{41} & Global Explanation & BS & \cmark & 12 & 4096 & 30 & model\_name=random\_forest, loss\_function=r2\_score \\
\textbf{42} & Global Explanation & BS & \cmark & 12 & 4096 & 30 & model\_name=gradient\_boosting, loss\_function=r2\_score \\
\midrule
\textbf{43} & Local Explanation & BS & \cmark & 12 & 4096 & 30 & model\_name=decision\_tree, imputer=marginal \\
\textbf{44} & Local Explanation & BS & \cmark & 12 & 4096 & 30 & model\_name=random\_forest, imputer=marginal \\
\textbf{45} & Local Explanation & BS & \cmark & 12 & 4096 & 30 & model\_name=gradient\_boosting, imputer=marginal \\
\textbf{46} & Local Explanation & BS & \cmark & 12 & 4096 & 30 & model\_name=decision\_tree, imputer=conditional \\
\textbf{47} & Local Explanation & BS & \cmark & 12 & 4096 & 30 & model\_name=random\_forest, imputer=conditional \\
\textbf{48} & Local Explanation & BS & \cmark & 12 & 4096 & 30 & model\_name=gradient\_boosting, imputer=conditional \\
\midrule
\textbf{49} & RF Ensemble Selection & BS & \cmark & 10 & 1024 & 30 & loss\_function=r2\_score, n\_members=10 \\
\midrule
\textbf{50} & Unsupervised FI. & BS & \cmark & 12 & 4096 & 1 & -- \\
\bottomrule
\end{tabular}
}
\label{tab_app_benchamrk_configs_1}
\end{table}

\begin{table}[!htb]
\caption{Overview of all Benchmark Configurations: Each configuration is assigned a distinctive identifier (ID), has a name (Benchmark) indicating dataset and application if available, is pre-computed (P.) if the player number $n$ does not exceed 16, compromises $|\mathcal{P}(N)|$ many coalitions to be evaluated, is iterated over multiple game instances ($g$), and has a set of parameters (Game Configuration).}
\vspace{0.5em}
\resizebox{\columnwidth}{!}{%
\begin{tabular}{cccccccc}
\toprule
\textbf{ID} & \textbf{Benchmark} & \textbf{Data} & \textbf{P.} & \textbf{$n$} & \textbf{$|\mathcal{P}(N)|$} & \textbf{$g$} & \textbf{Game Configuration} \\\midrule
\textbf{51} & Cluster Explanation & CH & \cmark & 8 & 256 & 1 & cluster\_method=kmeans, score\_method=silhouette\_score \\
\textbf{52} & Cluster Explanation & CH & \cmark & 8 & 256 & 1 & cluster\_method=agglomerative, score\_method=calinski\_harabasz\_score \\
\midrule
\textbf{53} & Data Valuation & CH & \cmark & 15 & 32768 & 10 & model\_name=decision\_tree, n\_data\_points=15 \\
\textbf{54} & Data Valuation & CH & \cmark & 15 & 32768 & 10 & model\_name=random\_forest, n\_data\_points=15 \\
\midrule
\textbf{55} & Data Valuation & CH & \cmark & 10 & 1024 & 30 & model\_name=decision\_tree, player\_sizes=increasing, n\_players=10 \\
\textbf{56} & Data Valuation & CH & \cmark & 10 & 1024 & 30 & model\_name=random\_forest, player\_sizes=increasing, n\_players=10 \\
\textbf{57} & Data Valuation & CH & \cmark & 10 & 1024 & 30 & model\_name=gradient\_boosting, player\_sizes=increasing, n\_players=10 \\
\textbf{58} & Data Valuation & CH & \cmark & 14 & 16384 & 5 & model\_name=decision\_tree, player\_sizes=increasing, n\_players=14 \\
\midrule
\textbf{59} & Ensemble Selection & CH & \cmark & 10 & 1024 & 30 & loss\_function=r2\_score, n\_members=10 \\
\midrule
\textbf{60} & Feature Selection & CH & \cmark & 8 & 256 & 30 & model\_name=decision\_tree \\
\textbf{61} & Feature Selection & CH & \cmark & 8 & 256 & 30 & model\_name=random\_forest \\
\textbf{62} & Feature Selection & CH & \cmark & 8 & 256 & 30 & model\_name=gradient\_boosting \\
\midrule
\textbf{63} & Global Explanation & CH & \cmark & 8 & 256 & 30 & model\_name=decision\_tree, loss\_function=r2\_score \\
\textbf{64} & Global Explanation & CH & \cmark & 8 & 256 & 30 & model\_name=random\_forest, loss\_function=r2\_score \\
\textbf{65} & Global Explanation & CH & \cmark & 8 & 256 & 30 & model\_name=gradient\_boosting, loss\_function=r2\_score \\
\textbf{66} & Global Explanation & CH & \cmark & 8 & 256 & 30 & model\_name=neural\_network, loss\_function=r2\_score \\
\midrule
\textbf{67} & Local Explanation & CH & \cmark & 8 & 256 & 30 & model\_name=decision\_tree, imputer=marginal \\
\textbf{68} & Local Explanation & CH & \cmark & 8 & 256 & 30 & model\_name=random\_forest, imputer=marginal \\
\textbf{69} & Local Explanation & CH & \cmark & 8 & 256 & 30 & model\_name=gradient\_boosting, imputer=marginal \\
\textbf{70} & Local Explanation & CH & \cmark & 8 & 256 & 30 & model\_name=neural\_network, imputer=marginal \\
\textbf{71} & Local Explanation & CH & \cmark & 8 & 256 & 30 & model\_name=decision\_tree, imputer=conditional \\
\textbf{72} & Local Explanation & CH & \cmark & 8 & 256 & 30 & model\_name=random\_forest, imputer=conditional \\
\textbf{73} & Local Explanation & CH & \cmark & 8 & 256 & 30 & model\_name=gradient\_boosting, imputer=conditional \\
\textbf{74} & Local Explanation & CH & \cmark & 8 & 256 & 30 & model\_name=neural\_network, imputer=conditional \\
\midrule
\textbf{75} & RF Ensemble Selection & CH & \cmark & 10 & 1024 & 30 & loss\_function=r2\_score, n\_members=10 \\
\midrule
\textbf{76} & Unsupervised FI. & CH & \cmark & 8 & 256 & 1 & -- \\
\midrule
\textbf{77} & Local Explanation & IC & \cmark & 14 & 16384 & 30 & model\_name=resnet\_18, n\_superpixel\_resnet=14 \\
\textbf{78} & Local Explanation & IC & \cmark & 9 & 512 & 30 & model\_name=vit\_9\_patches \\
\textbf{79} & Local Explanation & IC & \cmark & 16 & 65536 & 30 & model\_name=vit\_16\_patches \\
\midrule
\textbf{80} & Sum of Unanimity Model & Syn & \cmark & 15 & 32768 & 10 & n=15, n\_basis\_games=30, min\_interaction\_size=1, max\_interaction\_size=5 \\
\textbf{81} & Sum of Unanimity Model & Syn & \cmark & 15 & 32768 & 10 & n=15, n\_basis\_games=30, min\_interaction\_size=1, max\_interaction\_size=15 \\
\textbf{82} & Sum of Unanimity Model & Syn & \cmark & 15 & 32768 & 10 & n=15, n\_basis\_games=150, min\_interaction\_size=1, max\_interaction\_size=5 \\
\textbf{83} & Sum of Unanimity Model & Syn & \cmark & 15 & 32768 & 10 & n=15, n\_basis\_games=150, min\_interaction\_size=1, max\_interaction\_size=15 \\
\textbf{84} & Sum of Unanimity Model & Syn & \textbf{X} & 30 & $>2^{16}$ & 10 & n=30, n\_basis\_games=30, min\_interaction\_size=1, max\_interaction\_size=5 \\
\textbf{85} & Sum of Unanimity Model & Syn & \textbf{X} & 30 & $>2^{16}$ & 10 & n=30, n\_basis\_games=30, min\_interaction\_size=1, max\_interaction\_size=15 \\
\textbf{86} & Sum of Unanimity Model & Syn & \textbf{X} & 30 & $>2^{16}$ & 10 & n=30, n\_basis\_games=30, min\_interaction\_size=1, max\_interaction\_size=25 \\
\textbf{87} & Sum of Unanimity Model & Syn & \textbf{X} & 30 & $>2^{16}$ & 10 & n=30, n\_basis\_games=150, min\_interaction\_size=1, max\_interaction\_size=5 \\
\textbf{88} & Sum of Unanimity Model & Syn & \textbf{X} & 30 & $>2^{16}$ & 10 & n=30, n\_basis\_games=150, min\_interaction\_size=1, max\_interaction\_size=15 \\
\textbf{89} & Sum of Unanimity Model & Syn & \textbf{X} & 30 & $>2^{16}$ & 10 & n=30, n\_basis\_games=150, min\_interaction\_size=1, max\_interaction\_size=25 \\
\textbf{90} & Sum of Unanimity Model & Syn & \textbf{X} & 50 & $>2^{16}$ & 10 & n=50, n\_basis\_games=30, min\_interaction\_size=1, max\_interaction\_size=5 \\
\textbf{91} & Sum of Unanimity Model & Syn & \textbf{X} & 50 & $>2^{16}$ & 10 & n=50, n\_basis\_games=30, min\_interaction\_size=1, max\_interaction\_size=15 \\
\textbf{92} & Sum of Unanimity Model & Syn & \textbf{X} & 50 & $>2^{16}$ & 10 & n=50, n\_basis\_games=30, min\_interaction\_size=1, max\_interaction\_size=25 \\
\textbf{93} & Sum of Unanimity Model & Syn & \textbf{X} & 50 & $>2^{16}$ & 10 & n=50, n\_basis\_games=150, min\_interaction\_size=1, max\_interaction\_size=5 \\
\textbf{94} & Sum of Unanimity Model & Syn & \textbf{X} & 50 & $>2^{16}$ & 10 & n=50, n\_basis\_games=150, min\_interaction\_size=1, max\_interaction\_size=15 \\
\textbf{95} & Sum of Unanimity Model & Syn & \textbf{X} & 50 & $>2^{16}$ & 10 & n=50, n\_basis\_games=150, min\_interaction\_size=1, max\_interaction\_size=25 \\
\midrule
\textbf{96} & Local Explanation & MR & \cmark & 14 & 16384 & 30 & mask\_strategy=mask \\
\midrule
\textbf{97} & Tree Explanation & Syn & \textbf{X} & 30 & $>2^{16}$ & 10 & model\_name=decision\_tree, classification=True, n\_features=30 \\
\textbf{98} & Tree Explanation & Syn & \textbf{X} & 30 & $>2^{16}$ & 10 & model\_name=random\_forest, classification=True, n\_features=30 \\
\textbf{99} & Tree Explanation & Syn & \textbf{X} & 30 & $>2^{16}$ & 10 & model\_name=decision\_tree, classification=False, n\_features=30 \\
\textbf{100} & Tree Explanation & Syn & \textbf{X} & 30 & $>2^{16}$ & 10 & model\_name=random\_forest, classification=False, n\_features=30 \\
\bottomrule
\end{tabular}
}
\label{tab_app_benchamrk_configs_2}
\end{table}

\clearpage
\subsection{Datasets and Models Used}\label{appx_sec_datasets_used}
Our benchmark games are based on five datasets.
All of these datasets are publicly available.
The following contains a small description of all datasets: 
\begin{itemize}
    \item[\textbf{AC:}] The \textit{AdultCensus} \cite[][CC BY 4.0 license]{Kohavi.1996} dataset is a tabular classification dataset containing $n=14$ features. The dataset was obtained from \texttt{openml} \cite{Feurer.2020} (id: \emph{1590}) and is available at \url{https://github.com/mmschlk/shapiq/blob/v1/data/adult_census.csv} for reproducibility.
    \item[\textbf{BS:}] The \textit{BikeSharing} \cite[][CC BY 4.0 license]{FanaeeT.2014} dataset is a tabular regression dataset containing $n=12$ features. The dataset was obtained from \texttt{openml} \cite{Feurer.2020} (id: \emph{42712}) and is available at \url{https://github.com/mmschlk/shapiq/blob/v1/data/bike.csv} for reproducibility.
    \item[\textbf{CH:}] The \textit{CaliforniaHousing}~\cite[][CC0 public domain]{Kelley.1997} dataset is a tabular regression dataset containing $n=8$ features. The target of this dataset is to predict property prices. The dataset was obtained from \texttt{scikit-learn}~\cite{pedregosa2011scikitlearn} and is available at \url{https://github.com/mmschlk/shapiq/blob/v1/data/california_housing.csv} for reproducibility.
    \item[\textbf{MR:}] The \textit{MovieReview} is also known as the IMBD dataset \cite[][custom research license]{Maas.2011} contains moview review excerpts. We simplifiy the dataset to only contain sentences parts containing $n \leq 14$ words. The simplified dataset can be found at \url{https://github.com/mmschlk/shapiq/blob/v1/benchmark/data/simplified_imdb.csv} for reproducibility. 
    \item[\textbf{IC:}] The \textit{ImageClassification} data contains test images from \emph{Imagenet} \cite[][custom research license]{Deng.2009}. The example images can be found at \url{https://github.com/mmschlk/shapiq/tree/v1/shapiq/games/benchmark/imagenet_examples} for reproducibility.
\end{itemize}

All models used for the benchmark games are defined in the code repository. We use decision tree, random forest, k-nearest neighbour, linear/logistic regression models from \texttt{scikit-learn} \cite{pedregosa2011scikitlearn}. Moreover, we use gradient-boosted tree classifiers and regressors from \texttt{xgboost} \cite{chen2016xgboost}. We train small neural networks with \texttt{PyTorch} \cite{Paszke.2017} and use \texttt{PyTorch}'s \texttt{ResNet18} architecture. The movie review language model and the vision transformer is derived from the \texttt{transformers} API \cite{Wolf.2020}.

\clearpage
\subsection{Benchmarking Approximators of SIs with \shapiq}\label{appx_benchmark_listing}

\cref{listing:benchmark} shows an API for benchmarking 4 approximation algorithms on an Dataset Valuation game based on the \emph{AdultCensus} dataset and a gradient boosting decision tree model.\footnote{For details, refer to the notebook example at \url{https://shapiq.readthedocs.io/en/latest/notebooks/benchmark_approximators.html}.}

\begin{listing}[h]
\centering
{
\scriptsize
\begin{boxminted}{}{Python}
import shapiq
from shapiq.games.benchmark.benchmark_config import (
    load_games_from_configuration, 
    print_benchmark_configurations
)
from shapiq.games.benchmark.plot import plot_approximation_quality, add_legend
# print all available games and benchmark configurations
print_benchmark_configurations()
>> Game: AdultCensusDatasetValuation
>> Player ID: 0
>> Number of Players: 10
>> Number of configurations: 3
>> Is the Benchmark Pre-computed: True
>> Iteration Parameter: random_state
>> Configurations:
>> Configuration 1: {'model_name': 'decision_tree', 'player_sizes': 'increasing', 'n_players': 10}
>> Configuration 2: {'model_name': 'random_forest', 'player_sizes': 'increasing', 'n_players': 10}
>> Configuration 3: {'model_name': 'gradient_boosting', 'player_sizes': 'increasing', 'n_players': 10}
>> ...
# load the game files from disk / or download
games = load_games_from_configuration(game_class="AdultCensusDataValuation", n_player_id=0, config_id=2)
games = list(games)  # convert to list (the generator is consumed)
n_players = games[0].n_players
# define the approximators to benchmark
sv_approximators = [
    shapiq.PermutationSamplingSII(n=n_players, index="k-SII", random_state=0), 
    shapiq.SHAPIQ(n=n_players, random_state=0),
    shapiq.SVARMIQ(n=n_players, random_state=0), 
    shapiq.KernelSHAPIQ(n=n_players, random_state=0)
]
# run the benchmark with the chosen parameters
results = shapiq.games.benchmark.run_benchmark(
    index="k-SII",
    order=2,
    games=games,
    approximators=sv_approximators,
    save_path="benchmark_results.json",
    budget_steps=[500, 1000, 2000, 4000],
    n_jobs=8
)
# plot the results
plot_approximation_quality(results)
\end{boxminted}
}
\caption{Exemplary code for benchmarking approximators with \shapiq.}
\label{listing:benchmark}
\end{listing}

\clearpage
\section{Marginal and Conditional Imputers}\label{appx_imputers}

When computing \glspl{SV} and \glspl{SI}, especially for structured tabular data that has a natural interpretation of feature distribution, there is a choice for marginalizing feature influence over either a marginal or a conditional distribution~\citep{Lundberg.2017,Chen.2020,Lundberg.2020,Covert.2020,Sundararajan.2020b,aas2021explaining,olsen2022using,lin2023robustness,olsen2024comparative}. 

For a concrete example~\citep{lin2023robustness}, consider a supervised learning task where a model $f: \mathcal{X} \rightarrow \mathbb{R}$ is used to predict the response variable given an input $\bx$, which consists of individual features $(\bx_1, \bx_2, \ldots , \bx_n)$. Let $p(\bx)$ to represent the data distribution with support on $\mathcal{X} \subseteq \mathbb{R}^n$. We use bold symbols $\bx$ to denote random variables and normal symbols $x$ to denote values. Let $\bx_S$ and $x_S$ denote a subset of features, i.e. players in a game, and values for different $S \subseteq N$, respectively. Then, a cooperative game $\nu: \mathcal{P}(N) \rightarrow \mathbb{R}$ for estimating Shapley-based feature attributions and interactions is defined as 
\begin{align*}
    \nu(S) \coloneqq f_S(\xs) \coloneqq \mathbb{E}_{q(\bxsb)} \big[f(\xs, \bxsb)\big] = \int f(\xs, \xsb) q(\xsb) d\xsb,
\end{align*}
where $\bar{S} = N \setminus S$ denotes the set complement. The feature distribution $q(\bxsb)$ most often considered in the literature is either a \emph{marginal distribution} when $q(\bxsb) \coloneqq p(\bxsb)$~\citep{Lundberg.2017,Covert.2020}, or a \emph{conditional distribution} when $q(\bxsb) \coloneqq p(\bxsb \mid \xs)$~\citep{Frye.2021,aas2021explaining,olsen2022using}. 

In general, empirical estimation of a conditional feature distribution is challenging~\citep{Lundberg.2017,Covert.2020,aas2021explaining,olsen2022using}. Most recently in~\citep{olsen2024comparative}, the authors benchmark several methods for approximating \glspl{SV} based on marginalizing features with a conditional distribution $p(\bxsb \mid \xs)$, without a clear best, i.e. different methods are appropriate in different practical situations. Thus, we combine the decision tree-based and sampling approaches~\citep{olsen2024comparative} to implement a baseline \emph{conditional imputer} in \texttt{shapiq.ConditionalImputer}. The class can be easily extended to include more algorithms, which we leave as future work. The rather standard imputation with a marginal distribution $p(\bxsb)$ is implemented in \texttt{shapiq.MarginalImputer}. Both imputers are used by the appropriate game benchmarks, and available for approximating feature interaction explanations in \texttt{shapiq.TabularExplainer} via the \texttt{imputer} parameter.

\cref{listing:imputer} shows a more advanced API for setting a specific imputer and approximator in \shapiq.\footnote{For details, refer to the notebook example at \url{https://shapiq.readthedocs.io/en/latest/notebooks/conditional_imputer.html}.}

\begin{listing}[h]
\centering
{
\footnotesize
\begin{boxminted}{}{Python}
X, model = ...
import shapiq
# create an imputer object
imputer = shapiq.ConditionalImputer(model=model, data=X, sample_size=100)
# create an approximator object
approximator = shapiq.KernelSHAPIQ(n=X.shape[1], index="SII", max_order=3)
# create an explainer object
explainer = shapiq.Explainer(model, X, imputer=imputer, approximator=approximator)
# choose a sample point to be explained
x = X[0]
# approximate feature interactions given the specificed budget 
interaction_values = explainer.explain(x=x, budget=1024)
# retrieve 3-order feature interactions 
interaction_values.get_n_order_values(3)
# visualize 1-order and 2-order feature interactions on a graph 
interaction_values.plot_network(feature_names=...)
\end{boxminted}
}
\caption{Exemplary code for defining an imputer and approximator for explanation with \shapiq.}
\label{listing:imputer}
\end{listing}

\clearpage
\section{Details of the Experimental Setting and Reproducibility}\label{appx_sec_experiment_description}

\subsection{Generated Cooperative Games}

The game-theoretical quantification of interaction demands a formal cooperative game specified by a player set $N$ and value function $\nu : \mathcal{P}(N) \to \mathbb{R}$.
The players for each benchmark game are already given in \cref{tab_benchmark_overview}, leaving the value functions left open to be specified, with what we catch up here.

\paragraph{Local Explanation.}

For a specified datapoint $x$, the worth of a coalition of features $S$ is given by the model's predicted value $h(x_S)$ using only the features in $S$.
The features outside of $S$ are made absent in $x_S$ by imputing them with a surrogate value in order to remove their information.
For tabular datasets such as \textit{AdultCensus}, \textit{BikeSharing}, and \textit{CaliforniaHousing} this is done by marginal or conditional imputation.
For the language model predicting the sentiment of movie review excerpts, missing words are set to the masked token.
Missing pixels (patches) for the vision transformer image classifier are also set to the masked token.
For the ResNet image classifier removed superpixels are collectively set to a mean value (gray).

\paragraph{Global Explanation.}
Instead of specifying a single datapoint and considering the model's output, the model's loss is averaged over a number of fixed datapoints $x_1,\ldots,x_M$.
The model's loss for a coalition $S$ and datapoint $x_m$ is computed by comparing its prediction $h({x_m}_S)$ with the ground truth target value.
The imputation of absent features is done as for local explanations.

\paragraph{Tree Explanation.}
This is a specialization of local explanations for tree models, made feasible by the capabilities of TreeSHAP-IQ to compute ground-truth \glspl{SV} and \glspl{SI} values, which allows the evaluation games with substantially more features.
Features are imputed according to the tree distribution \cite{Lundberg.2020,Muschalik.2024}.
Consequently, the worth of the empty coalition containing no features is the tree's average prediction, e.g.\ baseline value.


\paragraph{Uncertainty Explanation.}
Similar to local explanations, the model's prediction with missing features imputed to a fixed datapoint is evaluated.
Instead of referring to the predicted value, the value function is given by the prediction's uncertainty for which three measures are available: total, epistemic, and aleatoric uncertainty.
Hence, the Shapley values of the features attribute their individual contribution to the decrease in uncertainty caused by their information.


\paragraph{Feature Selection.}
The available data is split into a training set $\mathcal{D}_{\text{Train}}$ and test set $\mathcal{D}_{\text{Test}}$.
Given a learning algorithm $\mathcal{A}$, a coalitions worth $\nu(S)$ is given by the generalization performance of the model $h_S$ on $\mathcal{D}_{\text{Test}}$ that results from applying $\mathcal{A}$ on $\mathcal{D}_{\text{Train}}$ using only features in $S$, known as \emph{remove-and-refit}.
The worth of the empty coalition is set to 0.


\paragraph{Ensemble Selection.}
Replacing features in feature selection by weak learners, and adapting the learning algorithm to construct an ensemble out of those, leads to ensemble selection.
Each coalition $S$ of base learners is evaluated by the performance of the resulting ensemble on a separate test set, known as \emph{remove-and-re-evaluate}.
Likewise, we set $\nu(\emptyset)=0$.


\paragraph{Data Valuation.}
Continuing in the spirit of \emph{remove-and-refit}, a new model is fitted to each coalition of datapoints.
The generalization performance of the resulting model on a separate test set is set to be the coalition's worth.
The value of the empty coalition is set to 0.


\paragraph{Dataset Valuation.}
The setup is analogous to data valuation, where instead of single datapoints are being understood as players, the available data is partioned and each subset is viewd as a player.


\paragraph{Cluster Explanation.}
Similar to feature selection, \emph{remove-and-refit} is applied.
Instead of fitting a model, a clustering algorithm forms multiple clusters on the dataset using only the available features of a coalition $S$. 
The worth $\nu(s)$ is given by a cluster evaluation score (see \cref{tab_app_benchamrk_configs_1,tab_app_benchamrk_configs_2} for details).
A cluster score of 0 is assigned to the empty coalition.


\paragraph{Unsupervised Feature Importance.}

Given a coalition of features $S$, a set of datapoints can be understood as observations generated by a joint distribution of $S$ and used to estimate this distribution by measuring the frequencies of feature values.
This in turn allows to measure entropy and thus also total correlation of a subset of features which is used as the worth of $S$.
Since the total correlation measures the amount of shared information, each feature's assigned Shapley value quantifies its contributed information to the group.
The total correlation of the empty set is naturally 0.


\paragraph{Sum of Unanimity Models (SOUMs).}
A unanimity game is a synthetic game for a coalition $U \subseteq N$ with $\nu_U(T) := \mathbf{1}_{T\supseteq U}$, i.e. outputs one, if all players of $U$ are present, and zero otherwise.
The sum of unanimity model (SOUM) is a linear combination of randomly sampled unanimity games.
For uniformly sampled coefficients $a_1,\dots,a_m \in [-1,1]$ and subsets $U_1,\dots,U_m \subseteq N$ uniformly sampled by size, where we restrict the SOUM to specific maximum subset sizes.
The value function then reads as
\begin{equation*}
    \nu(T) := \sum_{\ell=1}^m a_\ell \nu_{U_\ell}(T).
\end{equation*}

For SOUMs, the \glspl{MI} as well as all \glspl{SI} can be efficiently computed in linear time, cf.~\citep[Appendix~B.7]{Fumagalli.2024}.

\subsection{Computational Resources}\label{appx_sec_comp_resources}
This section contains additional information regarding the computational resources required for the empirical evaluation of this work.
The main computational burden stems from pre-computing the benchmark games for $n \leq 16$ players and from running all of \shapiq's \gls{SV} and $2$-SII approximation methods.
Still, the experiments require only a modest range of computational resources.
The games are pre-computed on a ``11th Gen Intel(R) Core(TM) i7--11800H 2.30GHz'' machine requiring around 240 CPU hours.
The approximation experiments have been run on a compute cluster using 80 CPUs of four ``AMD EPYC 7513 32--Core Processor'' units for 24 hours resulting in about 1920 CPU hours.

\subsection{Data Availability and Reproducability}\label{appx_section_data_availability}
The data to the pre-computed games is available at \url{https://github.com/mmschlk/shapiq/tree/v1}.
Utility functions exist in \shapiq that automatically download and instantiate the games.
The code for reproducing the experimental evaluation can be found at \url{https://github.com/mmschlk/shapiq/tree/v1/benchmark} and \url{https://github.com/mmschlk/shapiq/tree/v1/complexity_accuracy}.

\clearpage
\section{Additional Benchmark Approximation Results}\label{appx_sec_approx_results}

\begin{figure}[!htb]
    \centering
    \includegraphics[width=0.5\textwidth]{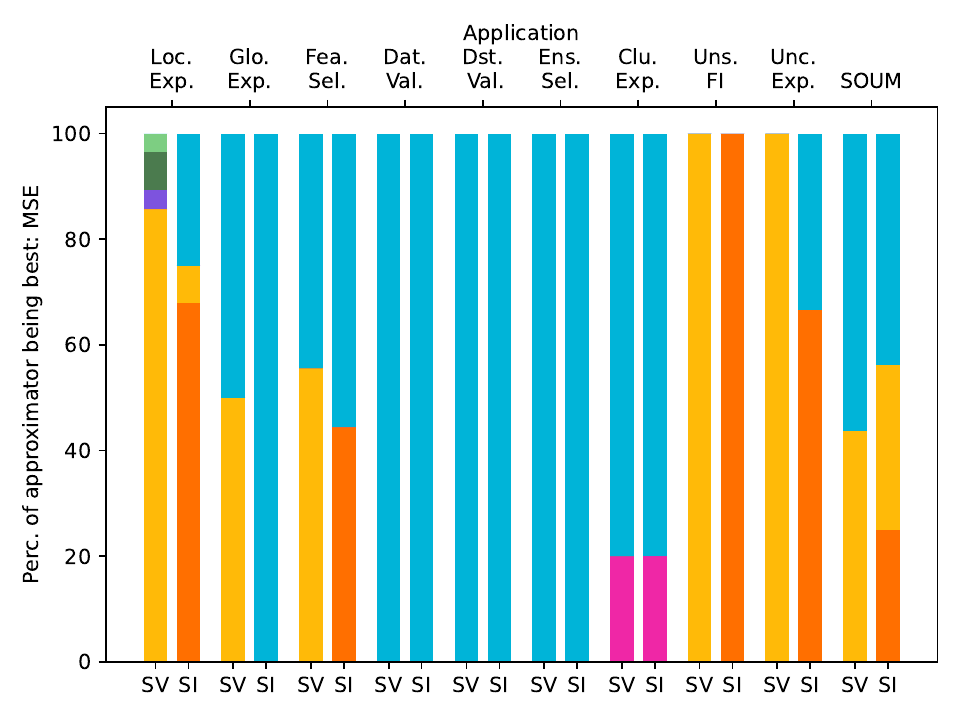}
    \\
    \includegraphics[width=0.5\textwidth]{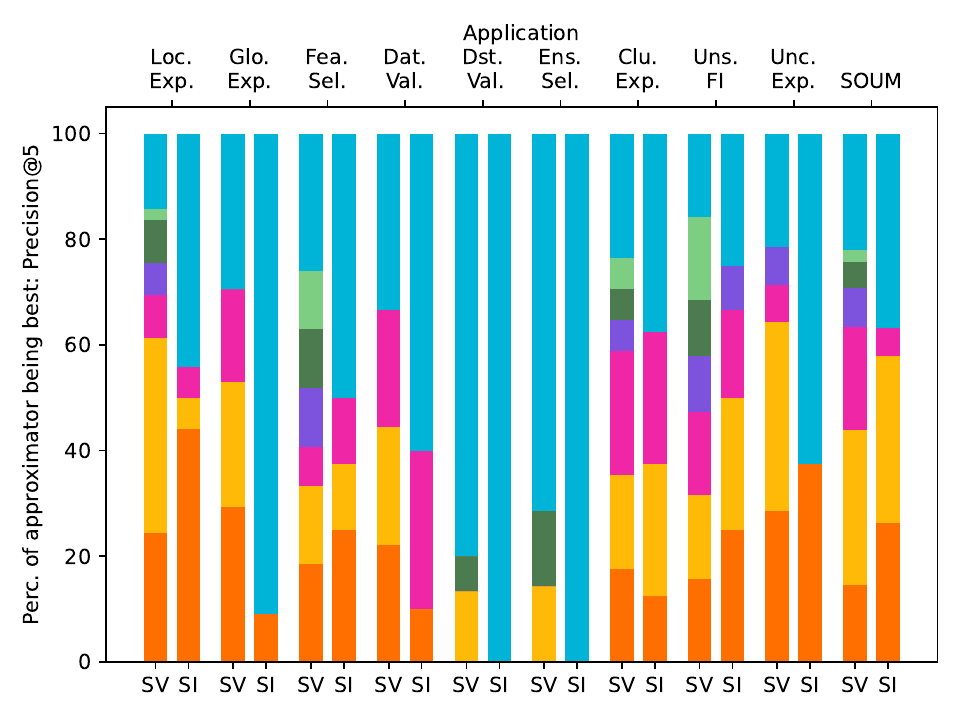}
    \\
    \includegraphics[width=0.5\textwidth]{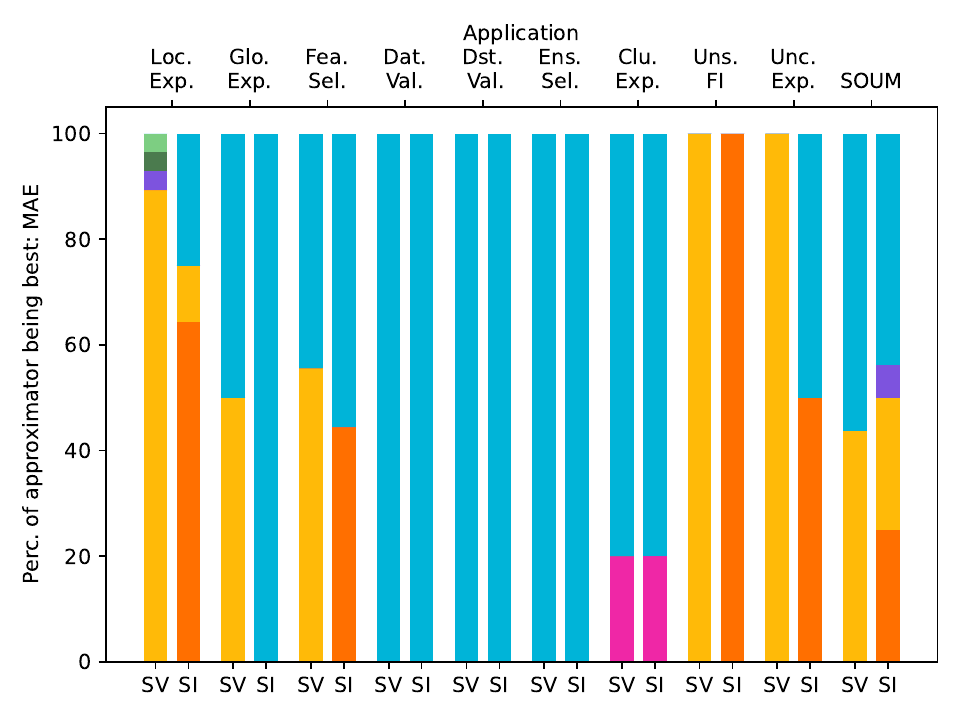}
    \caption{Benchmark overview approximation results for \gls{MSE} (top), \gls{prec} (middle), and \gls{MAE} (bottom).}
    \label{appx_figure_overview}
\end{figure}

\begin{figure}[!htb]
    \centering
    \hspace{1.5em}
    \includegraphics[width=\textwidth]{figures/legend.pdf}
    \\[0.5em]
    \begin{minipage}[c]{0.24\textwidth}
        \includegraphics[width=\textwidth]{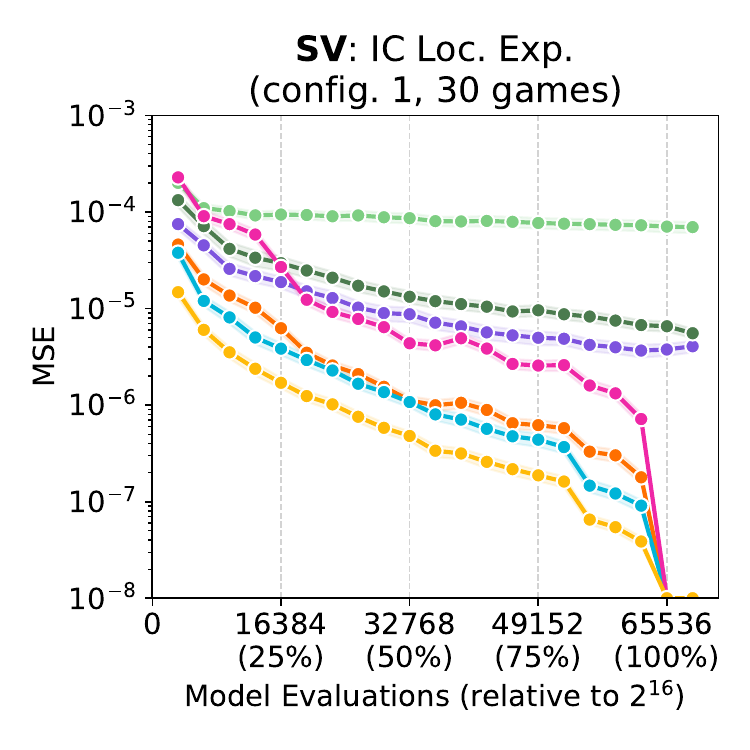}
    \end{minipage}
    \begin{minipage}[c]{0.24\textwidth}
        \includegraphics[width=\textwidth]{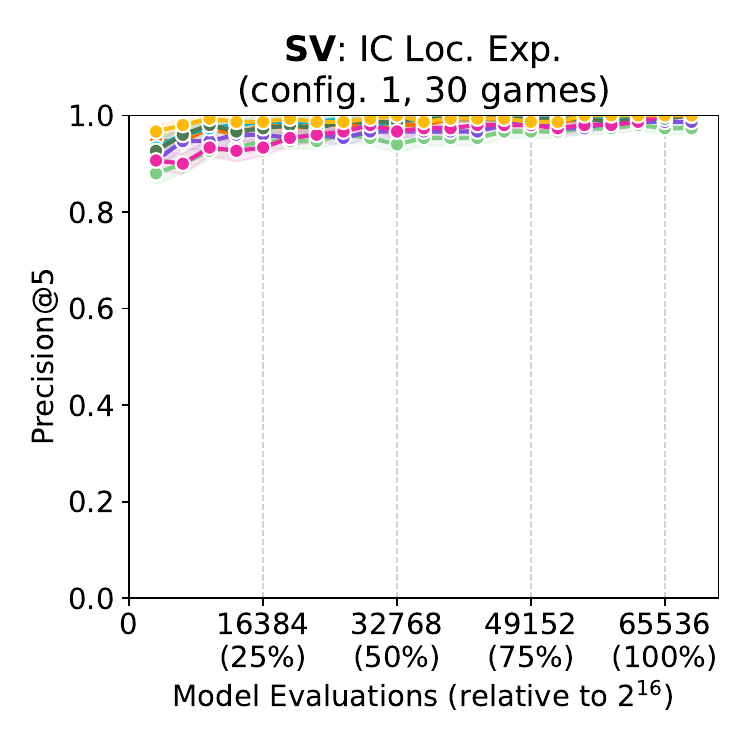}
    \end{minipage}
    \begin{minipage}[c]{0.24\textwidth}
        \includegraphics[width=\textwidth]{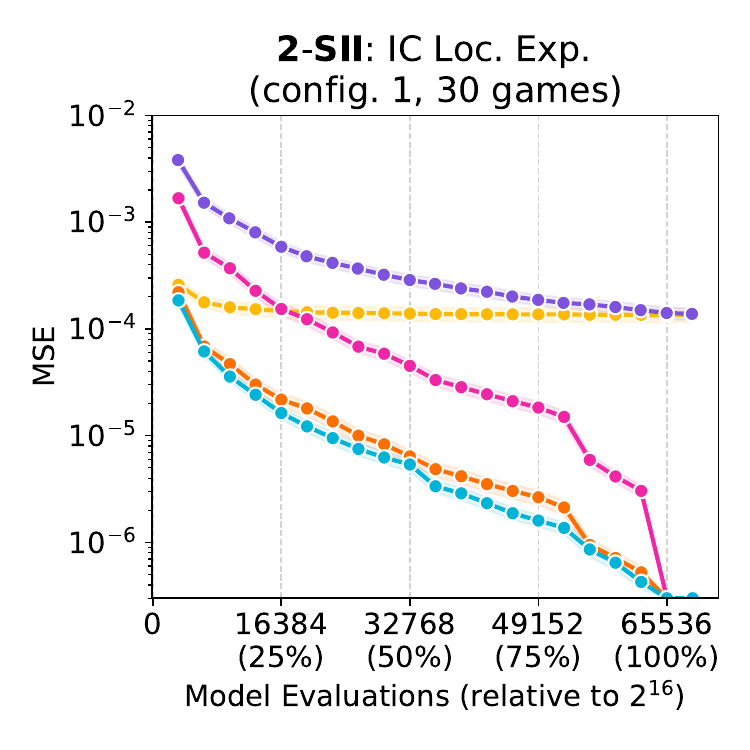}
    \end{minipage}
    \begin{minipage}[c]{0.24\textwidth}
        \includegraphics[width=\textwidth]{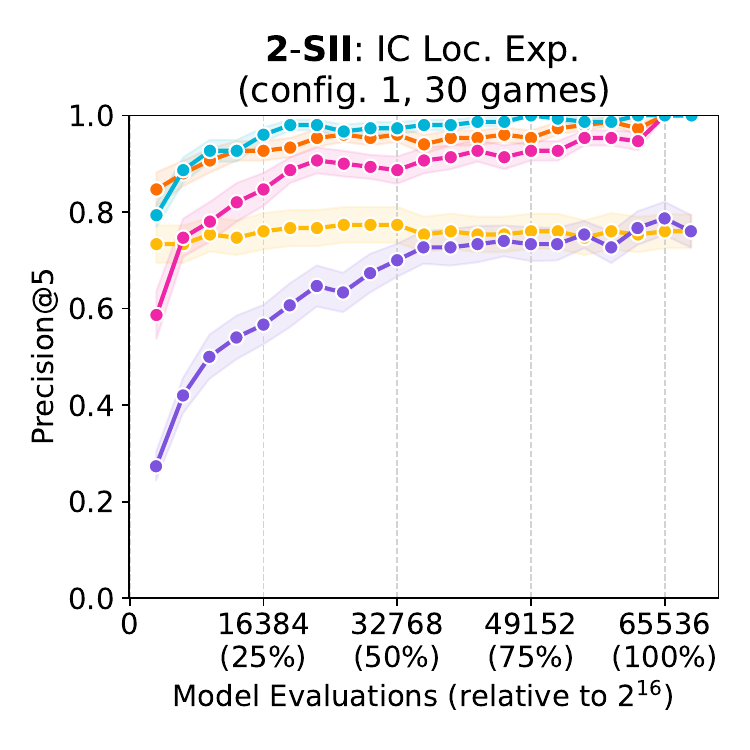}
    \end{minipage}
    \\
    \begin{minipage}[c]{0.24\textwidth}
        \includegraphics[width=\textwidth]{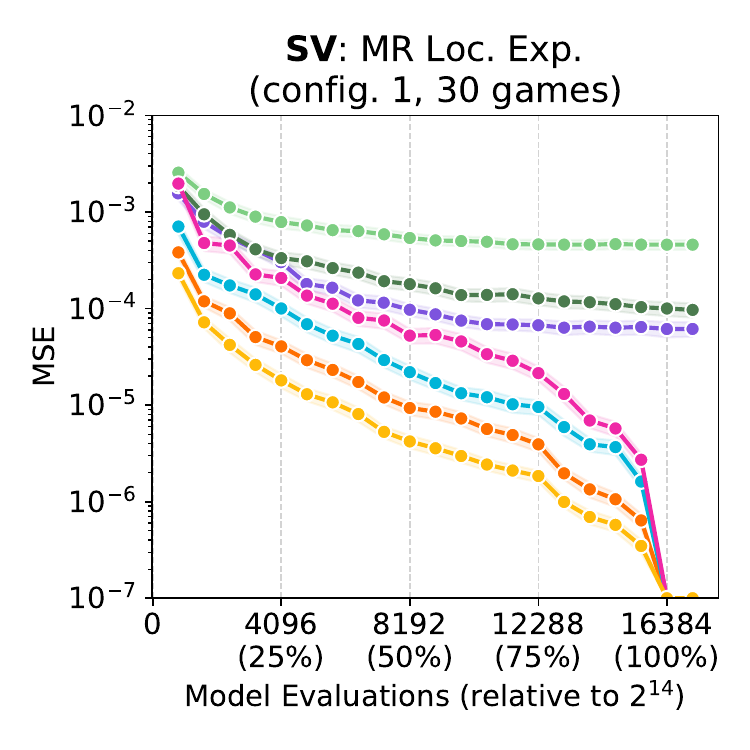}
    \end{minipage}
    \begin{minipage}[c]{0.24\textwidth}
        \includegraphics[width=\textwidth]{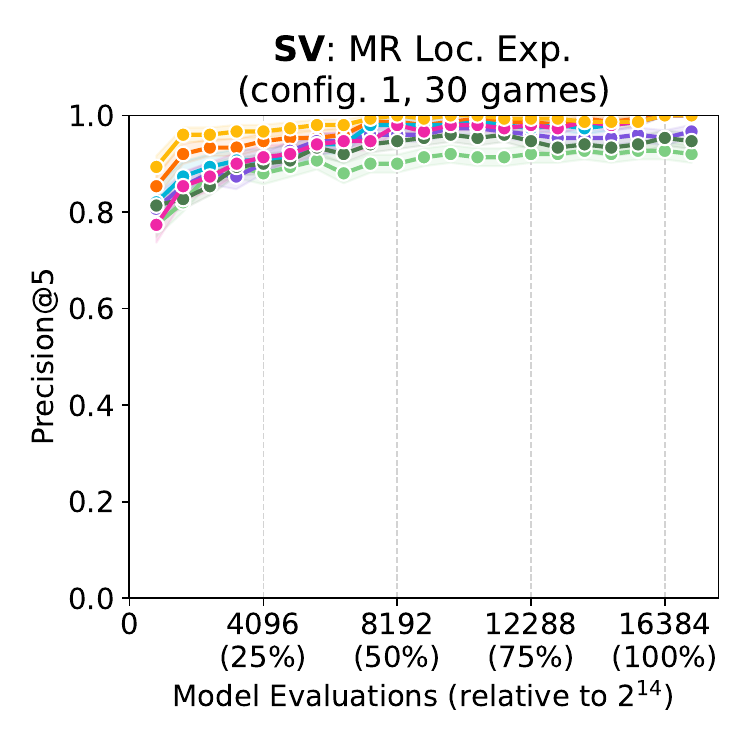}
    \end{minipage}
    \begin{minipage}[c]{0.24\textwidth}
        \includegraphics[width=\textwidth]{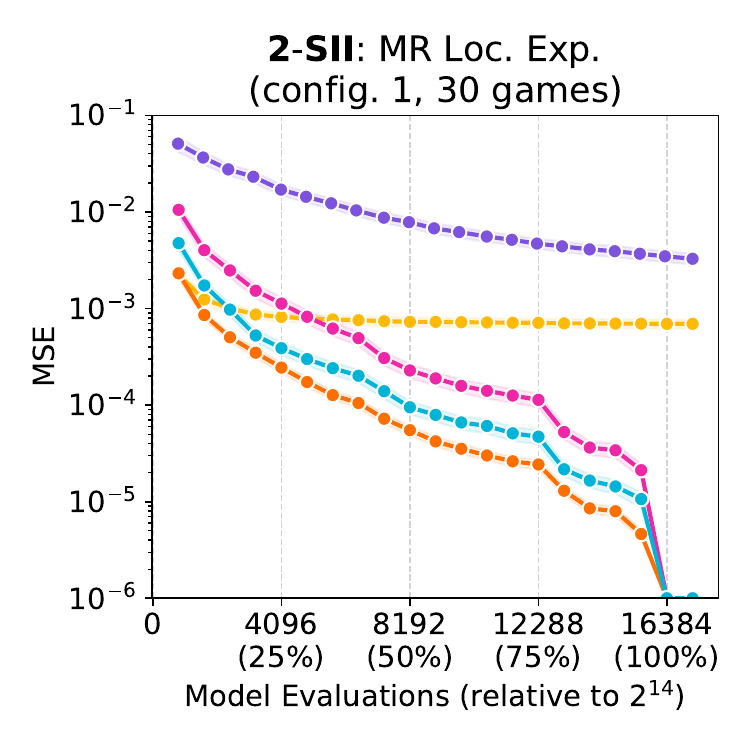}
    \end{minipage}
    \begin{minipage}[c]{0.24\textwidth}
        \includegraphics[width=\textwidth]{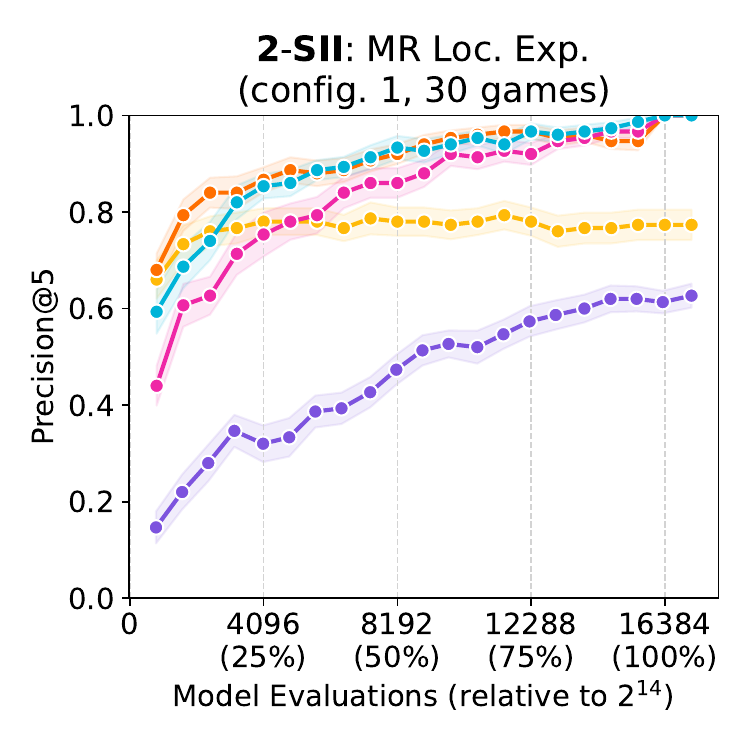}
    \end{minipage}
    \\
    \begin{minipage}[c]{0.24\textwidth}
        \includegraphics[width=\textwidth]{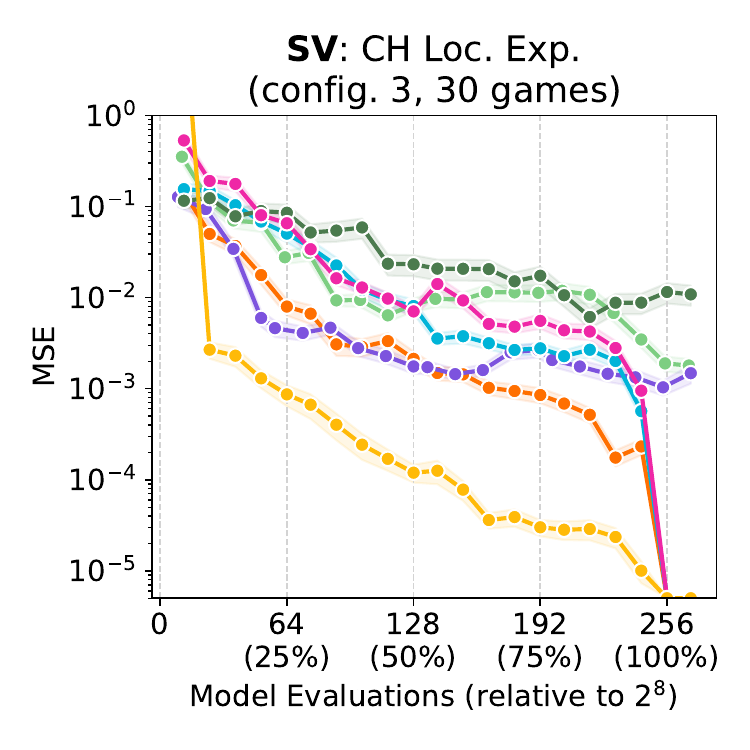}
    \end{minipage}
    \begin{minipage}[c]{0.24\textwidth}
        \includegraphics[width=\textwidth]{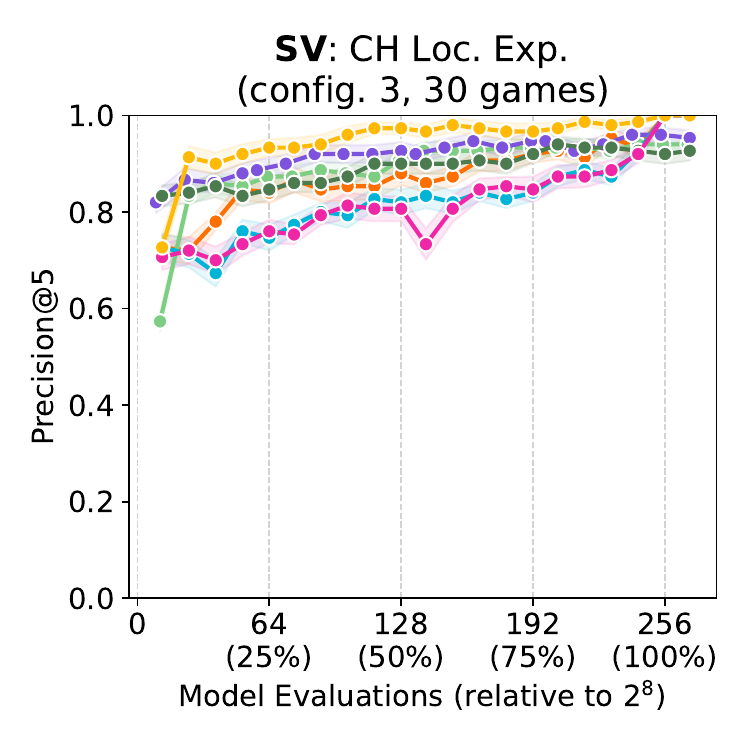}
    \end{minipage}
    \begin{minipage}[c]{0.24\textwidth}
        \includegraphics[width=\textwidth]{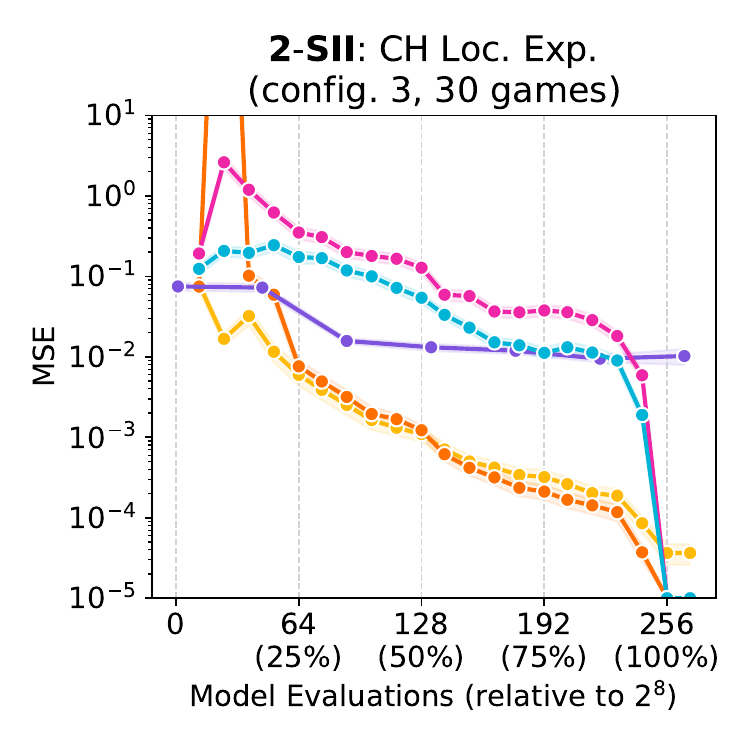}
    \end{minipage}
    \begin{minipage}[c]{0.24\textwidth}
        \includegraphics[width=\textwidth]{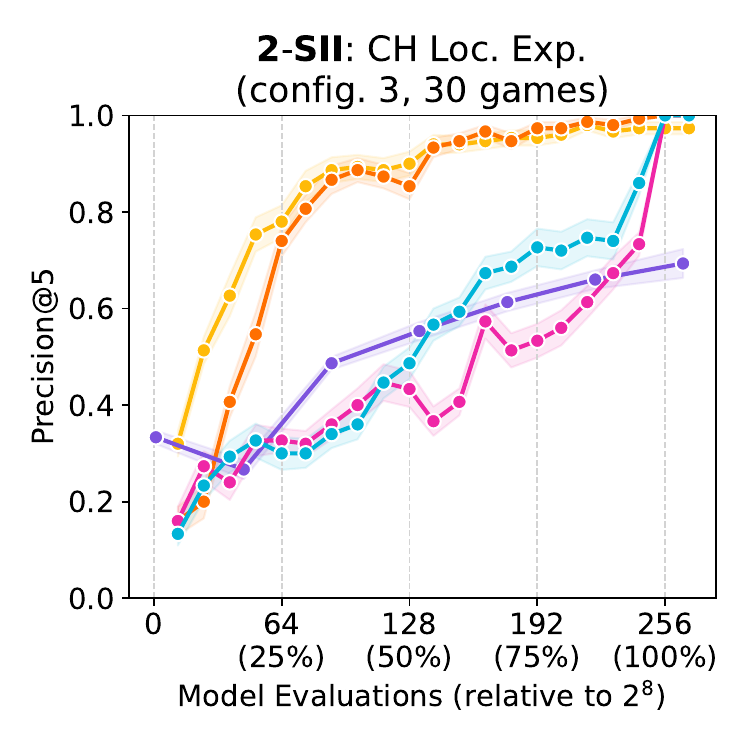}
    \end{minipage}
    \\
    \begin{minipage}[c]{0.24\textwidth}
        \includegraphics[width=\textwidth]{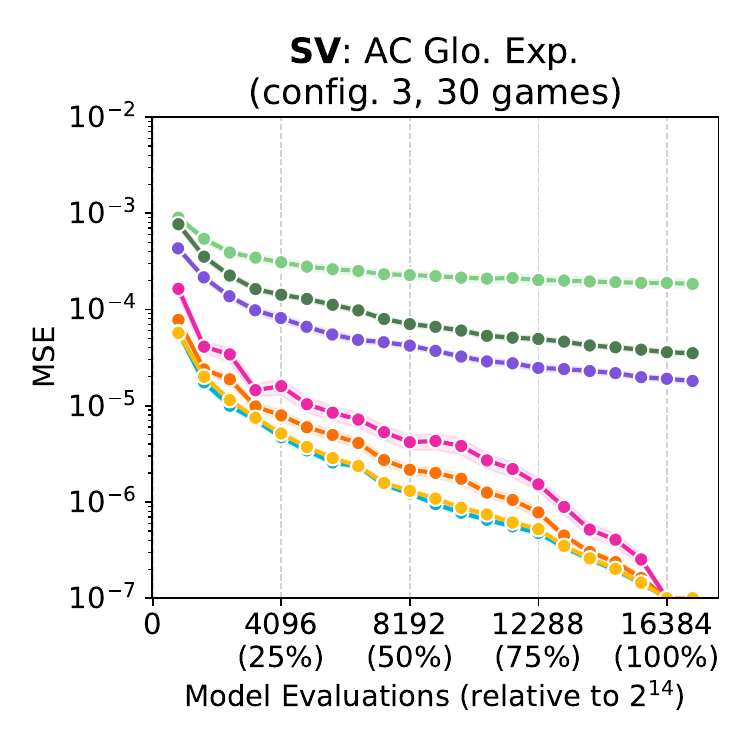}
    \end{minipage}
    \begin{minipage}[c]{0.24\textwidth}
        \includegraphics[width=\textwidth]{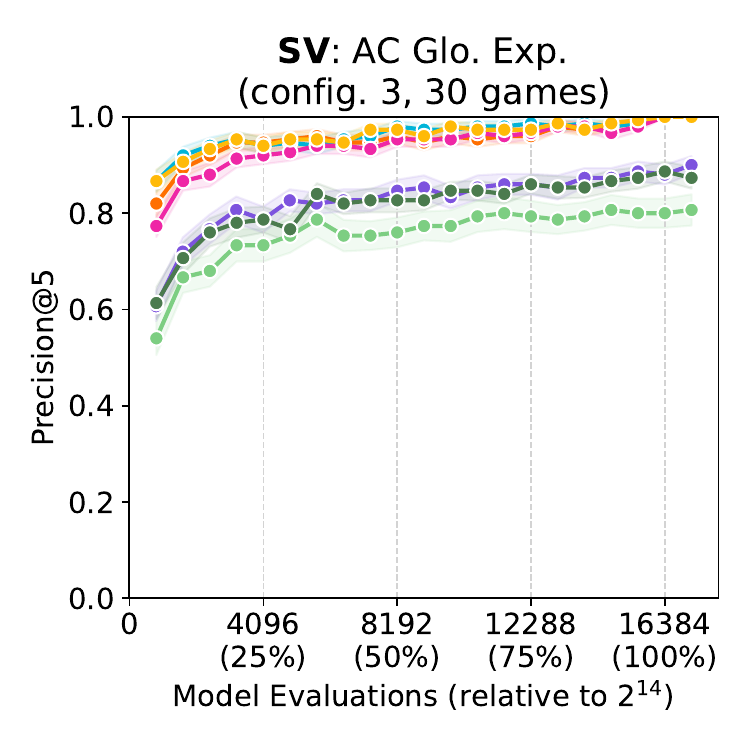}
    \end{minipage}
    \begin{minipage}[c]{0.24\textwidth}
        \includegraphics[width=\textwidth]{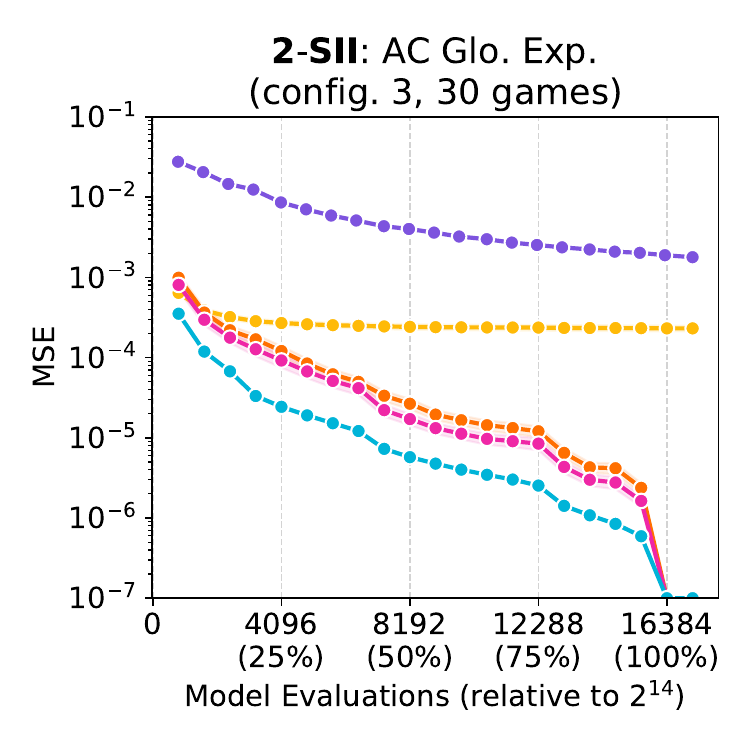}
    \end{minipage}
    \begin{minipage}[c]{0.24\textwidth}
        \includegraphics[width=\textwidth]{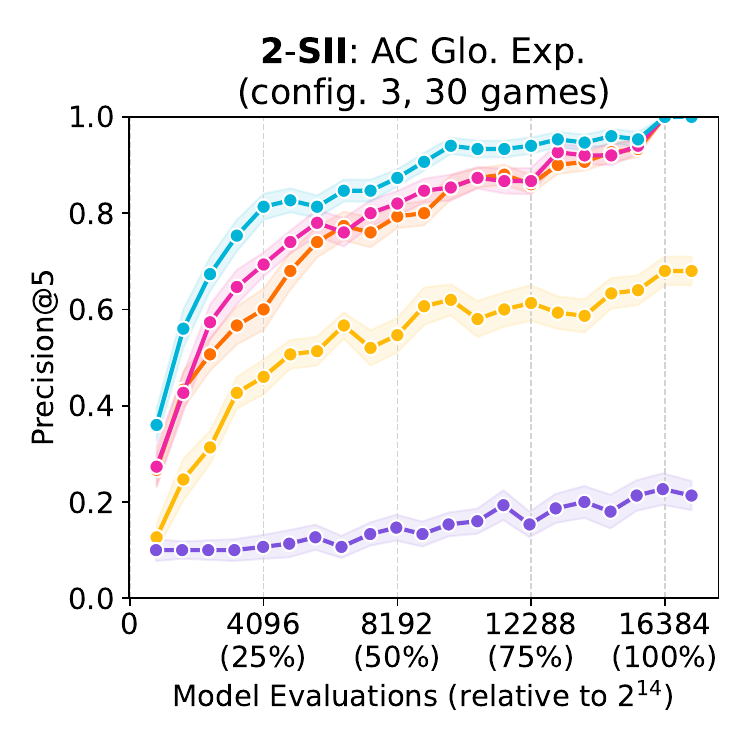}
    \end{minipage}
    \caption{Additional \gls{SV} (column one and two) and \gls{SI} (column three and four) approximation results for different benchmark games from the Local Explanation (first row, vision transformer image classifier with $n=16$ patches), Local Explanation (second row, language model predicting movie review sentiment with $n=14$ words), Local Explanation (third row, dataset \textit{CaliforniaHousing} with $n=8$ features) and Global Explanation (fourth row, dataset \textit{AdultCensus} with $n=14$ features) domain.}
    \label{appx_fig_approximation_1}
\end{figure}

\begin{figure}[!htb]
    \centering
    \hspace{1.5em}
    \includegraphics[width=\textwidth]{figures/legend.pdf}
    \\[0.5em]
    \begin{minipage}[c]{0.24\textwidth}
        \includegraphics[width=\textwidth]{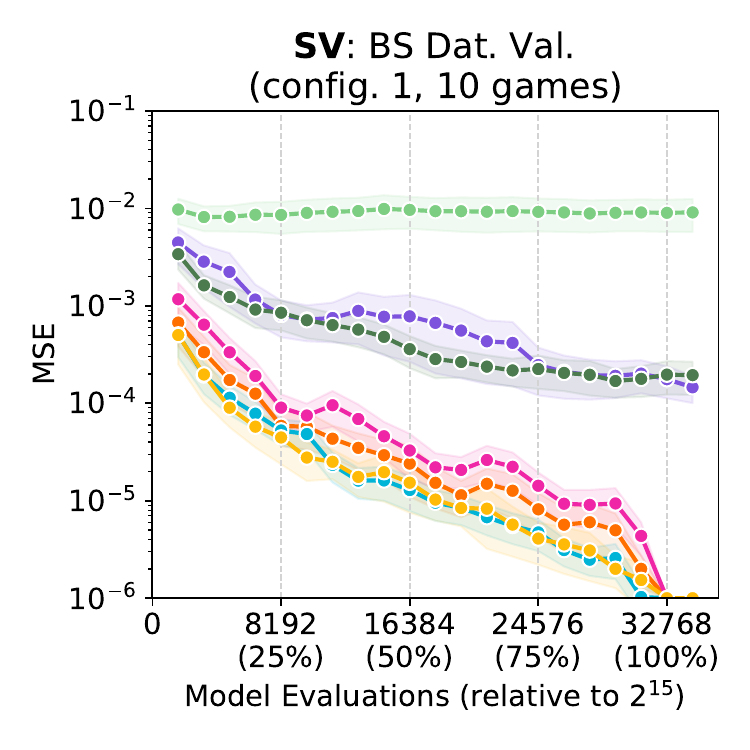}
    \end{minipage}
    \begin{minipage}[c]{0.24\textwidth}
        \includegraphics[width=\textwidth]{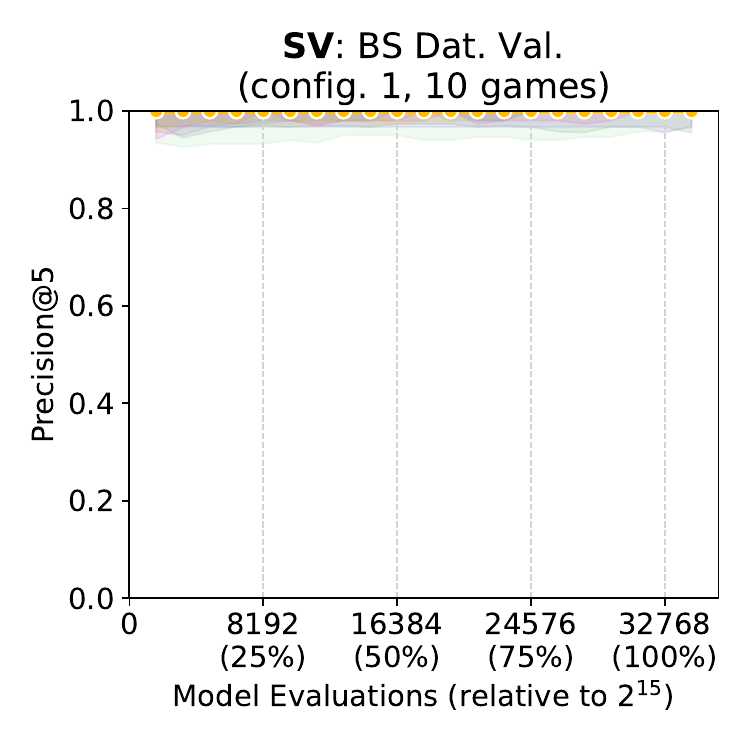}
    \end{minipage}
    \begin{minipage}[c]{0.24\textwidth}
        \includegraphics[width=\textwidth]{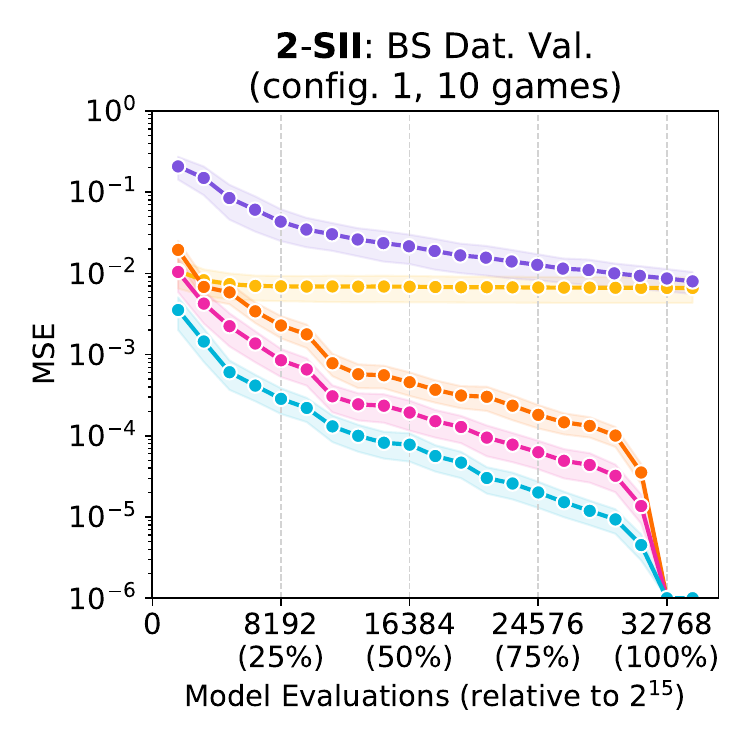}
    \end{minipage}
    \begin{minipage}[c]{0.24\textwidth}
        \includegraphics[width=\textwidth]{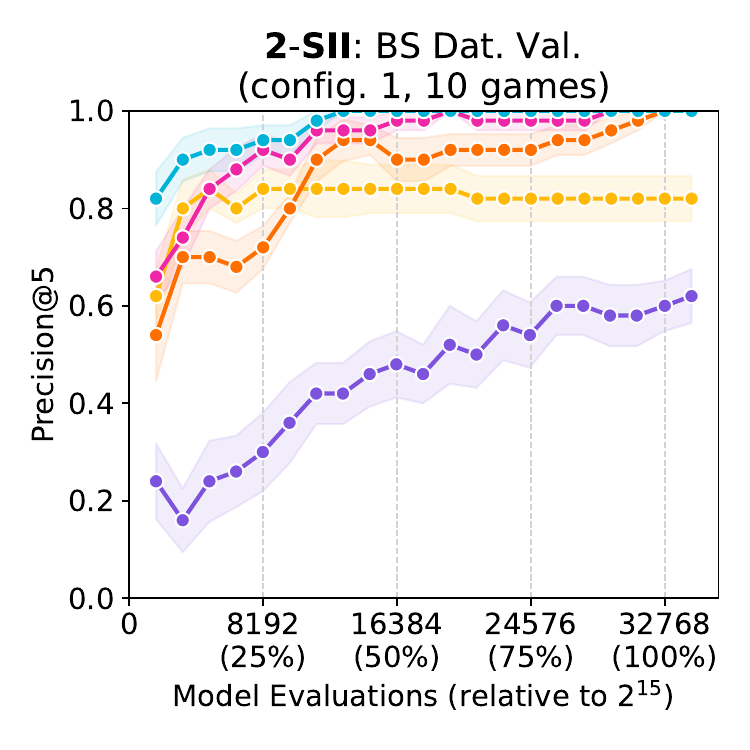}
    \end{minipage}
    \\
    \begin{minipage}[c]{0.24\textwidth}
        \includegraphics[width=\textwidth]{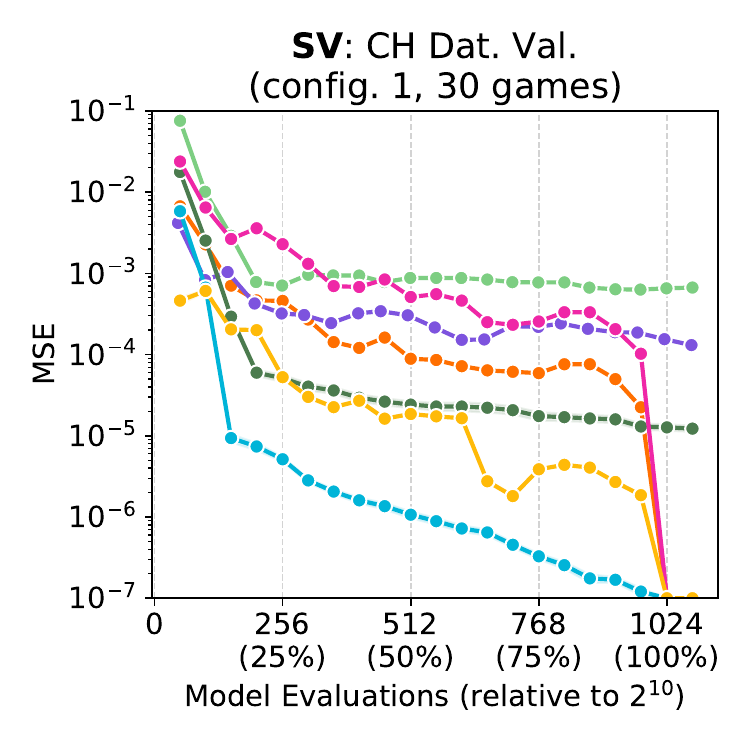}
    \end{minipage}
    \begin{minipage}[c]{0.24\textwidth}
        \includegraphics[width=\textwidth]{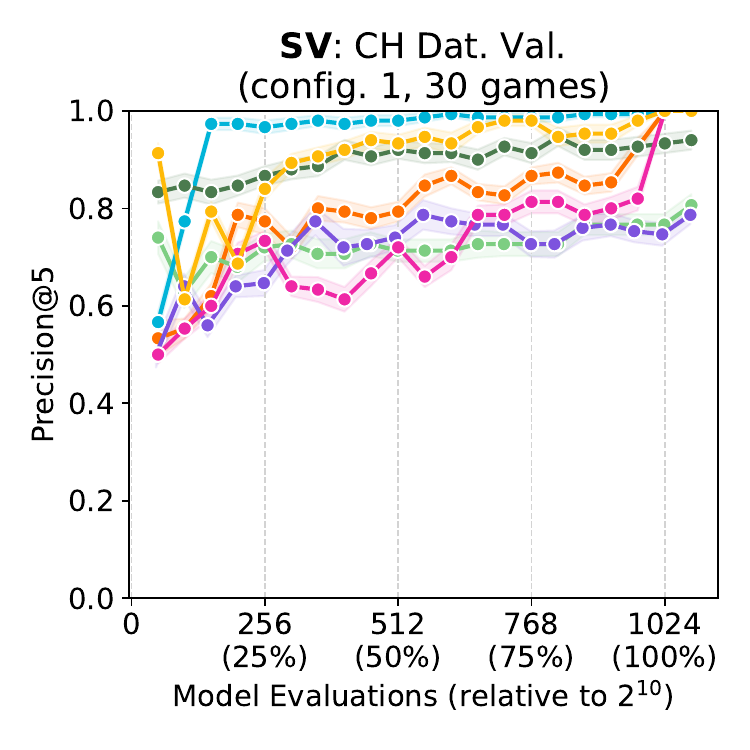}
    \end{minipage}
    \begin{minipage}[c]{0.24\textwidth}
        \includegraphics[width=\textwidth]{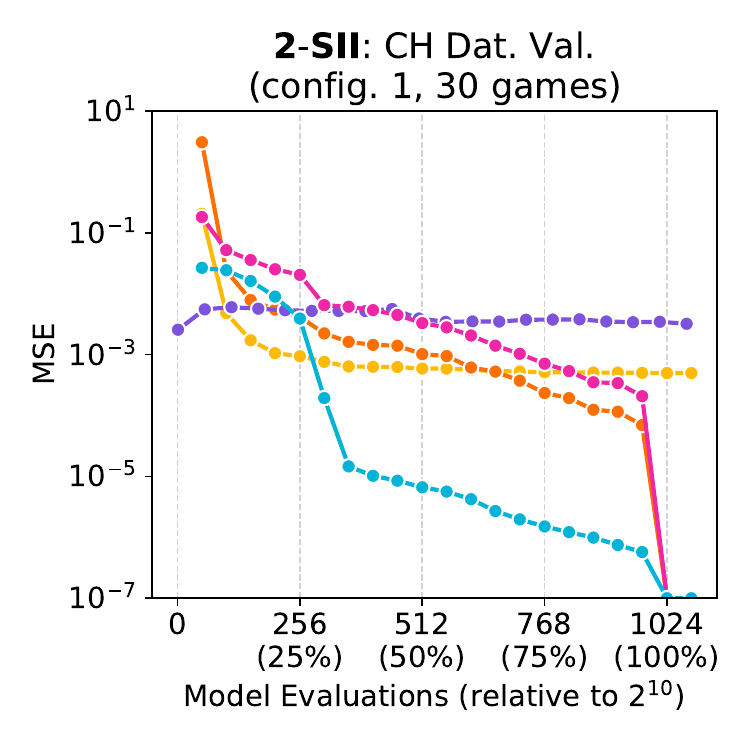}
    \end{minipage}
    \begin{minipage}[c]{0.24\textwidth}
        \includegraphics[width=\textwidth]{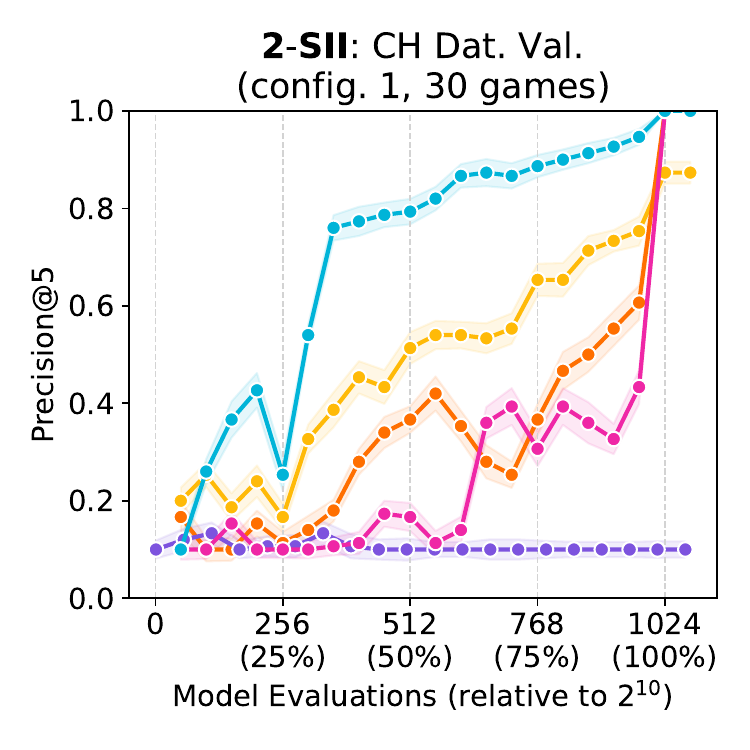}
    \end{minipage}
    \\
    \begin{minipage}[c]{0.24\textwidth}
        \includegraphics[width=\textwidth]{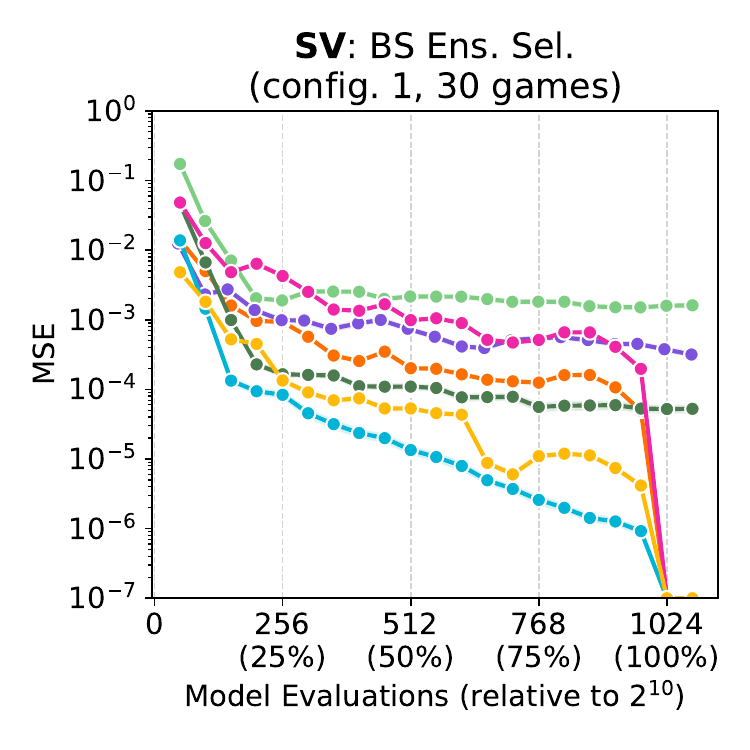}
    \end{minipage}
    \begin{minipage}[c]{0.24\textwidth}
        \includegraphics[width=\textwidth]{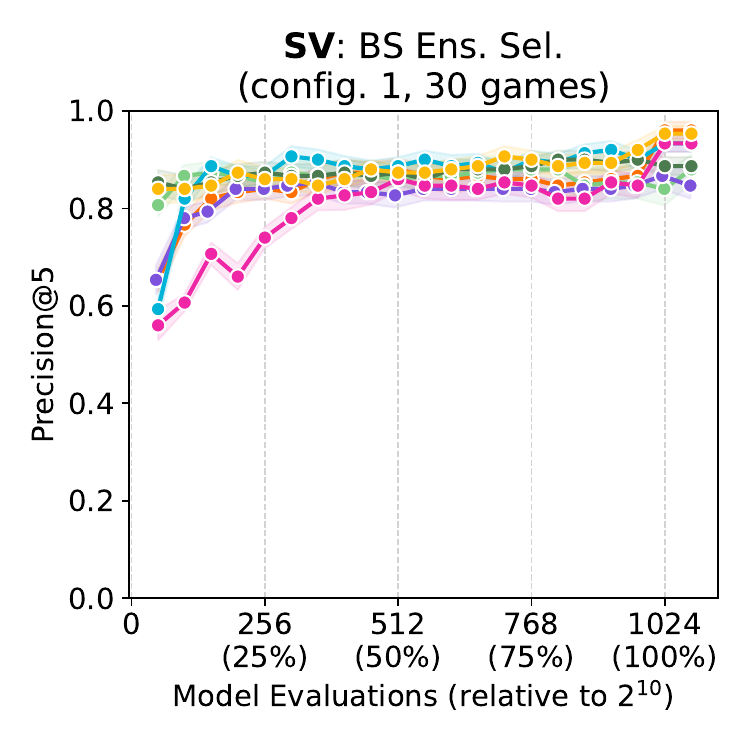}
    \end{minipage}
    \begin{minipage}[c]{0.24\textwidth}
        \includegraphics[width=\textwidth]{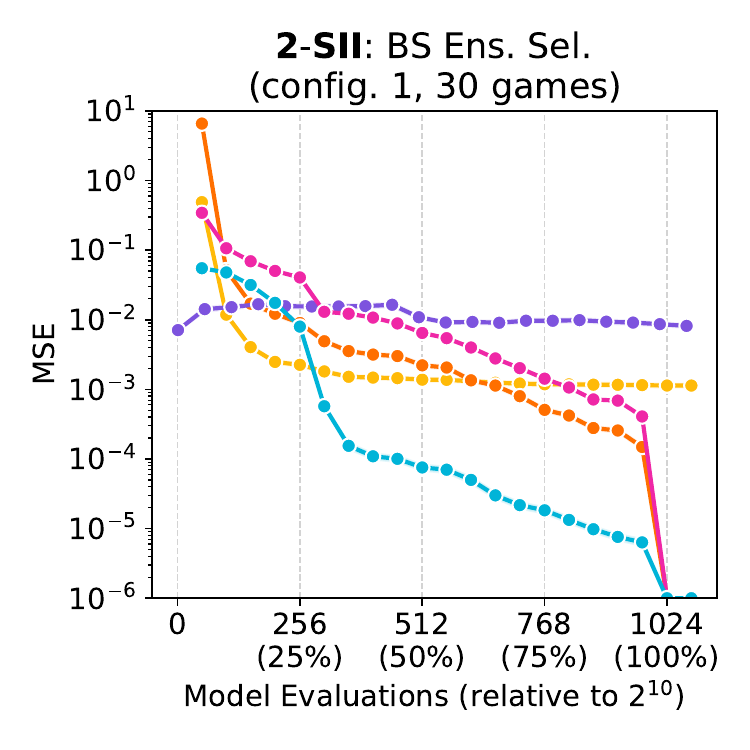}
    \end{minipage}
    \begin{minipage}[c]{0.24\textwidth}
        \includegraphics[width=\textwidth]{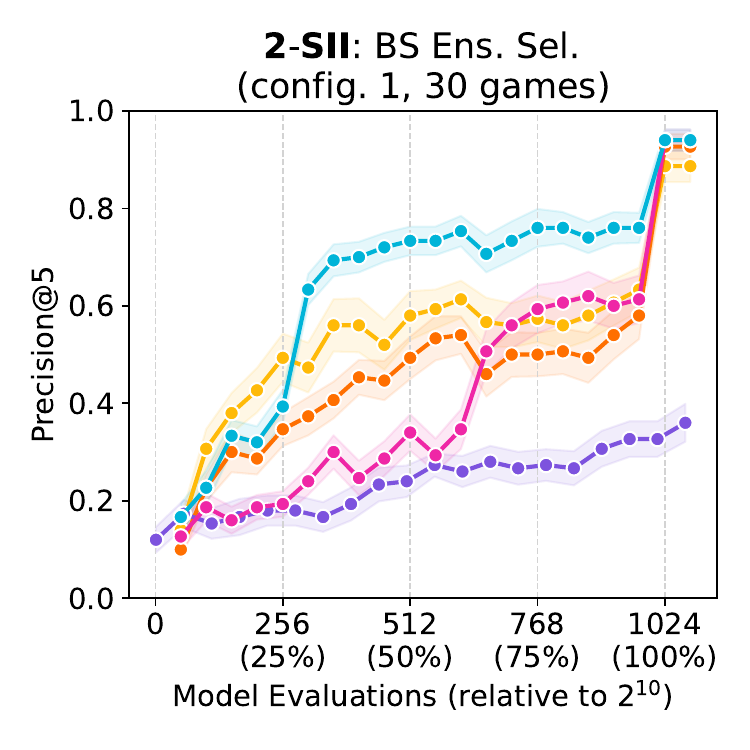}
    \end{minipage}
    \\
    \begin{minipage}[c]{0.24\textwidth}
        \includegraphics[width=\textwidth]{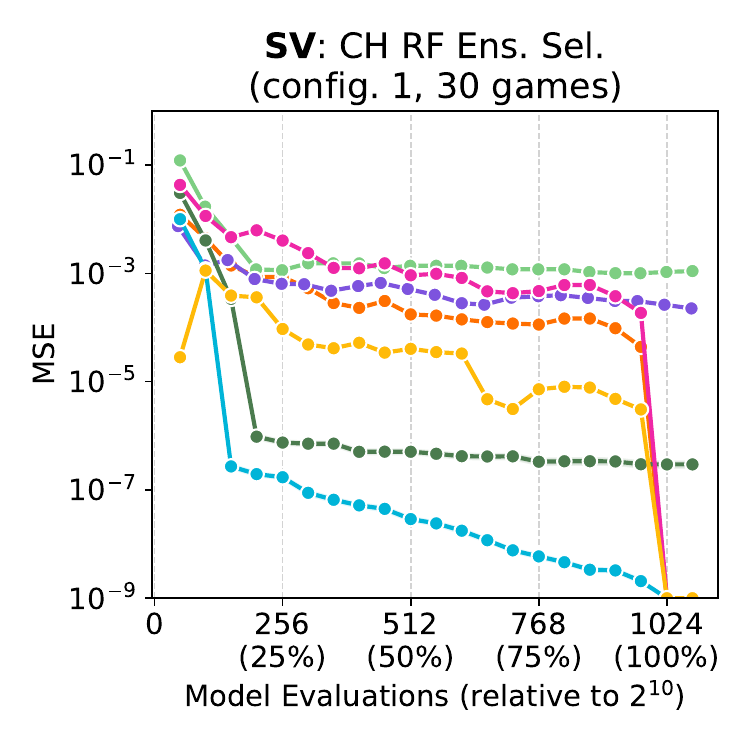}
    \end{minipage}
    \begin{minipage}[c]{0.24\textwidth}
        \includegraphics[width=\textwidth]{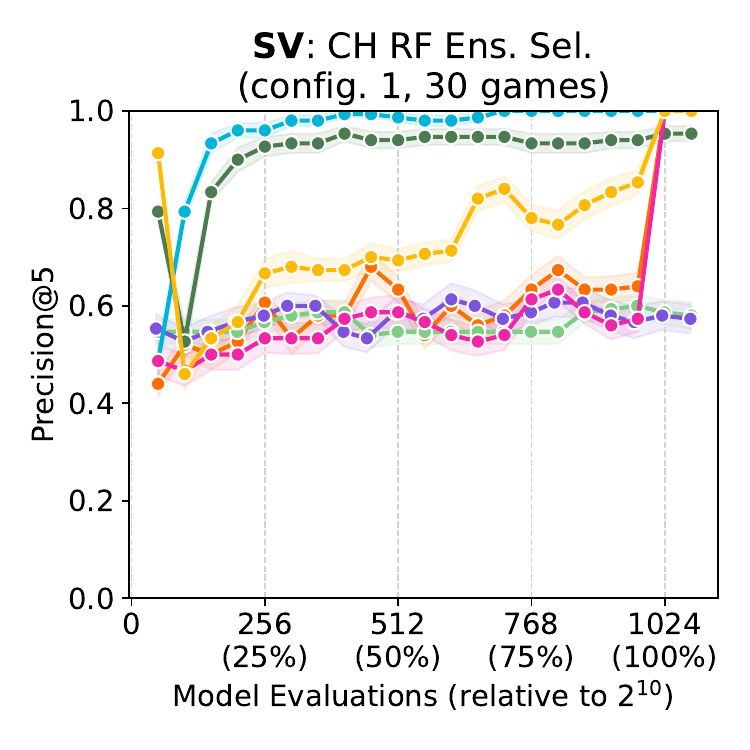}
    \end{minipage}
    \begin{minipage}[c]{0.24\textwidth}
        \includegraphics[width=\textwidth]{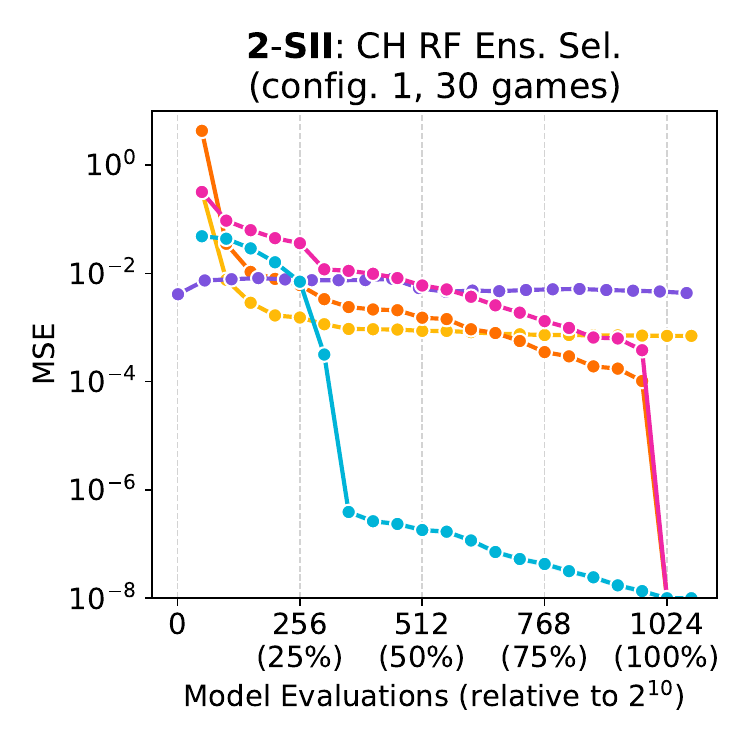}
    \end{minipage}
    \begin{minipage}[c]{0.24\textwidth}
        \includegraphics[width=\textwidth]{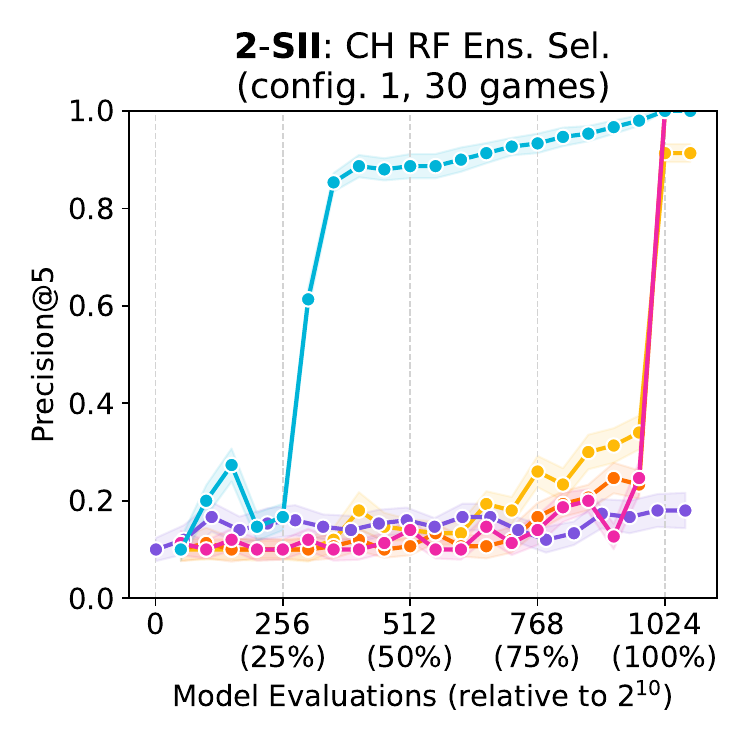}
    \end{minipage}
    \caption{Additional \gls{SV} (column one and two) and \gls{SI} (column three and four) approximation results for different benchmark games from the Data Valuation (first row, \textit{BikeSharing} with $n=12$ features), Dataset Valuation (second row, \textit{CaliforniaHousing} with $n=8$ features), Ensemble Selection (third row, dataset \textit{BikeSharing} with $n=12$ features) and Random Forest Ensemble Selection (fourth row, dataset \textit{CaliforniaHousing} with $n=8$ features) domain.}
    \label{appx_fig_approximation_2}
\end{figure}

\begin{figure}[!htb]
    \centering
    \hspace{1.5em}
    \includegraphics[width=\textwidth]{figures/legend.pdf}
    \\[0.5em]
    \begin{minipage}[c]{0.24\textwidth}
        \includegraphics[width=\textwidth]{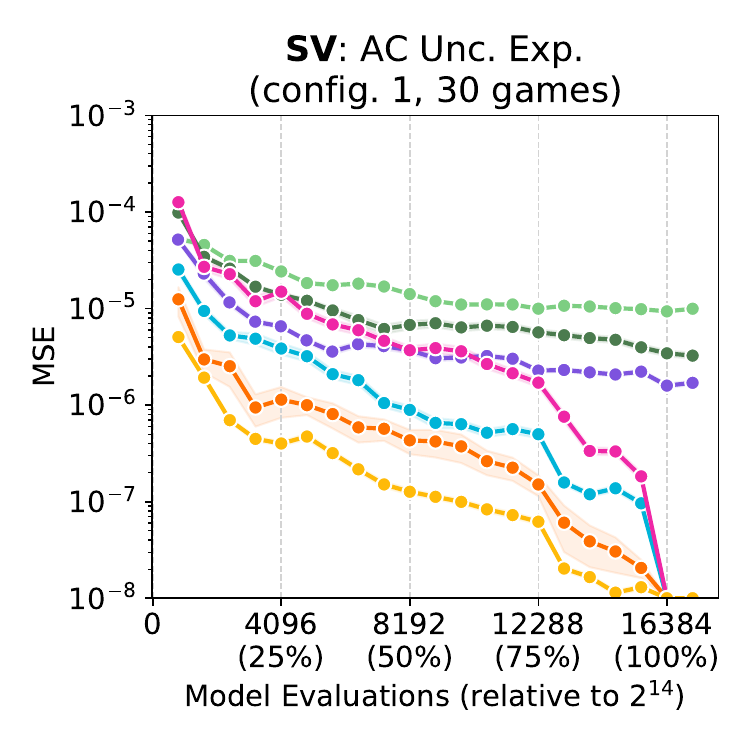}
    \end{minipage}
    \begin{minipage}[c]{0.24\textwidth}
        \includegraphics[width=\textwidth]{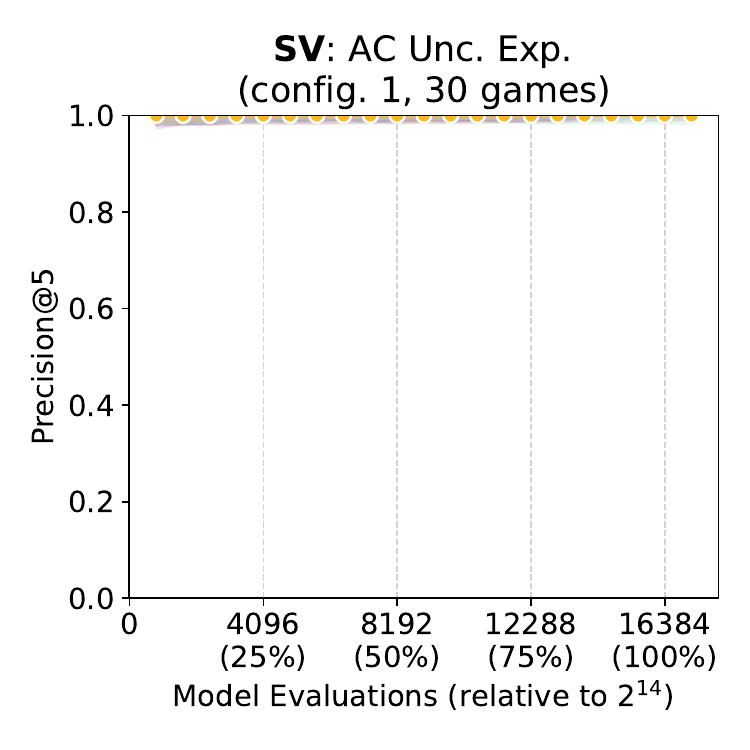}
    \end{minipage}
    \begin{minipage}[c]{0.24\textwidth}
        \includegraphics[width=\textwidth]{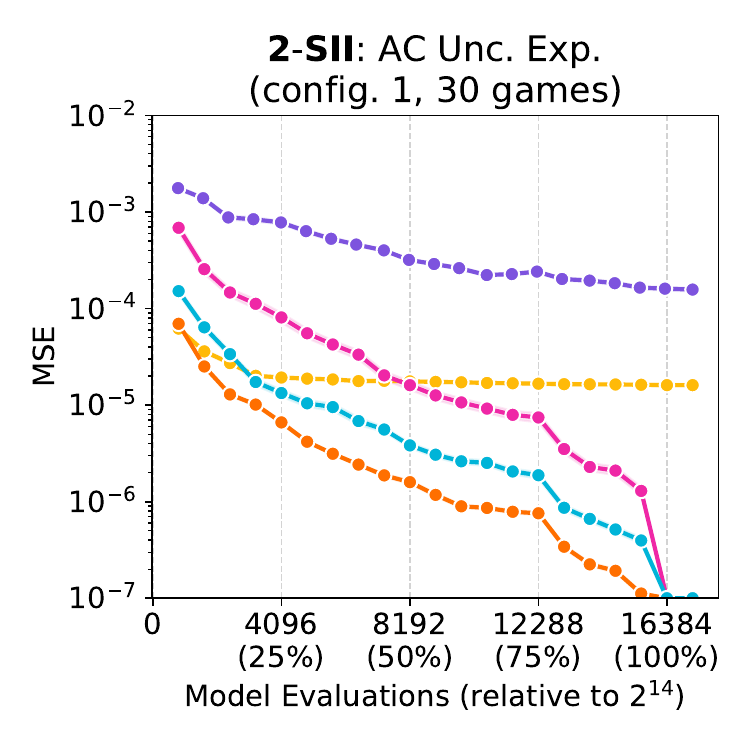}
    \end{minipage}
    \begin{minipage}[c]{0.24\textwidth}
        \includegraphics[width=\textwidth]{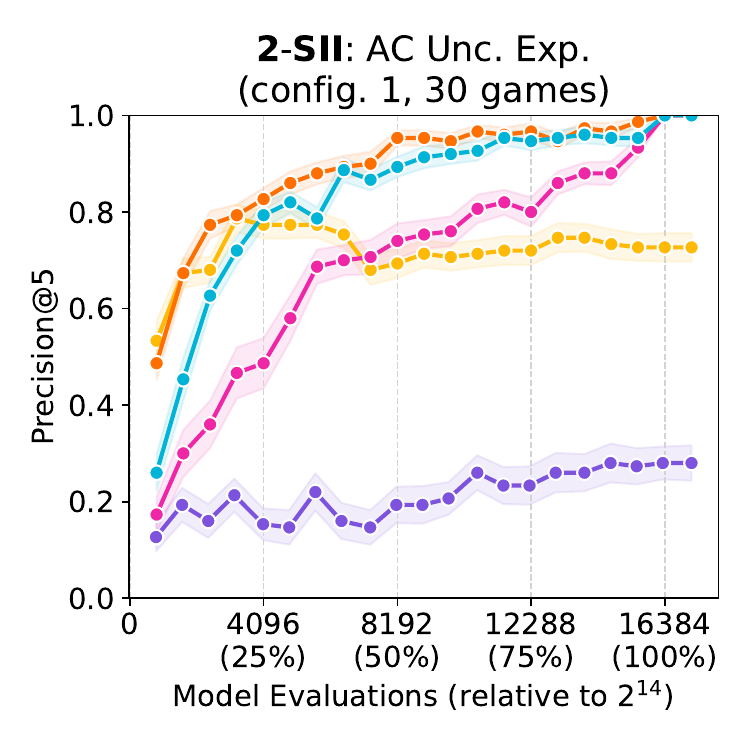}
    \end{minipage}
    \\
    \begin{minipage}[c]{0.24\textwidth}
        \includegraphics[width=\textwidth]{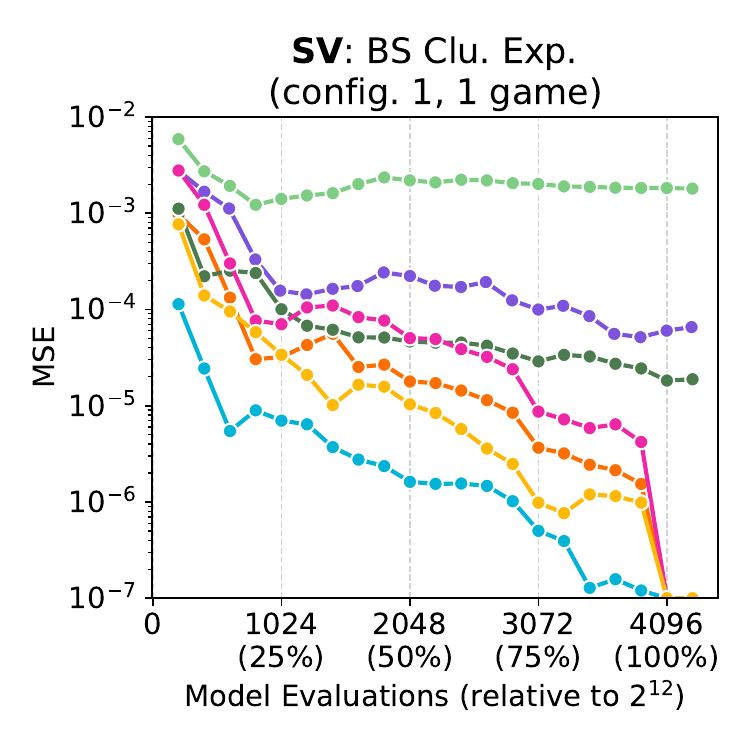}
    \end{minipage}
    \begin{minipage}[c]{0.24\textwidth}
        \includegraphics[width=\textwidth]{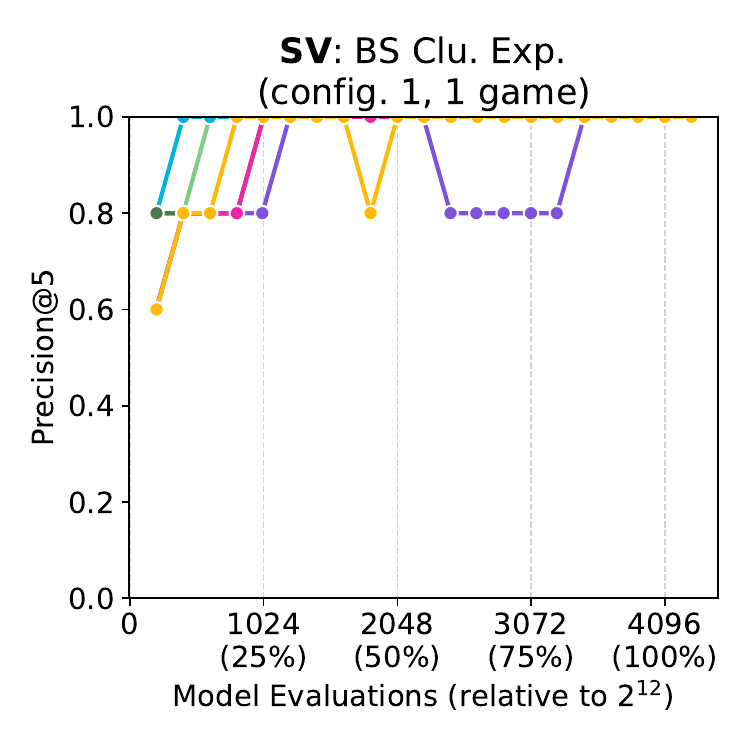}
    \end{minipage}
    \begin{minipage}[c]{0.24\textwidth}
        \includegraphics[width=\textwidth]{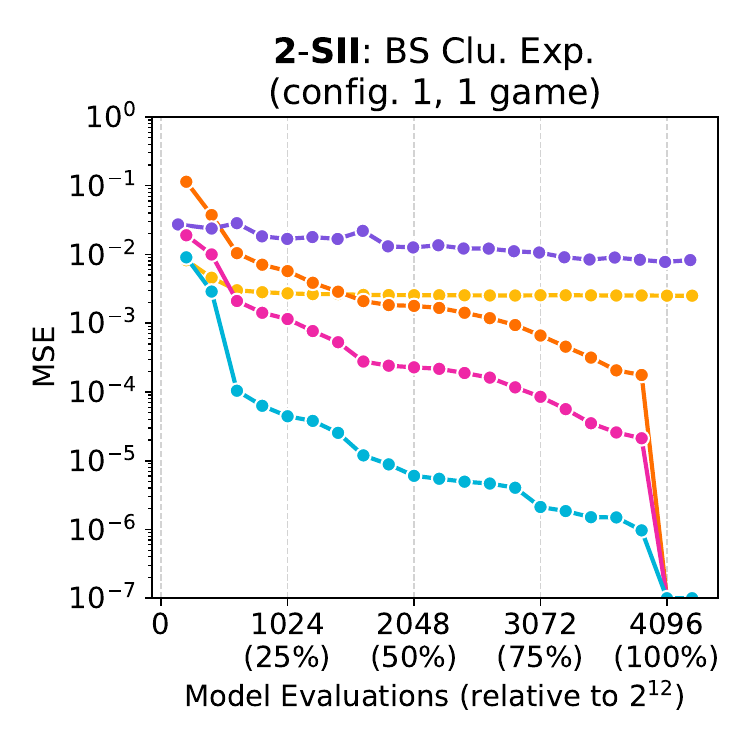}
    \end{minipage}
    \begin{minipage}[c]{0.24\textwidth}
        \includegraphics[width=\textwidth]{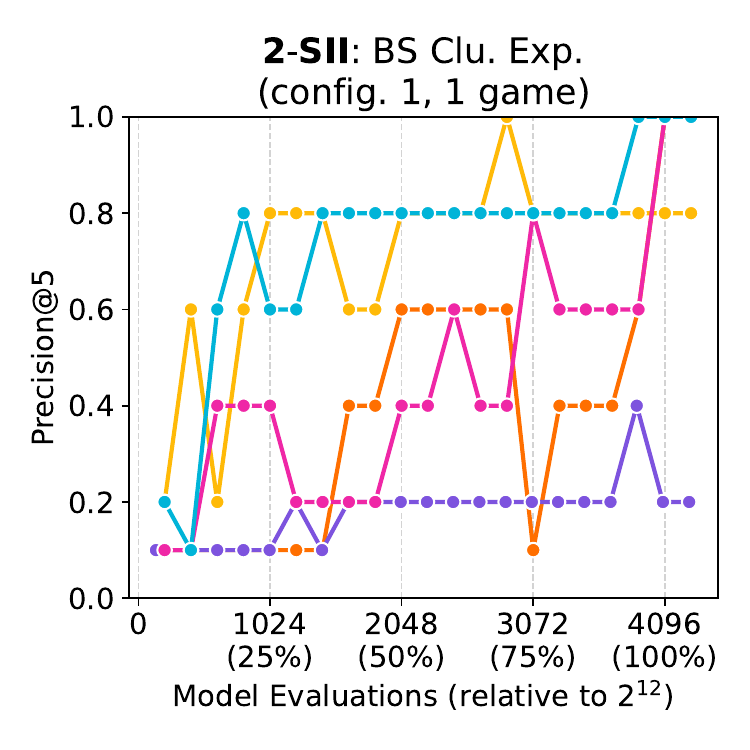}
    \end{minipage}
    \\
    \begin{minipage}[c]{0.24\textwidth}
        \includegraphics[width=\textwidth]{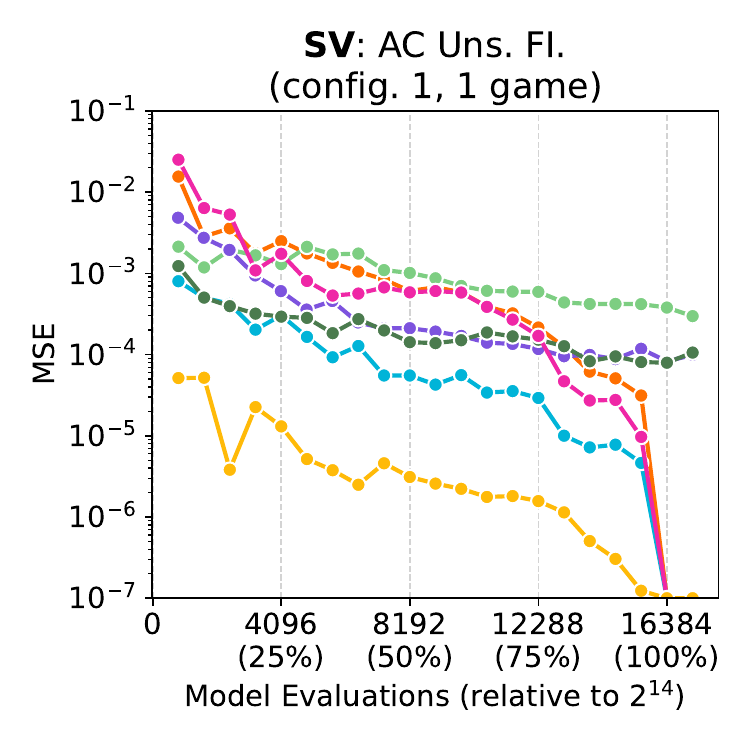}
    \end{minipage}
    \begin{minipage}[c]{0.24\textwidth}
        \includegraphics[width=\textwidth]{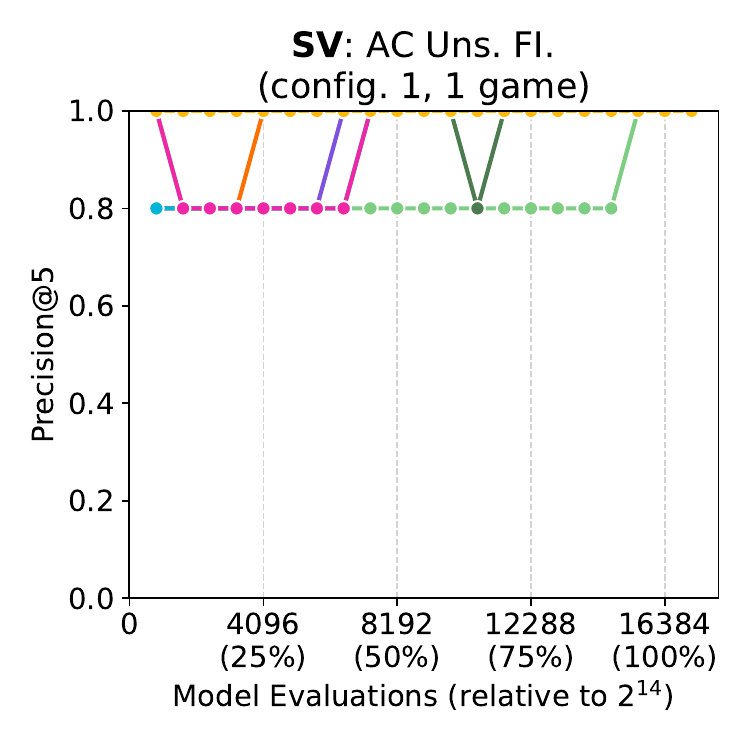}
    \end{minipage}
    \begin{minipage}[c]{0.24\textwidth}
        \includegraphics[width=\textwidth]{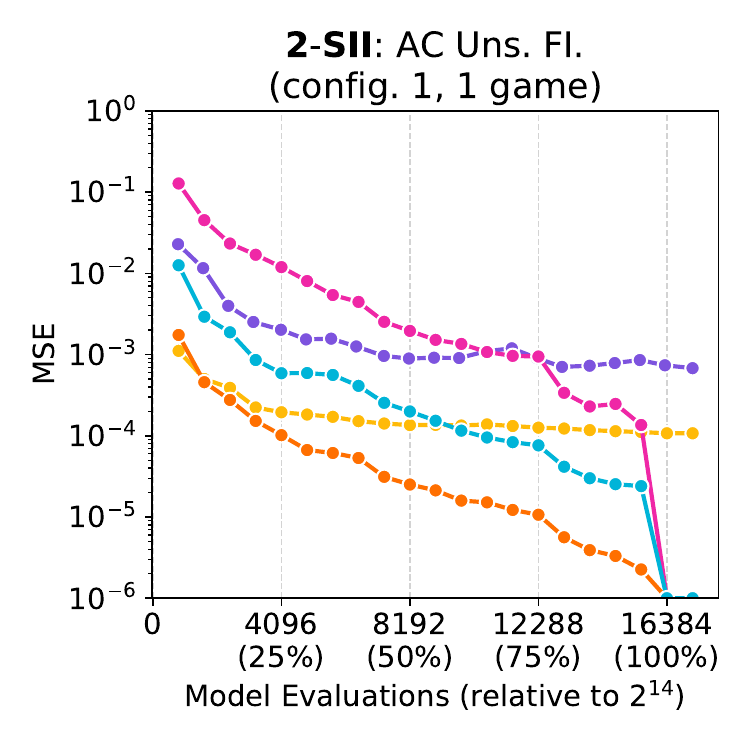}
    \end{minipage}
    \begin{minipage}[c]{0.24\textwidth}
        \includegraphics[width=\textwidth]{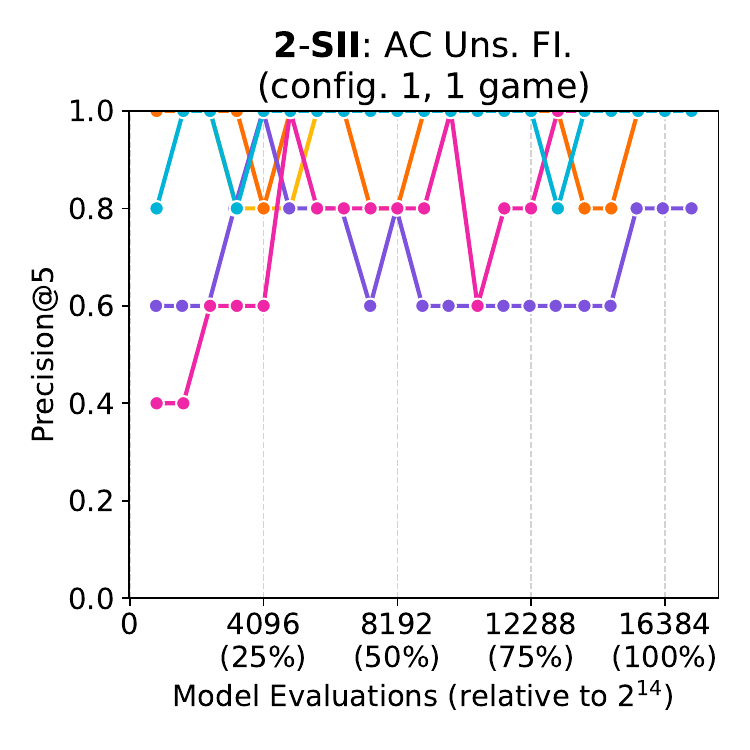}
    \end{minipage}
    \caption{Additional \gls{SV} (column one and two) and \gls{SI} (column three and four) approximation results for different benchmark games from the Uncertainty Explanation (first row, \textit{AdultCensus} with $n=14$ features), Cluster Explanation (second row, \textit{BikeSharing} with $n=12$ features), and Unsupervised Feature Importance (third row, dataset \textit{AdultCensus} with $n=14$ features) domain.}
    \label{appx_fig_approximation_3}
\end{figure}

This section contains additional experimental results. 
Mainly, this section contains exemplary \gls{MSE} and \gls{prec} approximation curves for a benchmark game of each application domain.
These results can be found in \cref{appx_fig_approximation_1,appx_fig_approximation_2,appx_fig_approximation_3}.
The dataset names used to for the benchmark games are abbreviated as described in \cref{appx_sec_overview_benchmark}.  
\cref{appx_figure_overview} shows the overview of the benchmark results additionally to \cref{fig_approximation} for the \gls{MAE}.

\begin{figure}[!htb]
    \centering
    \hspace{1.5em}
    \includegraphics[width=\textwidth]{figures/legend.pdf}
    \\[0.5em]
    \begin{minipage}[c]{0.32\textwidth}
        \includegraphics[width=\textwidth]{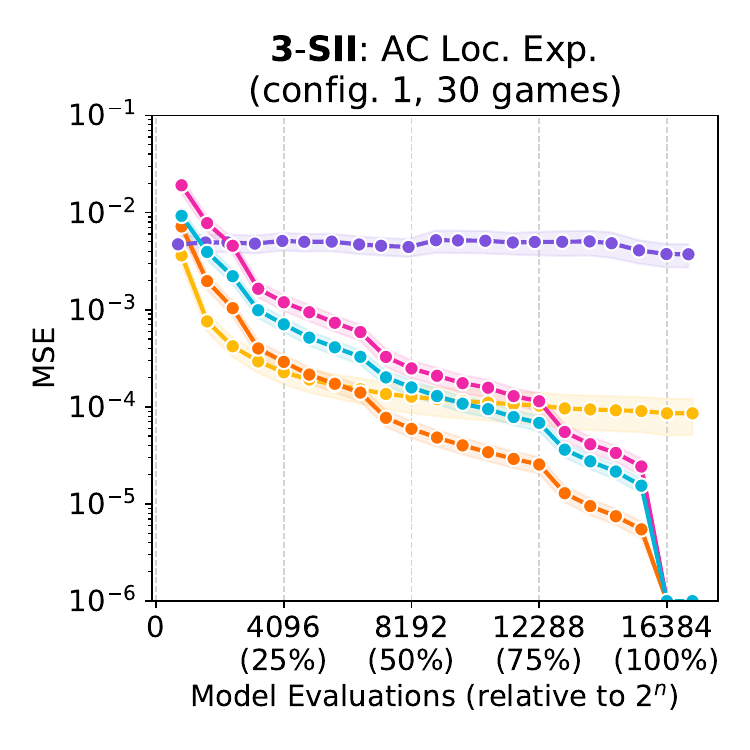}
    \end{minipage}
    \hfill
    \begin{minipage}[c]{0.32\textwidth}
        \includegraphics[width=\textwidth]{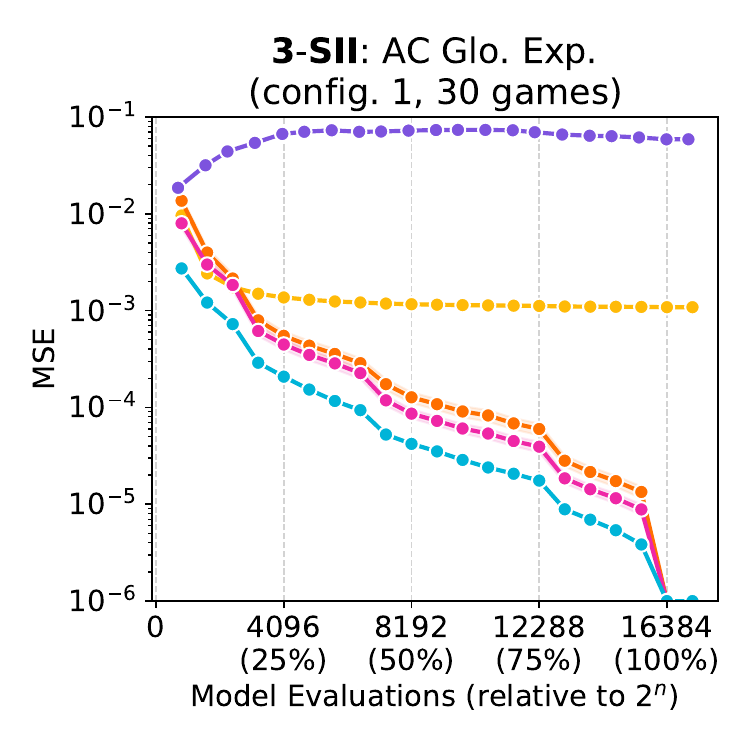}
    \end{minipage}
    \hfill
    \begin{minipage}[c]{0.32\textwidth}
        \includegraphics[width=\textwidth]{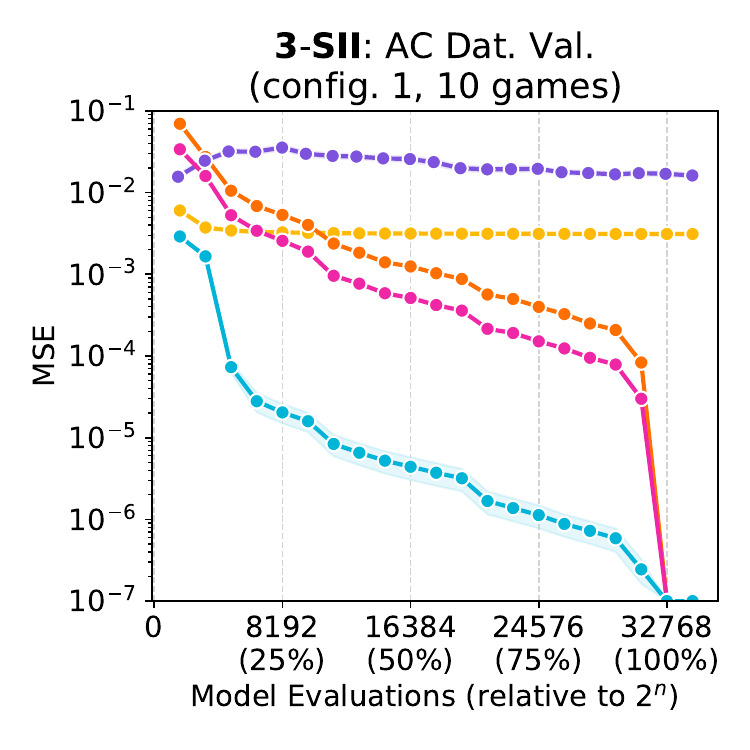}
    \end{minipage}
    \\
    \begin{minipage}[c]{0.32\textwidth}
        \includegraphics[width=\textwidth]{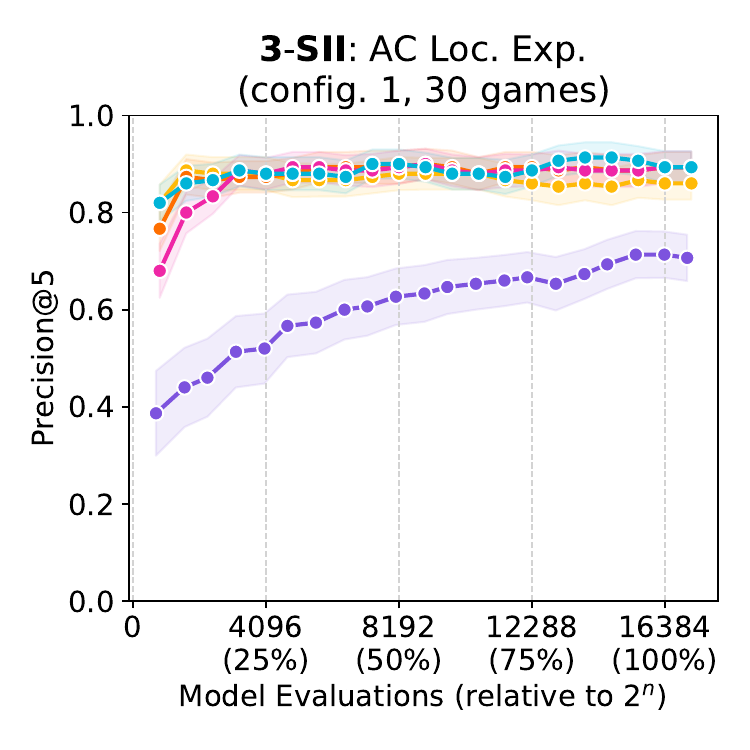}
    \end{minipage}
    \hfill
    \begin{minipage}[c]{0.32\textwidth}
        \includegraphics[width=\textwidth]{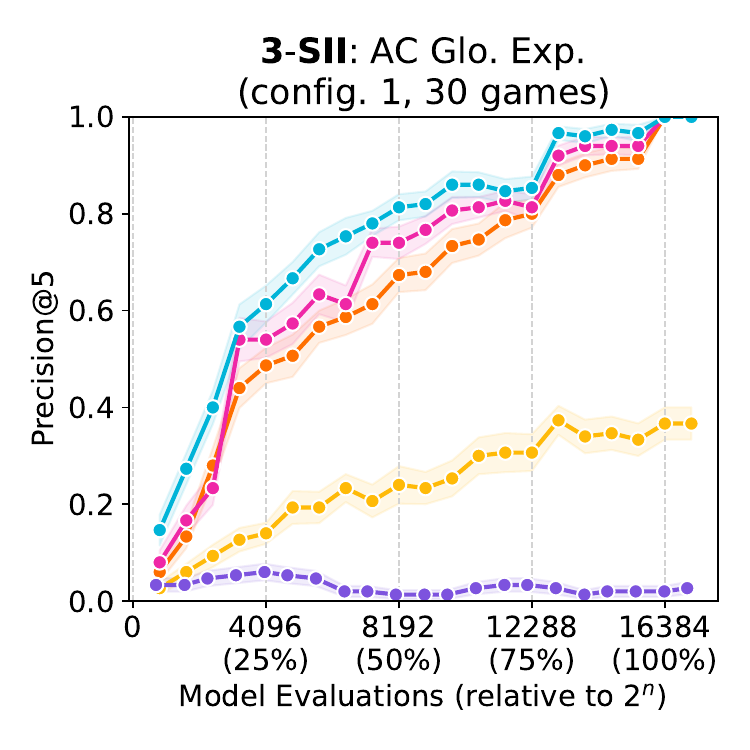}
    \end{minipage}
    \hfill
    \begin{minipage}[c]{0.32\textwidth}
        \includegraphics[width=\textwidth]{figures/appendix/approximation/Precision5_AdultCensus_LocalExplanation_Game_model_name=decision_tree_imputer=marginal_k-SII_3_n_games=30}
    \end{minipage}
    
    \caption{Approximation qualities in terms of \gls{MSE} \textbf{(top)} and \gls{prec} \textbf{(bottom)} for $3$-SII higher-order interactions for three benchmark settings based on the \textit{AdultCensus} (AC) dataset including Local Explanation \textbf{(left)}, Global Explanation \textbf{(middle)}, and Data Valuation \textbf{(right)}.}
    \label{appx_fig_3_SII}
\end{figure}

\clearpage
\section{Glossary of Acronyms}
\printglossary[type=\acronymtype, title=]

\end{document}